\documentclass{article}

\usepackage[main,final]{neurips_2026}

\usepackage[utf8]{inputenc}
\usepackage[T1]{fontenc}

\usepackage{xcolor}
\definecolor{ourblue}{RGB}{219,234,254}
\definecolor{citeblue}{rgb}{0.12,0.49,0.85}

\usepackage{graphicx}
\usepackage{booktabs}
\usepackage{url}
\usepackage{amsmath}
\usepackage{amssymb}
\usepackage{amsfonts}
\usepackage{mathtools}
\usepackage{nicefrac}
\usepackage{microtype}

\usepackage{subcaption}
\usepackage{multirow}
\usepackage{array}
\usepackage{makecell}
\usepackage{colortbl}
\usepackage{rotating}
\usepackage{longtable}

\usepackage{placeins}

\usepackage{amsthm}

\usepackage[colorlinks=true,citecolor=citeblue,linkcolor=citeblue,urlcolor=citeblue,pagebackref,breaklinks]{hyperref}
\usepackage{cleveref}

\theoremstyle{plain}
\newtheorem{lemma}{Lemma}[section]
\newtheorem{proposition}[lemma]{Proposition}

\newcommand{\R}{\mathbb{R}}
\newcommand{\vv}{\mathbf{v}}
\newcommand{\vw}{\mathbf{w}}
\newcommand{\ee}{\mathbf{e}}

\newcommand{\zz}{\mathbf{z}}
\newcommand{\vs}{\mathbf{s}}
\newcommand{\Ia}{\mathbf{I}_{\mathrm{mix}}}

\newcommand{\vmctl}[3]{%
  \makecell[c]{#1\\#2\,/\,#3}%
}

\newcommand{\vmcell}[4]{%
  \makecell[c]{%
    \makebox[2.5em][r]{#1}\,$\vert$\,\makebox[2.5em][r]{#2}\\
    \makebox[2.5em][r]{#3}\,$\vert$\,\makebox[2.5em][r]{#4}%
  }%
}

\title{Beyond Bit Matching: Orthogonal Watermarks for Collusion-Resistant Image Fingerprinting}

\author{%
  Zhakshylyk Nurlanov \qquad Tobias Wei\ss{}berg \qquad Florian Bernard \\
  University of Bonn \qquad Lamarr Institute
}

\hypersetup{
  pdftitle={Beyond Bit Matching: Orthogonal Watermarks for Collusion-Resistant Image Fingerprinting},
  pdfauthor={Zhakshylyk Nurlanov, Tobias Wei\ss{}berg, Florian Bernard}
}

\begin{document}

\maketitle

\begin{abstract}
Image fingerprinting assigns each distributed copy a user-specific watermark for source tracing. Averaging collusion is a central threat: several recipients can average their differently marked copies to suppress each individual mark. Discrete bit-string watermarks are especially vulnerable because averaging drives bit evidence toward ambiguous decisions. We introduce \textbf{VectorMark}, which maps user keys to near-orthogonal high-dimensional unit vectors and identifies colluders by cosine similarity to the extracted watermark direction. Under ideal averaging of $K$ vector keys, the normalized average preserves a cosine signal of order $1/\sqrt{K}$ for each colluder, while unrelated keys remain concentrated near zero. The spherical geometry of random directions gives a shifted-Beta null distribution for cosine similarities, enabling analytic per-key false-positive-rate control after validation on unwatermarked images. VectorMark uses a neural watermark encoder and centered extractor trained with JND masking and a progressive curriculum for photometric, geometric, and collusion robustness. Experiments on MS-COCO, OpenImages, and AI-generated images show that VectorMark maintains robust watermark attribution under photometric and geometric distortions, and achieves the best strict all-colluder detection among evaluated methods, while preserving high visual quality. Our code is publicly available: \url{https://github.com/nurlanov-zh/VectorMark}
\end{abstract}

% One paragraph lay summary:
% When a photographer or company shares an image with several people, it can hide a different invisible fingerprint in each copy to help trace an unauthorized leak. However, several recipients can combine their copies to weaken these fingerprints and hide where the leaked image came from. Many existing methods try to read a single sequence of zeros and ones, which becomes unreliable when different fingerprints are mixed. VectorMark instead hides distinctive numerical patterns that are designed to remain distinguishable after mixing. A trained neural network recovers the combined pattern, and the method searches for traces of several recipients rather than forcing it into one identifier. This improves identification of the contributing copies while keeping the image looking almost unchanged. In an averaging experiment, VectorMark identified all ten contributors exactly in 98% of cases among one million possible recipients. The approach could help creators and image distributors investigate leaks. Moreover, we show that the same mechanism can also help writing and extracting multiple provenance labels in one image.

\begin{figure}[tb]
  \centering
  \includegraphics[width=0.95\textwidth]{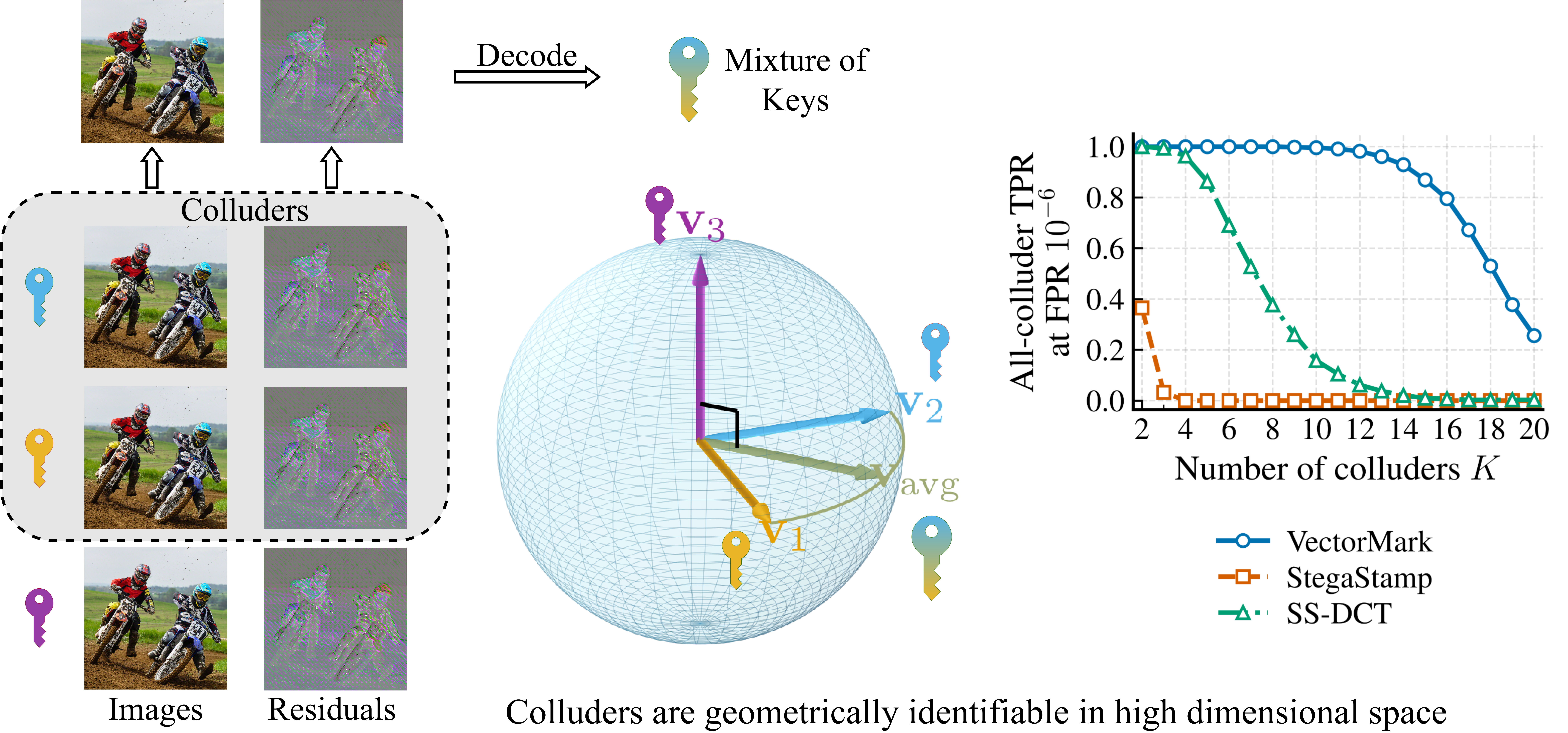}
  \caption{\textbf{Orthogonal watermarks preserve colluder evidence under averaging.}
\textbf{Left:} Under a $K$-user averaging attack, each individual image-space
watermark residual is attenuated by $1/K$.
\textbf{Center:} When user keys are represented as near-orthogonal directions
in $\R^D$, the normalized average direction retains cosine similarity
approximately $1/\sqrt{K}$ with each colluder key, while unrelated keys remain
near zero.
\textbf{Right:} VectorMark achieves near-perfect strict all-colluder detection
up to $K=10$ at per-key $\mathrm{FPR}=10^{-6}$, substantially outperforming
both discrete bit-matching and classical continuous-correlation baselines.}
  \label{fig:teaser}
\end{figure}

\section{Introduction}
\label{sec:intro}
The proliferation of generative AI has intensified long-standing concerns over
intellectual-property protection for digital media. A canonical protection
model is \emph{multi-user fingerprinting}: an author distributes
personalized copies of an image, each invisibly marked with a user-specific key
identifying the licensee~\cite{cox2008digital}. If a copy leaks, the embedded
key can trace it back to the responsible party. The problem becomes
substantially harder when several users collude by averaging, interleaving, or
otherwise combining their marked copies before release.

Most modern neural image watermarks expose a discrete payload interface
\cite{zhu2018hidden,bui2025trustmark}. This interface is natural for blind single-copy
recovery: the decoder predicts a bit string, which is then matched to a user
identifier. However, when several differently marked copies are averaged, bit
evidence is pushed toward an ambiguous middle state. Even if a detector still
recognizes that the image is watermarked, all-colluder attribution can fail
because no individual user's bit string remains cleanly recoverable.

VectorMark changes the object being recovered. Instead of decoding a bit string
directly, we map user keys to high-dimensional near-orthogonal unit vectors and
train an image encoder/extractor to preserve vector direction. Attribution is
then performed by searching a registered key database with cosine similarity.
Under an ideal equal-weight average of $K$ near-orthogonal keys, the normalized
average direction retains cosine similarity of order $1/\sqrt{K}$ with each
colluder key~\cite{muirhead1982aspects}, while unrelated keys remain
concentrated near zero. This separates the problem into two parts: the neural
watermark handles visual robustness, and high-dimensional directional retrieval
handles collusion attribution.

The same geometry also gives a principled detection rule. Under a centered
isotropic null, cosine similarities between unwatermarked extractions and fixed
registered keys follow a shifted-Beta distribution. We validate this model on held-out unwatermarked samples and use the resulting analytic threshold
for per-key false-positive-rate control. For baselines whose null model is
rejected, we instead use fitted null thresholds.

Our experiments show that VectorMark preserves strong single-key detection under
photometric and geometric distortions and achieves robust strict all-colluder
detection under averaging and non-averaging collusion attacks, including
coalitions up to $K=20$. In settings where bit-matching baselines collapse,
continuous directional retrieval preserves a usable attribution signal (\Cref{fig:teaser}).

Our contributions are:
\begin{enumerate}
  \item A neural watermarking formulation for \emph{near-orthogonal continuous
  fingerprinting}, where each registered user is represented by a
  high-dimensional direction paired with an optional discrete payload.
  \item Analytically calibrated per-key detection using the shifted-Beta null
  distribution of random-vector cosine similarities.
  \item A multi-user collusion attribution framework in which averaging
  preserves a $1/\sqrt{K}$ directional signal for each colluder.
  \item A collusion-aware training objective that encourages image-space
  averaging of marked copies to behave like embedding the corresponding average
  key direction.
  \item A progressive robustness curriculum that combines JND masking with
  photometric, geometric, and collusion augmentations, yielding strong visual
  quality and robust detection.
\end{enumerate}

\section{Related Work}
\label{sec:related}

\paragraph{Discrete neural watermarking.}
HiDDeN~\cite{zhu2018hidden} established the encoder-noise-decoder paradigm for
end-to-end learned image watermarking. Subsequent systems such as
StegaStamp~\cite{tancik2020stegastamp}, MBRS~\cite{jia2021mbrs},
RivaGAN~\cite{zhang2019robust}, TrustMark~\cite{bui2025trustmark},
VINE~\cite{lu2025robust}, MaskWM~\cite{hu2025mask}, VideoSeal~\cite{fernandez2024videoseal}, PixelSeal~\cite{soucek2025pixelseal}, and ChunkySeal~\cite{petrov2025chunkyseal} improve robustness,
localization, visual quality, or payload capacity. These methods primarily
recover discrete payloads, which is well suited to blind single-copy
provenance. However, after several differently marked copies are mixed,
colluder attribution requires recovering multiple user identities from a single
composite signal. This is poorly matched to exact bit-string recovery.

\paragraph{Continuous correlation watermarks.}
Classical spread-spectrum watermarking detects continuous signals by
correlation~\cite{cox1997secure,lin2001rotation,cox2008digital}. Such methods
are naturally more compatible with averaging than discrete bit matching,
because correlation scores can remain informative after signal attenuation.
However, classical spread-spectrum schemes are not trained end-to-end for the
photometric and geometric transformations encountered in modern image
distribution. VectorMark combines the attribution advantages of continuous
correlation with a neural encoder and extractor trained for robustness and
collusion.

\paragraph{Collusion-resistant fingerprinting.}
Classical fingerprinting codes, including Boneh--Shaw codes~\cite{boneh1998collusion}
and Tardos codes~\cite{tardos2008optimal}, provide traitor-tracing guarantees
under combinatorial marking assumptions. Tardos codes achieve length
$O(K^2\log(N/\varepsilon))$ for $N$ users and target error probability~$\varepsilon$. These guarantees do not directly transfer to
neural image watermarking, where payload capacity is limited and attacks occur
through real-valued image transformations. VectorMark instead uses
high-dimensional directional separation, colluder keys remain statistically
distinguishable from unrelated keys after averaging, and attribution is
performed by calibrated cosine retrieval.

\section{Method}
\label{sec:method}

VectorMark formulates multi-user image fingerprinting as directional retrieval
over a high-dimensional unit sphere. Each registered user is assigned a
near-orthogonal vector key. The image encoder embeds this direction into the
host image, and the extractor recovers a continuous direction from a possibly
distorted image. Detection is then performed by cosine similarity against a
registered key database.

\subsection{Interface}
\label{subsec:interface}
Let $I$ be a host image and let $u$ denote a registered user. VectorMark assigns
user $u$ a unit vector $\vv_u\in\mathbb{S}^{D-1}$. The encoder produces a personalized watermarked image
$I_u=E_\theta(I,\vv_u)$. Given a possibly distorted image $\tilde I$, the
extractor predicts $\hat{\zz}=F_\phi(\tilde I)$, which is centered and
normalized as
$\hat{\vw}=(\hat{\zz}-\mu_0)/\|\hat{\zz}-\mu_0\|_2$, where $\mu_0$ is the
held-out unwatermarked extractor mean. Detection uses the cosine score
$s_u=\langle \hat{\vw},\vv_u\rangle$ and returns all registered users with
$s_u\geq\tau$. The threshold
$\tau$ is calibrated from the null distribution described in
\Cref{sec:calibration}. For non-colluded images, VectorMark can additionally
decode an optional discrete payload from the same continuous representation.

\subsection{Near-Orthogonal Keys}
\label{subsec:keys}
We map a binary user key $m_i\in\{0,1\}^L$ to a random linear codeword and then
to a unit vector by
\begin{equation}
  c_i=Gm_i \bmod 2,\qquad
  \ee_i=2c_i-\mathbf{1},\qquad
  \vv_i=\frac{R\ee_i}{\sqrt{D}},
  \label{eq:codeword}
\end{equation}
where $G\in GF(2)^{D\times L}$ is a fixed random binary matrix and
$R\in O(D)$ is a fixed random orthogonal matrix. The projection back to code
coordinates is $\Pi_R(\vv)=\sqrt{D}R^\top\vv$. For an extracted direction
$\hat{\vw}$, $\Pi_R(\hat{\vw})$ gives soft random-linear-code observations for
optional payload decoding.

The orthogonal rotation preserves pairwise inner products:
\begin{equation}
  \langle \vv_i,\vv_j\rangle
  =
  D^{-1}\langle \ee_i,\ee_j\rangle .
\end{equation}
For any two fixed distinct user keys $m_i$ and $m_j$, the random linear mapping
in \Cref{eq:codeword} makes the corresponding vector-key inner product have
mean zero and variance $1/D$. Thus, registered keys are
near-orthogonal with high probability (\Cref{sec:supp_theory_rewrite}).

\subsection{Collusion Geometry}
\label{subsec:collusion_geometry}

Consider equal-weight averaging of $K$ marked copies,
$\Ia=K^{-1}\sum_{i=1}^{K}E_\theta(I,\vv_i)$. If the encoder/extractor pair is
approximately linear around the embedded signal and the keys are
near-orthogonal, then the extracted direction is close to the normalized key
average. For ideal orthonormal keys,
\begin{equation}
  \left\langle
  \frac{\sum_{i=1}^{K}\vv_i}{\|\sum_{i=1}^{K}\vv_i\|_2}, \vv_j
  \right\rangle
  = \frac{1}{\sqrt{K}}.
  \label{eq:collusion_scaling}
\end{equation}
This equation gives the ideal cosine-similarity scaling for a $K$-colluder
averaging attack. In practice, encoder and extractor nonlinearity, recovery
noise, non-averaging attacks, image distortions, and key correlations
can reduce the observed margin. The training objective below encourages the
learned watermark to approximate this ideal directional behavior.

\subsection{Training Objective}
\label{subsec:training_summary}

\begin{figure}[t]
    \centering
    \includegraphics[width=\textwidth]{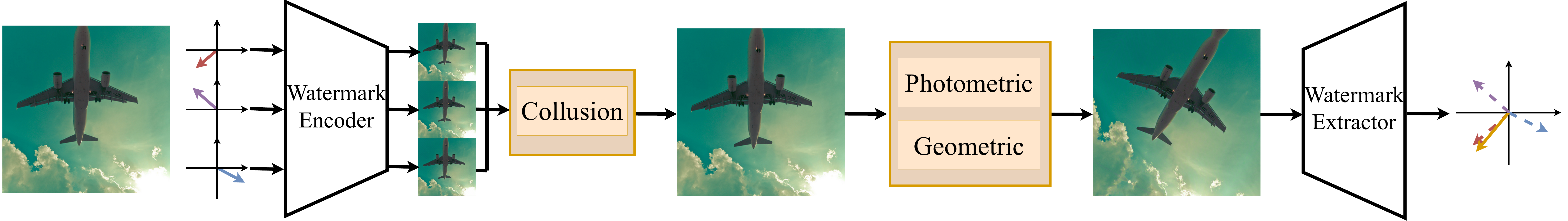}
      \caption{\textbf{VectorMark architecture.}
      Watermark encoder conditions on a continuous key vector $\vv$ and embeds a
      JND-masked residual into the host image $I$. During training,
      differentiable photometric, geometric, and collusion augmentations enforce
      robustness. A watermark extractor recovers a continuous vector, which is
      centered using the mean of unwatermarked images and normalized for
      cosine-similarity detection against registered keys.}
    \label{fig:pipeline}
\end{figure}

VectorMark uses U-Net-style encoder and extractor networks~\cite{ronneberger2015unet}.
The encoder converts the key vector into a spatial conditioning map,
concatenates it with the host image, and predicts an additive residual. This
residual is attenuated by a Just-Noticeable-Difference (JND) mask
\cite{wu2017enhanced} to concentrate watermark energy in perceptually tolerant
regions. The extractor maps the image back to a continuous vector through
U-Net features and a learned readout head. A centering layer subtracts the
held-out unwatermarked mean before cosine matching or payload decoding.

For single-key training, let $T\sim\mathcal{T}$ be a differentiable image
transformation sampled from the training distortion distribution. For
$I_\vv=E_\theta(I,\vv)$, define
$\bar F_\phi(\tilde I)=F_\phi(\tilde I)-\mu_0$ and
$\hat{\vw}_T=\bar F_\phi(T(I_\vv))/\|\bar F_\phi(T(I_\vv))\|_2$.
With target codeword $\ee=\sqrt{D}R^\top\vv$ and soft prediction
$\hat{\ee}=\sqrt{D}R^\top\bar F_\phi(T(I_\vv))$, the single-key loss is
\begin{equation}
\mathcal{L}_{\mathrm{single}}
=
\lambda_{\mathrm{img}}\operatorname{MSE}(I_\vv,I)
+
\lambda_{\mathrm{vec}}(1-\langle \hat{\vw}_T,\vv\rangle)
+
\frac{\lambda_{\mathrm{code}}}{D}\sum_{d=1}^{D}
\mathrm{ReLU}(1-\ee_d\hat{\ee}_d).
\label{eq:training_single}
\end{equation}
During collusion phases, we sample $K$ keys and define
$\vv_{\mathrm{avg}}=K^{-1}\sum_j\vv_j$,
$\bar{\vv}_{\mathrm{avg}}=\vv_{\mathrm{avg}}/\|\vv_{\mathrm{avg}}\|_2$,
$I_{\mathrm{mix}}=K^{-1}\sum_jE_\theta(I,\vv_j)$, and
$I_{\mathrm{dir}}=E_\theta(I,\bar{\vv}_{\mathrm{avg}})$. The collusion loss is
\begin{equation}
\begin{aligned}
\mathcal{L}_{\mathrm{col}}
&=
\lambda_{\mathrm{lin}}\operatorname{MSE}(I_{\mathrm{mix}},I_{\mathrm{dir}})
+
\lambda_{\mathrm{vec}}\Big[
2
-\cos(\bar F_\phi(T(I_{\mathrm{mix}})),\vv_{\mathrm{avg}})
-\cos(\bar F_\phi(T(I_{\mathrm{dir}})),\vv_{\mathrm{avg}})
\Big].
\end{aligned}
\label{eq:training_collusion}
\end{equation}
The first term encourages image-space averaging to match direct embedding of
the normalized average key, while the cosine terms train the extractor to
recover the average direction from both images. Codeword supervision is kept
only on the single-key path because $\vv_{\mathrm{avg}}$ is not a valid
codeword.

The total objective is
\begin{equation}
  \min_{\theta,\phi}
  \mathbb{E}_{I,\vv,T}\!\left[\mathcal{L}_{\mathrm{single}}\right]
  +
  \mathbf{1}_{\mathrm{col}}
  \mathbb{E}_{I,\vv_1,\ldots,\vv_K,T}\!\left[
  \mathcal{L}_{\mathrm{col}}\right],
  \label{eq:training_total}
\end{equation}
where $\mathbf{1}_{\mathrm{col}}$ is active only in scheduled collusion phases.

\section{Calibration and Scaling}
\label{sec:calibration}

\subsection{Null Model}
\label{subsec:null}

For a fixed registered unit vector $\vv$ and a null extracted direction
$\hat{\vw}$ from an unwatermarked image, assume that $\hat{\vw}$ is isotropic
on $\mathbb{S}^{D-1}$. Then the cosine similarity
$S=\langle \hat{\vw},\vv\rangle$ satisfies
\begin{equation}
  \frac{1+S}{2}\sim
  \mathrm{Beta}\!\left(\frac{D-1}{2},\frac{D-1}{2}\right).
  \label{eq:beta_null}
\end{equation}
We use this shifted-Beta distribution to set analytic per-key false-positive
thresholds. For $D=1024$ and per-key $\mathrm{FPR}=10^{-6}$, the resulting
threshold is approximately $\tau=0.148$.
We validate this assumption on held-out unwatermarked extractions paired
with random registered keys. For VectorMark, the empirical cosine distribution
is consistent with the shifted-Beta prediction, so we
 use the analytic threshold without per-query calibration. The
centering layer subtracts the held-out unwatermarked mean, removing systematic null bias. We apply the same
validation to continuous baselines. When the spherical null is rejected, as for
SS-FM due to a large nonzero null mean, we fit a parametric null distribution on
the held-out set and use its threshold instead.

\subsection{Database-Level Scaling}
\label{subsec:database}

The single-key benchmarks and initial collusion comparisons use a fixed
\emph{per-key} FPR. For a deployment with $N$
registered keys and family-wise false-positive budget $\delta$, a conservative
Bonferroni correction uses $\alpha_{\mathrm{per}}=\delta/N$. The higher threshold reduces the
ideal collusion capacity in \Cref{eq:collusion_scaling}.
\Cref{subsec:registry} evaluates full registries up to $N=10^6$.
\Cref{sec:supp_registry} gives the additional protocol details and scaling analysis.

VectorMark supports two complementary operating modes. In the single-user or
non-colluded setting, the optional blind payload decoder can recover an
explicit user ID, matching the interface of conventional discrete watermarking
systems. In the collusion or forensic setting, the extracted continuous vector
is compared against registered database keys, and all keys whose cosine scores
exceed the calibrated threshold are returned. For small and medium databases,
this search can be exact matrix-vector multiplication. For larger databases,
standard approximate nearest-neighbor retrieval can be used for candidate
generation. Payload decoding
provides compatibility with existing watermark deployments, while cosine search
is the mechanism that remains informative when averaging collusion destroys
exact bit recovery.
\section{Experiments}
\label{sec:experiments}
We evaluate VectorMark along four axes: image quality, single-key detection,
robustness to benign transformations, and multi-user attribution under
collusion. The central question is whether a continuous near-orthogonal key
space preserves enough directional signal to identify all colluders after their
marked copies are combined. 

\subsection{Setup}
\label{subsec:setup}

\paragraph{Training and evaluation.}
We train VectorMark on the MS-COCO 2014 train split
\cite{lin2014microsoft}, containing 82,783 images, using
$256\times256$ crops. At evaluation time, watermarks are embedded at the
model resolution and the residual is scaled to $512\times512$, following
TrustMark~\cite{bui2025trustmark}. We evaluate on 1,000 MS-COCO 2014
validation images, 1,000 OpenImages V7 validation images
\cite{kuznetsova2020openimages}, and 1,000 AI-generated images.
The AI-generated set contains 200 images from each of five generators:
DALL-E 3, FLUX, OpenAI-4o, Nano-Banana, and SDXL
\cite{christodoulou2024finding,rapidata_flux_sd3_mj_dalle3_hf,
rapidata_openai4o_t2i_human_preference_hf,bitmind_nano_banana_hf,
ash12321_sdxl_generated_10k_hf}. Unless stated otherwise, VectorMark uses
dimension $D=1024$ and is trained with AdamW~\cite{loshchilov2019decoupled}.

\paragraph{Metrics.}
We report image quality using \emph{PSNR} and \emph{LPIPS}. Detection is measured as \emph{true
positive rate (TPR)} at a fixed per-key \emph{false positive rate}
$\mathrm{FPR}=10^{-6}$. For continuous-key methods, the detection score is the
cosine similarity between the extracted direction and a registered key. The
threshold is set from the shifted-Beta null when the null validation passes,
and from a fitted held-out null distribution otherwise. For binary-key methods,
the score is bit agreement and the threshold is set using the corresponding
binomial null model.

For single-key detection, a trial is positive if the embedded key is detected.
For collusion experiments with $K$ users, we report strict \emph{all-colluder TPR}: a trial succeeds when all $K$ true colluders are detected. In full-registry experiments, \emph{exact-set recovery}
requires detecting all and only the true colluders, \emph{recall} measures the detected fraction of true colluders, and \emph{precision} is their fraction in the returned detected set. \emph{No-FA} is the rate of
returning no innocent identity, and \emph{empirical family-wise error rate (FWER)} is $1-\mathrm{No\text{-}FA}$.

We evaluate robustness to benign transformations commonly encountered during
image transmission. The photometric suite contains 15 configurations, including
JPEG compression, Gaussian blur and noise, median filtering, salt-and-pepper
noise, color and brightness shifts, and downscaling. The geometric suite
contains 30 configurations, including horizontal flips, rotations
($5^\circ$--$15^\circ$) with different padding strategies, perspective warps,
crop-and-resize, non-uniform stretching, and pad-and-resize transforms. The photometric and geometric columns report macro-averaged values across transformations.

\paragraph{Baselines.}
We compare against nine binary-payload watermarking methods:
HiDDeN \cite{zhu2018hidden}, DWT-DCT \cite{al2007combined},
DWT-DCT-SVD \cite{navas2008dwt}, RivaGAN \cite{zhang2019robust},
StegaStamp \cite{tancik2020stegastamp}, MBRS \cite{jia2021mbrs},
TrustMark \cite{bui2025trustmark}, VINE \cite{lu2025robust}, and
MaskWM \cite{hu2025mask}. We use the checkpoints from official repositories. To separate the effect of continuous keys from the
effect of learned robustness, we also compare against three continuous
spread-spectrum baselines: SS-FM \cite{lin2001rotation}, SS-DCT, and
SS-DFT \cite{cox1997secure}. Each method is embedded at its native resolution
when applicable, and the residuals are resized to $512\times512$ for a common
evaluation resolution.

\subsection{Main MS-COCO Results}
\label{subsec:main_results}

\begin{table}[tb]
\caption{\textbf{MS-COCO detection and collusion attribution.}
We compare VectorMark with continuous-key and binary-key baselines at per-key
$\mathrm{FPR}=10^{-6}$. Robustness columns report single-key TPR on clean,
photometric, and geometric transformations. Collusion columns report strict
all-colluder TPR under $K$-user averaging attacks. VectorMark is the only
evaluated method that combines high geometric robustness with near-perfect
strict all-colluder attribution up to $K=10$.}
  \label{tab:main_results_coco}
  \centering
  \setlength{\tabcolsep}{3pt}
  \small
  \begin{tabular}{@{} c l r  r c  c c c  c c c @{}}
    \toprule
    & & & \multicolumn{2}{c}{Quality}
        & \multicolumn{3}{c}{Robustness}
        & \multicolumn{3}{c}{All-colluder detect.} \\
    \cmidrule(lr){4-5}\cmidrule(lr){6-8}\cmidrule(l){9-11}
    & Method & Dim
      & PSNR$\uparrow$ & LPIPS$\downarrow$
      & Clean & Photo. & Geo.
      & $K{=}2$ & $K{=}5$ & $K{=}10$ \\
    \midrule
    \multirow{4}{*}{\rotatebox[origin=c]{90}{\itshape Cont. keys}}
    & \textbf{VectorMark}
      & 1024 & 40.29 & .041
      & \textbf{1.00} & \textbf{1.00} & \textbf{.985}
      & \textbf{1.00} & \textbf{1.00} & \textbf{.996} \\
    & SS-DCT~\cite{cox1997secure}
      & 1024 & 34.25 & .198 & {1.00} & {1.00} & .000
      & \underline{1.00} & .829 & .092 \\
    & SS-DFT~\cite{cox1997secure}
      & 1024 & 33.85 & .207 & {1.00} & .997 & .000
      & \underline{.999} & \underline{.945} & \underline{.171} \\
    & SS-FM~\cite{lin2001rotation}
      & 74 & 32.84 & .103 & .813 & .680 & .041
      & .060 & .000 & .000 \\
    \midrule
    \multirow{9}{*}{\rotatebox[origin=c]{90}{\itshape Binary keys}}
    & VINE~\cite{lu2025robust}
      & 100 & 36.98 & {.017} & {1.00} & {1.00} & .000
      & .307 & .000 & .000 \\
    & TrustMark~\cite{bui2025trustmark}
      & 100 & 40.53 & {.026} & {1.00} & .978 & .349
      & \underline{.346} & .000 & .000 \\
    & StegaStamp~\cite{tancik2020stegastamp}
      & 100 & 29.34 & .112 & {1.00} & {1.00} & .161
      & \underline{.365} & .000 & .000 \\
    & MaskWM~\cite{hu2025mask}
      & 64 & 36.42 & .146 & {1.00} & {1.00} & .409
      & .070 & .000 & .000 \\
    & MBRS~\cite{jia2021mbrs}
      & 64 & {43.06} & .029 & {1.00} & .967 & .000
      & .052 & .000 & .000 \\
    & HiDDeN~\cite{zhu2018hidden}
      & 30 & 32.83 & .099 & .293 & .425 & .048
      & .000 & .000 & .000 \\
    & RivaGAN~\cite{zhang2019robust}
      & 32 & 39.99 & .076 & .996 & .971 & \underline{.491}
      & .000 & .000 & .000 \\
    & DWT-DCT~\cite{al2007combined}
      & 64 & 36.79 & .056 & .716 & .467 & .003
      & .022 & .000 & .000 \\
    & DWT-DCT-SVD~\cite{navas2008dwt}
      & 64 & 36.76 & .042 & .996 & .561 & .005
      & .080 & .000 & .000 \\
    \bottomrule
  \end{tabular}
\end{table}

\Cref{tab:main_results_coco} summarizes the in-domain MS-COCO results.
Several binary-payload methods achieve strong clean and photometric detection,
and some preserve high image quality. However, their attribution performance
collapses under multi-user averaging: at $K=5$, the short-payload binary-key baselines have
zero strict all-colluder TPR. Continuous spread-spectrum baselines are more
resistant to averaging, confirming the value of continuous correlation scores,
but they are brittle to geometric transformations. VectorMark combines both
properties: it achieves high visual quality, near-perfect clean and
photometric detection, strong geometric robustness, and near-perfect strict
all-colluder detection through $K=10$. Detailed per-noise robustness results for single-key detection are given in \Cref{tab:supp_tpr_all}.

\subsection{Cross-Domain Generalization}

\begin{table}[tb]
\centering
\caption{\textbf{Cross-domain evaluation on OpenImages and AI-generated images.}
We report image quality, single-key robustness, and strict all-colluder TPR at
per-key $\mathrm{FPR}=10^{-6}$. VectorMark preserves both geometric robustness
and multi-user attribution across real and generated image domains.}
\label{tab:cross_domain}
\small
\begin{tabular}{@{}llccccccc@{}}
\toprule
Dataset & Method & PSNR & Clean & Photo. & Geo. & $K{=}2$ & $K{=}5$ & $K{=}10$ \\
\midrule
OpenImages & \textbf{VectorMark} & 40.36 & \textbf{1.00} & \textbf{1.00} & \textbf{1.00} & \textbf{1.00} & \textbf{1.00} & \textbf{1.00} \\
OpenImages & SS-DFT & 35.22 & .998 & .997 & .249 & .994 & .934 & .184 \\
OpenImages & TrustMark & 40.81 & {1.00} & {1.00} & .750 & .369 & .000 & .000 \\
\midrule
AI-generated & \textbf{VectorMark} & 41.53 & \textbf{1.00} & \textbf{1.00} & \textbf{1.00} & \textbf{1.00} & \textbf{1.00} & \textbf{1.00} \\
AI-generated & SS-DFT & 36.68 & {1.00} & {1.00} & .250 & {1.00} & .938 & .194 \\
AI-generated & TrustMark & 42.21 & {1.00} & {1.00} & .750 & .388 & .000 & .000 \\
\bottomrule
\end{tabular}
\end{table}

\Cref{tab:cross_domain} evaluates generalization of VectorMark to other domains, including natural images from OpenImages and a dataset of
AI-generated images. We compare with the post-hoc watermarking methods, so the watermarking methods integrated into image generation are outside our evaluation scope. VectorMark maintains 1.00 TPR for clean, photometric, and
geometric single-key detection on both datasets, and achieves 1.00 strict
all-colluder TPR through $K=10$. The comparison with SS-DFT and TrustMark
mirrors the MS-COCO results: SS-DFT remains partially robust to averaging but
has much lower geometric robustness, while TrustMark remains strong for
single-key detection but fails strict all-colluder recovery once $K\geq5$.

\subsection{Collusion Beyond Averaging}

\begin{figure}[tb]
\centering
\includegraphics[width=0.92\linewidth]{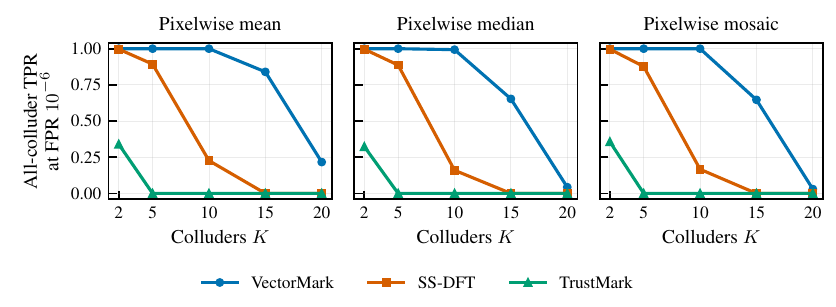}
\caption{\textbf{Collusion attacks on OpenImages.}
Each curve shows strict all-colluder TPR at per-key
$\mathrm{FPR}=10^{-6}$ for up to $K=20$ colluders. We compare standard
averaging collusion with pixelwise median and pixelwise mosaic non-averaging
collusions. TrustMark uses bit-agreement scores with a binomial threshold;
VectorMark and SS-DFT use calibrated cosine-similarity thresholds.}
\label{fig:nonaverage_collusion}
\end{figure}

Averaging is the canonical collusion attack and the one most directly explained
by the $1/\sqrt{K}$ geometry. We also evaluate two non-averaging attacks:
pixelwise median collusion and pixelwise mosaic collusion. In the median
attack, each output pixel/channel is the median across the $K$ marked copies.
In the mosaic attack, each output pixel is sampled from one of the marked
copies, so local watermark evidence comes from individual colluders rather
than from an arithmetic average.
\Cref{fig:nonaverage_collusion} shows strict all-colluder TPR on OpenImages for
up to $K=20$. VectorMark remains substantially more robust than TrustMark and
SS-DFT across all three attacks. Performance is strongest under mean averaging,
as predicted by the linear geometry, and degrades under median and mosaic
collusion. Nevertheless, the extractor continues to recover directions
correlated with the participating keys, enabling multi-user attribution beyond
the averaging setting used in the main training objective.

\subsection{Payload Recovery}

\begin{table}[tb]
\centering
\caption{\textbf{MS-COCO payload recovery accuracy.}
VectorMark decodes a 100-bit payload from a 1024-dimensional random linear
codeword. We report raw codeword sign accuracy, single-pass ECC-decoded payload
accuracy, and blind two-pass ECC-decoded payload accuracy. Baselines are
reported using their native payload lengths and sorted by mean accuracy over
all transformations.}
\label{tab:main_payload_coco}
\setlength{\tabcolsep}{2pt}
\scriptsize
\resizebox{\textwidth}{!}{%
\begin{tabular}{@{}lcccccccccccc@{}}
\toprule
& \multicolumn{3}{c}{VectorMark} & \multicolumn{9}{c}{Baselines, sorted by all-transform mean} \\
\cmidrule(lr){2-4}\cmidrule(l){5-13}
Setting & \makecell{Ours\\raw} & \makecell{Ours\\ECC} & \makecell{Ours\\2-pass\\ECC} & \makecell{Riva\\GAN} & MaskWM & \makecell{Trust\\Mark} & HiDDeN & \makecell{Stega\\Stamp} & VINE & MBRS & \makecell{DWT-\\DCT\\SVD} & DWT-DCT \\
\textit{bits predicted} & $1024^b$ & $100^b$ & $100^b$ & $32^b$ & $64^b$ & $100^b$ & $30^b$ & $100^b$ & $100^b$ & $64^b$ & $64^b$ & $64^b$ \\
\midrule
Clean & \textbf{1.00} & \textbf{1.00} & \textbf{1.00} & .997 & \textbf{1.00} & \textbf{1.00} & .884 & .999 & \textbf{1.00} & \textbf{1.00} & .998 & .855 \\
Photometric & .945 & .956 & .957 & .987 & \textbf{1.00} & .981 & .866 & .991 & .994 & .983 & .770 & .730 \\
Geometric & .838 & .827 & \textbf{.975} & .787 & .725 & .668 & .686 & .588 & .500 & .499 & .535 & .519 \\
All trans. avg. & .874 & .870 & \textbf{.969} & .854 & .816 & .772 & .746 & .722 & .665 & .660 & .613 & .589 \\
\bottomrule
\end{tabular}%
}
\end{table}

Although cosine retrieval is the primary forensic interface, VectorMark can also
recover a conventional blind payload in the single-key setting. The extracted
continuous direction is rotated back into the codeword basis, producing soft
bipolar reliability scores. Because each 100-bit user payload is mapped to a
1024-dimensional random linear codeword, we decode the payload using
reliability-ordered OSD-0~\cite{fossorier1995soft}.
\Cref{tab:main_payload_coco} reports payload bit accuracy on MS-COCO. The raw
column measures sign accuracy over the 1024-dimensional codeword, while the ECC
columns measure decoded 100-bit payload accuracy. The two-pass decoder evaluates
both the original image and a horizontally flipped version, then selects the
decoded payload with the higher internal codeword-match score. This simple
blind test substantially improves geometric robustness. Detailed per-noise results are shown in \Cref{tab:supp_payload_scores}. Overall, VectorMark
retains competitive single-key payload recovery while providing the continuous
retrieval mechanism needed for collusion attribution.

\subsection{Qualitative Results}

\begin{figure}[tb]
  \centering
    \setlength{\tabcolsep}{2pt}  % horizontal spacing between images
    \renewcommand{\arraystretch}{1.2}  % vertical spacing between rows

    \begin{tabular}{cccc}
        \small\textbf{Original} &
        \small\textbf{Watermarked} &
        \small\textbf{Residual} &
        \small\textbf{Distorted Sample} \\[4pt]

        \includegraphics[width=0.225\linewidth]{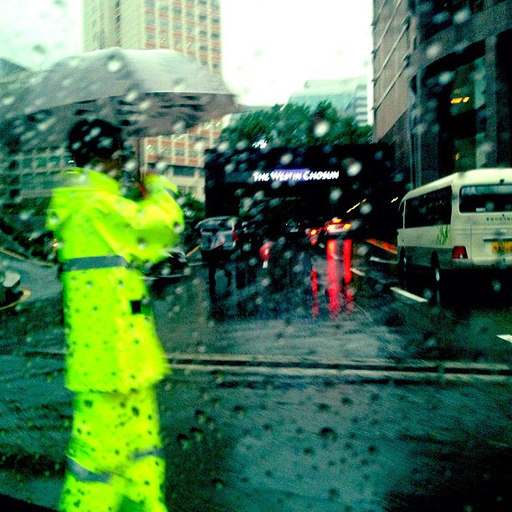} &
        \includegraphics[width=0.225\linewidth]{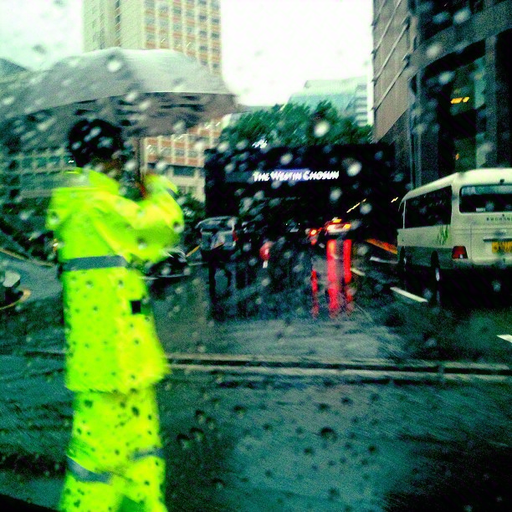} &
        \includegraphics[width=0.225\linewidth]{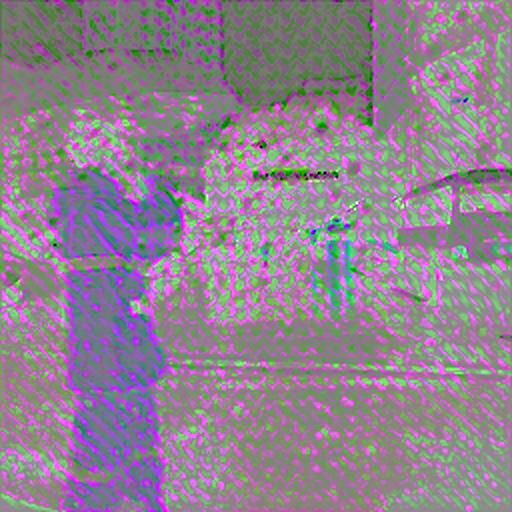} &
        \includegraphics[width=0.225\linewidth]{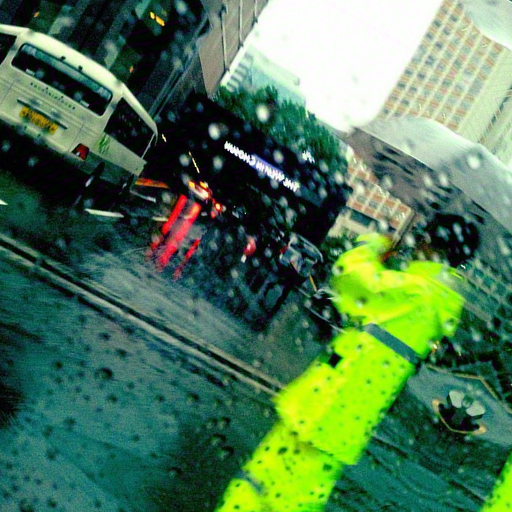} \\

    \end{tabular}
\caption{\textbf{Qualitative VectorMark watermark examples.}
From left to right: original image, watermarked image, absolute residual
magnified by $10\times$, and a representative distorted sample. The JND mask
concentrates watermark energy in perceptually tolerant regions.}
  \label{fig:quality}
\end{figure}

Quantitative metrics such as PSNR and LPIPS are useful proxies for image
quality, but they do not fully capture the visibility of structured watermark
artifacts \cite{gowal2025synthid}. \Cref{fig:quality} therefore shows original
images, watermarked images, magnified residuals, and representative distorted
samples. VectorMark concentrates residual energy in textured regions through the
JND mask \cite{wu2017enhanced}, where human perception is less sensitive to
small perturbations. This helps preserve visual quality while maintaining
extractability under distortions.

\subsection{Database-Scale Attribution}
\label{subsec:registry}

To evaluate performance of VectorMark in a realistic large-registry setting with a large-scale database of user identities, we generate registries of different sizes ($N=10^3$--$10^6$) with randomly sampled user identities. There, we select $K$ users who carry out a collusion attack by mixing their marked images. For user attribution, we test every identity in registry by computing similarity score against the predicted watermark vector and select the ones with similarities exceeding pre-determined threshold values. The thresholds are set analytically to satisfy fixed false positive rates. Per-key FPR $\alpha=\delta/N$ is set for a family-wise false-positive budget $\delta=0.01$.
We use 2,000 test
images and three independently sampled registries. \Cref{tab:registry_main} shows that at $N=10^6$, $K=10$, VectorMark achieves 0.998 recall,
0.999 precision, 0.990 all-colluder TPR, and 0.980 exact-set recovery.
Empirical family-wise error rate (FWER) is 0.011, close to the nominal 1\% budget.
At the same registry size, exact-set recovery decreases to 0.521 for $K=15$
and 0.007 for $K=20$. \Cref{tab:registry_full} reports all coalition sizes.

\begin{table}[tbp]
\centering
\caption{\textbf{Registry scaling evaluation with $K=10$ colluders and $N$ random user keys.} Full-registry matching at nominal family-wise false-positive budget
$\delta=0.01$. The complete $N\times K$ results are in \Cref{tab:registry_full}.}
\label{tab:registry_main}
\small
\begin{tabular}{@{}rcccccc@{}}
\toprule
$N$ & Recall & Precision & All-colluder & No-FA & Exact set & FWER \\
\midrule
$10^{3}$ & 1.000 & .999 & .999 & .990 & .989 & .010 \\
$10^{4}$ & 1.000 & .999 & .998 & .990 & .988 & .010 \\
$10^{5}$ & .999 & .999 & .996 & .988 & .984 & .012 \\
$10^{6}$ & .998 & .999 & .990 & .989 & .980 & .011 \\
\bottomrule
\end{tabular}
\end{table}

\subsection{Controlled Comparisons and Ablations}
\label{subsec:ablation}

\begin{table}[tb]
\centering
\small
\caption{\textbf{Ablations on collusion training and key dimension.}
Short training runs evaluate the effect of the collusion objective and the
embedding dimension $D$. Robustness columns report single-key TPR, and
collusion columns report strict all-colluder TPR under averaging attacks at
per-key $\mathrm{FPR}=10^{-6}$. Higher-dimensional key spaces improve the
separation between colluder and null cosine scores, while collusion training
improves large-$K$ attribution.}
\label{tab:main_ablation_studies}
\begin{tabular}{@{}cc c ccc ccccc@{}}
  \toprule
  Dim.
    & $K^{\mathrm{train}}$
    & PSNR
    & Clean
    & Photo.
    & Geo.
    & $K{=}2$
    & $K{=}5$
    & $K{=}10$
    & $K{=}15$
    & $K{=}20$ \\
  \midrule
  1024 & 20 & 32.97 & 1.00 & 1.00 & .901 & \textbf{1.00} & \textbf{1.00} & \textbf{1.00} & \textbf{.977} & \textbf{.650} \\
  1024 & --   & 37.94 & 1.00 & 1.00 & .926 & 1.00 & 1.00 & 1.00 & .927 & .234 \\
  \midrule
   100 & 20 & 32.38 & 1.00 & .991 & .969 & 1.00 & .008 & .000 & .000 & .000 \\
   256 & 20 & 31.89 & 1.00 & 1.00 & .989 & 1.00 & .985 & .007 & .000 & .000 \\
   576 & 20 & 32.43 & 1.00 & 1.00 & .995 & 1.00 & 1.00 & .988 & .321 & .001 \\
  1024 & 20 & 32.97 & 1.00 & 1.00 & .901 & 1.00 & 1.00 & 1.00 & .977 & .650 \\
  \bottomrule
\end{tabular}%
\end{table}

\paragraph{Effect of collusion training.}
The first two rows of \Cref{tab:main_ablation_studies} compare short runs with
and without the collusion objective. Even without explicit collusion training,
the model retains non-trivial averaging robustness, achieving perfect strict
all-colluder TPR through $K=10$. However, performance drops sharply at
$K=20$. Adding the collusion objective substantially improves large-$K$
attribution, increasing $K=20$ TPR from 0.234 to 0.650. This suggests that
high-dimensional near-orthogonal keys already provide a useful inductive bias,
while explicit linearity training is needed to approach the ideal averaging
geometry at larger coalition sizes.

\paragraph{Effect of embedding dimension.}
The remaining rows examine the role of key dimension $D$. The results follow
the predicted scaling from the shifted-Beta null model: larger $D$ concentrates
null cosine scores more tightly around zero, lowers the detection threshold,
and supports larger colluder sets. At $D=100$, strict all-colluder detection
fails before $K=5$; at $D=256$, it remains strong through $K=5$ but collapses
by $K=10$; at $D=576$, it remains high through $K=10$; and at $D=1024$, it
remains effective through $K=20$. The gap between the theoretical limit and
empirical performance reflects non-ideal encoder/extractor linearity, image
distortions, and finite training capacity.

\begin{table}[tbp]
\centering
\caption{\textbf{Dimension and scoring controls.} Exact-set detection of $K$ colluders in database with
$N=10^6$ keys and nominal family-wise false-positive budget $\delta=0.01$. Paired rows use the same marked outputs, changing only the hard or soft matcher.}
\label{tab:controls_main}
\small
\setlength{\tabcolsep}{4pt}
\begin{tabular}{@{}lr l ccr@{}}
\toprule
Method & $D$ & Score & $K=5$ & $K=10$ & PSNR \\
\midrule
\textbf{VectorMark} & 1024 & Soft cosine & \textbf{.992} & \textbf{.974} & 40.28 \\
VectorMark & 1024 & Hard agreement & .991 & .624 & 40.28 \\
\midrule
ChunkySeal & 1024 & Logit cosine & .949 & .441 & 44.70 \\
ChunkySeal & 1024 & Hard agreement & .923 & .002 & 44.70 \\
\midrule
SS-DFT & 1024 & Soft cosine & .862 & .008 & 33.85 \\
DWT-DCT & 1024 & Hard agreement & .490 & .036 & 36.79 \\
PixelSeal & 256 & Logit cosine & .433 & .000 & 46.13 \\
VideoSeal & 256 & Logit cosine & .091 & .000 & 45.07 \\
\bottomrule
\end{tabular}
\end{table}

\paragraph{Soft evidence versus hard decisions.}
We use analytical thresholds for VectorMark and separately calibrate the baseline
scoring rules in a registry containing $N=10^6$ identities, nominal family-wise false-positive budget $\delta=0.01$.
On {identical VectorMark outputs}, detection of colluders using soft cosine similarity gives 0.974 exact-set
recovery at $K=10$, versus 0.624 after hard codeword decisions
(\Cref{tab:controls_main}). ChunkySeal improves from 0.002 to 0.441 by retaining its logits for matching. The DWT-DCT sweep increases $K=5$ exact-set recovery from 0.003
at 256 bits to 0.490 at 1024 bits (\Cref{sec:supp_controls}). Thus dimension,
soft matching, and collusion-resistant training jointly contribute to the performance. 

\subsection{Scheme-Aware and Black-Box Attacks}
\label{subsec:attacks}

\begin{table}[tbp]
\centering
\caption{\textbf{VectorMark under scheme-aware and black-box attacks.} Collusion detection in registry with
$N=10^6$ identities, $K$ colluders and nominal family-wise false-positive budget $\delta=0.01$. PSNR of the attacked image is measured against the unmarked host. The stronger attacks can erase the watermark signal while also perturbing the image significantly.}
\label{tab:attacks_main}
\small
\setlength{\tabcolsep}{3.5pt}
\begin{tabular}{@{}lrcccc@{}}
\toprule
Attack & $K$ & Exact set & Recall & Precision & PSNR \\
\midrule
Mean averaging & 10 & .979 & .998 & .999 & 41.03 \\
\midrule
Subspace removal, $\gamma=0.25$ & 10 & .704 & .945 & .999 & 37.73 \\
Subspace removal, $\gamma=0.50$ & 10 & .083 & .448 & .878 & 34.03 \\
Subspace removal, $\gamma=1$ & 10 & .000 & .012 & .058 & 28.99 \\
\midrule
WmForger & 5 & .725 & .838 & .904 & 32.68 \\
Regen-VAE & 5 & .980 & .998 & .998 & 36.81 \\
SD1.5 img2img & 5 & .000 & .000 & .000 & 25.50 \\
\bottomrule
\end{tabular}
\end{table}

The scheme-aware attacker knows the construction and owns aligned marked
copies, but has no clean host, secret keys, target-model parameters, or
detector outputs. Our scheme-adaptive subspace removal attack subtracts estimated patchwise residual directions from the averaged colluded image.
We also apply black-box attacks such as WmForger~\cite{soucek2025forging}, Regen-VAE~\cite{zhao2024removable}, and SD1.5
img2img denoising~\cite{rombach2022high}. The full six-method results and constructions
are in \Cref{sec:supp_attacks}.
\Cref{tab:attacks_main} shows the results of VectorMark against the attacks for $K$ colluders detection in a database of $N=10^6$ keys. Moderate subspace
removal retains 0.704 exact-set recovery, and WmForger retains 0.725. SD1.5 denoising also removes attribution at the cost of significant image perturbation with 25.50\,dB PSNR. Thus, the stronger attack can erase the watermark signal while also substantially reducing fidelity to the original image.

\section{Conclusion}
\label{sec:conclusion}

VectorMark combines near-orthogonal vector fingerprints, a neural watermark encoder and extractor, and calibrated directional soft retrieval. It achieves strong photometric and geometric robustness with the state-of-the-art detection under collusion-based averaging, including 98.0\% exact-set recovery for ten colluders in a million-identity registry at a nominal 1\% family-wise FPR budget. Controlled ablations attribute the performance gains to high vector dimensionality, soft evidence matching, and collusion-aware training.

\paragraph*{Acknowledgments.}
This work is supported by the ERC starting grant no.~101160648 (Harmony).

{
\small
\bibliographystyle{unsrt}
\bibliography{main}

@String(IJCV = {International Journal of Computer Vision})

@String(CVPR= {Conference on Computer Vision and Pattern Recognition})

@String(ICCV= {Proceedings of the IEEE/CVF International Conference on Computer Vision})

@String(ECCV= {European Conference on Computer Vision})

@String(NIPS= {Advances in Neural Information Processing Systems})

@String(TIP  = {IEEE Transactions on Image Processing})

@String(ACMMM= {ACM International Conference on Multimedia})

@String(ICLR = {International Conference on Learning Representations})

@String(TIT    = {IEEE Transactions on Information Theory})

@String(MICCAI = {International Conference on Medical Image Computing and Computer Assisted Intervention})

@inproceedings{hu2025mask,
  title={Mask Image Watermarking},
  author={Hu, Runyi and Zhang, Jie and Zhao, Shiqian and Lukas, Nils and Li, Jiwei and Guo, Qing and Qiu, Han and Zhang, Tianwei},
  booktitle=NIPS,
  year={2025}
}

@article{wu2017enhanced,
  title={Enhanced Just Noticeable Difference Model for Images With Pattern Complexity},
  author={Wu, Jinjian and Li, Leida and Dong, Weisheng and Shi, Guangming and Lin, Weisi and Kuo, C-C Jay},
  journal=TIP,
  volume={26},
  number={6},
  pages={2682--2693},
  year={2017},
  publisher={IEEE Press Piscataway, NJ, USA}
}

@article{lin2001rotation,
  title={Rotation, scale, and translation resilient watermarking for images},
  author={Lin, C-Y and Wu, Min and Bloom, Jeffrey A and Cox, Ingemar J and Miller, Matthew L and Lui, Yui Man},
  journal=TIP,
  volume={10},
  number={5},
  pages={767--782},
  year={2001},
  publisher={IEEE}
}

@inproceedings{zhu2018hidden,
  title={Hidden: Hiding data with deep networks},
  author={Zhu, Jiren and Kaplan, Russell and Johnson, Justin and Fei-Fei, Li},
  booktitle=ECCV,
  pages={657--672},
  year={2018}
}

@inproceedings{jia2021mbrs,
  title={Mbrs: Enhancing robustness of dnn-based watermarking by mini-batch of real and simulated jpeg compression},
  author={Jia, Zhaoyang and Fang, Han and Zhang, Weiming},
  booktitle=ACMMM,
  pages={41--49},
  year={2021}
}

@article{zhang2019robust,
  title={Robust invisible video watermarking with attention},
  author={Zhang, Kevin Alex and Xu, Lei and Cuesta-Infante, Alfredo and Veeramachaneni, Kalyan},
  journal={arXiv preprint arXiv:1909.01285},
  year={2019}
}

@inproceedings{bui2025trustmark,
  title={Trustmark: Robust watermarking and watermark removal for arbitrary resolution images},
  author={Bui, Tu and Agarwal, Shruti and Collomosse, John},
  booktitle=ICCV,
  pages={18629--18639},
  year={2025}
}

@inproceedings{sander2025watermark,
  title={Watermark anything with localized messages},
  author={Sander, Tom and Fernandez, Pierre and Durmus, Alain and Furon, Teddy and Douze, Matthijs},
  booktitle=ICLR,
  year={2025}
}

@inproceedings{lu2025robust,
  title={Robust Watermarking Using Generative Priors Against Image Editing: From Benchmarking to Advances},
  author={Lu, Shilin and Zhou, Zihan and Lu, Jiayou and Zhu, Yuanzhi and Kong, Adams Wai-Kin},
  booktitle=ICLR,
  year={2025}
}

@article{boneh1998collusion,
  title={Collusion-secure fingerprinting for digital data},
  author={Boneh, Dan and Shaw, James},
  journal=TIT,
  volume={44},
  number={5},
  pages={1897--1905},
  year={1998},
}

@article{tardos2008optimal,
  title={Optimal probabilistic fingerprint codes},
  author={Tardos, G{\'a}bor},
  journal={Journal of the ACM},
  volume={55},
  number={2},
  pages={1--24},
  year={2008},
}

@book{cox2008digital,
  title={Digital watermarking},
  author={Cox, Ingemar J and Miller, Matthew L and Bloom, Jeffrey A and Fridrich, Jessica and Kalker, Ton},
  year={2008},
  publisher={Morgan Kaufmann}
}

@inproceedings{ronneberger2015unet,
  title={U-net: Convolutional networks for biomedical image segmentation},
  author={Ronneberger, Olaf and Fischer, Philipp and Brox, Thomas},
  booktitle=MICCAI,
  pages={234--241},
  year={2015},
}

@inproceedings{loshchilov2019decoupled,
  title={Decoupled Weight Decay Regularization},
  author={Loshchilov, Ilya and Hutter, Frank},
  booktitle=ICLR,
  year={2019}
}

@article{fossorier1995soft,
  title={Soft-decision decoding of linear block codes based on ordered statistics},
  author={Fossorier, Marc PC and Lin, Shu},
  journal=TIT,
  volume={41},
  number={5},
  pages={1379--1396},
  year={1995},
}

@inproceedings{lin2014microsoft,
  title={Microsoft coco: Common objects in context},
  author={Lin, Tsung-Yi and Maire, Michael and Belongie, Serge and Hays, James and Perona, Pietro and Ramanan, Deva and Doll{\'a}r, Piotr and Zitnick, C Lawrence},
  booktitle=ECCV,
  pages={740--755},
  year={2014},
  organization={Springer}
}

@article{cox1997secure,
  title={Secure spread spectrum watermarking for multimedia},
  author={Cox, Ingemar J and Kilian, Joe and Leighton, F Thomson and Shamoon, Talal},
  journal=TIP,
  volume={6},
  number={12},
  pages={1673--1687},
  year={1997},
  publisher={IEEE}
}

@inproceedings{tancik2020stegastamp,
  title={Stegastamp: Invisible hyperlinks in physical photographs},
  author={Tancik, Matthew and Mildenhall, Ben and Ng, Ren},
  booktitle=CVPR,
  pages={2117--2126},
  year={2020}
}

@book{muirhead1982aspects,
  title={Aspects of multivariate statistical theory},
  author={Muirhead, RJ},
  publisher={John Wiley \& Sons},
  year={1982},
}

@article{hoeffding1963probability,
  title={Probability inequalities for sums of bounded random variables},
  author={Hoeffding, Wassily},
  journal={Journal of the American statistical association},
  volume={58},
  number={301},
  pages={13--30},
  year={1963},
  publisher={Taylor \& Francis}
}

@inproceedings{rombach2022high,
  title={High-resolution image synthesis with latent diffusion models},
  author={Rombach, Robin and Blattmann, Andreas and Lorenz, Dominik and Esser, Patrick and Ommer, Bj{\"o}rn},
  booktitle=CVPR,
  pages={10684--10695},
  year={2022}
}

@article{gowal2025synthid,
  title={SynthID-Image: Image watermarking at internet scale},
  author={Gowal, Sven and Bunel, Rudy and Stimberg, Florian and Stutz, David and Ortiz-Jimenez, Guillermo and Kouridi, Christina and Vecerik, Mel and Hayes, Jamie and Rebuffi, Sylvestre-Alvise and Bernard, Paul and others},
  journal={arXiv preprint arXiv:2510.09263},
  year={2025}
}

@misc{shieldmnt2020invisible,
  author       = {{ShieldMnt}},
  title        = {invisible-watermark: {A} {Python} library for invisible image watermarks},
  year         = {2020},
  howpublished = {\url{https://github.com/ShieldMnt/invisible-watermark}},
  note         = {Github repository}
}

@article{fernandez2024videoseal,
  title={Video Seal: Open and Efficient Video Watermarking},
  author={Fernandez, Pierre and Elsahar, Hady and Yalniz, I. Zeki and Mourachko, Alexandre},
  journal={arXiv preprint arXiv:2412.09492},
  year={2024}
}

@article{soucek2025pixelseal,
  title={Pixel Seal: Adversarial-only Training for Invisible Image and Video Watermarking},
  author={Sou\v{c}ek, Tom\'{a}\v{s} and Fernandez, Pierre and Elsahar, Hady and Rebuffi, Sylvestre-Alvise and Lacatusu, Valeriu and Tran, Tuan and Sander, Tom and Mourachko, Alexandre},
  journal={arXiv preprint arXiv:2512.16874},
  year={2025}
}

@article{petrov2025chunkyseal,
  title={We can hide more bits: The unused watermarking capacity in theory and in practice},
  author={Petrov, Aleksandar and Fernandez, Pierre and Sou{\v{c}}ek, Tom{\'a}{\v{s}} and Elsahar, Hady},
  journal={arXiv preprint arXiv:2510.12812},
  year={2025}
}

@inproceedings{soucek2025forging,
  title={Transferable Black-Box One-Shot Forging of Watermarks via Image Preference Models},
  author={Sou\v{c}ek, Tom\'{a}\v{s} and Rebuffi, Sylvestre-Alvise and Fernandez, Pierre and Jovanović, Nikola and Elsahar, Hady and Lacatusu, Valeriu and Tran, Tuan and Mourachko, Alexandre},
  booktitle=NIPS,
  year={2025}
}

@article{kuznetsova2020openimages,
  title={The open images dataset v4: Unified image classification, object detection, and visual relationship detection at scale},
  author={Kuznetsova, Alina and Rom, Hassan and Alldrin, Neil and Uijlings, Jasper and Krasin, Ivan and Pont-Tuset, Jordi and Kamali, Shahab and Popov, Stefan and Malloci, Matteo and Kolesnikov, Alexander and others},
  journal=IJCV,
  volume={128},
  number={7},
  pages={1956--1981},
  year={2020},
  publisher={Springer}
}

@inproceedings{navas2008dwt,
  title={Dwt-dct-svd based watermarking},
  author={Navas, KA and Ajay, Mathews Cheriyan and Lekshmi, M and Archana, Tampy S and Sasikumar, M},
  booktitle={International Conference on Communication Systems Software and Middleware and Workshops},
  pages={271--274},
  year={2008},
  organization={IEEE}
}

@article{al2007combined,
  title={Combined DWT-DCT digital image watermarking},
  author={Al-Haj, Ali},
  journal={Journal of Computer Science},
  volume={3},
  number={9},
  pages={740--746},
  year={2007}
}

@article{christodoulou2024finding,
  title={Finding the subjective truth: Collecting 2 million votes for comprehensive gen-ai model evaluation},
  author={Christodoulou, Dimitrios and Kuhlmann-J{\o}rgensen, Mads},
  journal={arXiv preprint arXiv:2409.11904},
  year={2024}
}

@misc{rapidata_flux_sd3_mj_dalle3_hf,
  author={{Rapidata}},
  title={{Rapidata Image Generation Preference Dataset}},
  year={2024},
  howpublished={\url{https://huggingface.co/datasets/Rapidata/700k_Human_Preference_Dataset_FLUX_SD3_MJ_DALLE3}},
}

@misc{rapidata_openai4o_t2i_human_preference_hf,
  author={{Rapidata}},
  title={{Rapidata OpenAI 4o Preference}},
  year={2025},
  howpublished={\url{https://huggingface.co/datasets/Rapidata/OpenAI-4o_t2i_human_preference}},
}

@misc{bitmind_nano_banana_hf,
  author={{BitMind}},
  title={{Nano-Banana Generated Images}},
  year={2025},
  howpublished={\url{https://huggingface.co/datasets/bitmind/nano-banana}},
}

@misc{ash12321_sdxl_generated_10k_hf,
  author={{ash12321}},
  title={{SDXL Generated Images Dataset (10,000 images)}},
  year={2025},
  howpublished={\url{https://huggingface.co/datasets/ash12321/sdxl-generated-10k}},
}

@article{zhao2024removable,
  title={Invisible image watermarks are provably removable using generative ai},
  author={Zhao, Xuandong and Zhang, Kexun and Su, Zihao and Vasan, Saastha and Grishchenko, Ilya and Kruegel, Christopher and Vigna, Giovanni and Wang, Yu-Xiang and Li, Lei},
  journal=NIPS,
  volume={37},
  pages={8643--8672},
  year={2024}
}

@article{balle2018variational,
  title={Variational image compression with a scale hyperprior},
  author={Ball{\'e}, Johannes and Minnen, David and Singh, Saurabh and Hwang, Sung Jin and Johnston, Nick},
  journal={arXiv preprint arXiv:1802.01436},
  year={2018}
}
}
\appendix
\clearpage

\setcounter{section}{0}
\setcounter{figure}{0}
\setcounter{table}{0}
\setcounter{equation}{0}

\renewcommand{\thesection}{\Alph{section}}
\renewcommand{\thefigure}{S\arabic{figure}}
\renewcommand{\thetable}{S\arabic{table}}
\renewcommand{\theequation}{S\arabic{equation}}
\renewcommand{\theHfigure}{supp.\arabic{figure}}
\renewcommand{\theHtable}{supp.\arabic{table}}
\renewcommand{\theHequation}{supp.\arabic{equation}}

\section{Theoretical Results}
\label{sec:supp_theory_rewrite}

The main paper uses high-dimensional geometry as a detection model. The results
below concern the null distribution, the random-code key ensemble, and the
ideal averaging-collusion model.

\begin{proposition}[Spherical cosine null]
\label{prop:supp_beta_null}
Let $\vv$ be a fixed unit vector in $\R^D$ and let $\vw$ be uniformly
distributed on $\mathbb{S}^{D-1}$. Then
$S=\langle \vv,\vw\rangle$ has density proportional to
$(1-S^2)^{(D-3)/2}$ on $[-1,1]$, equivalently
\begin{equation}
  \frac{1+S}{2}
  \sim
  \mathrm{Beta}\!\left(\frac{D-1}{2},\frac{D-1}{2}\right).
\end{equation}
\end{proposition}

\begin{proof}
By rotational invariance, assume $\vv=e_1$. Then $S$ is the first coordinate
of a uniformly random point on $\mathbb{S}^{D-1}$. The slice with first
coordinate $S$ is a $(D-2)$-sphere of radius $\sqrt{1-S^2}$, and the
surface-to-axis Jacobian gives density proportional to
$(1-S^2)^{(D-3)/2}$. The change of variables $x=(1+S)/2$ gives the stated
Beta distribution.
\end{proof}

\begin{proposition}[Near-orthogonality of random linear-code keys]
\label{prop:supp_rlc}
Let $G\in GF(2)^{D\times L}$ have independent fair bits and let distinct
messages $m_1,m_2$ be mapped to bipolar unit vectors
\begin{equation}
  \vv(m)=\frac{2(Gm\bmod 2)-\mathbf{1}}{\sqrt{D}}.
\end{equation}
For fixed $m_1\neq m_2$,
$\mathbb{E}\langle \vv(m_1),\vv(m_2)\rangle=0$, and
\begin{equation}
  \Pr\!\left(|\langle \vv(m_1),\vv(m_2)\rangle|\geq \epsilon\right)
  \leq 2\exp\!\left(-\frac{D\epsilon^2}{2}\right).
\end{equation}
For a registry of $N$ sampled messages, a union bound gives
\begin{equation}
  \Pr\!\left(\max_{i<j}|\langle \vv_i,\vv_j\rangle|\geq \epsilon\right)
  \leq N(N-1)\exp\!\left(-\frac{D\epsilon^2}{2}\right).
\end{equation}
\end{proposition}

\begin{proof}
For $\Delta=m_1\oplus m_2\neq 0$, each row dot product
$G_i\Delta$ is a fair Bernoulli bit. The bipolar agreement variables are
independent Rademacher variables, so the inner product is
$D^{-1}\sum_{i=1}^{D}X_i$ with $X_i\in\{-1,+1\}$. The mean is zero, and the
tail bound follows from Hoeffding's inequality~\cite{hoeffding1963probability}.
\end{proof}

\begin{proposition}[Ideal equal-weight collusion scaling]
\label{prop:supp_collusion_scaling}
If $\vv_1,\ldots,\vv_K$ are orthonormal and the extractor returns the
normalized key average
$\bar{\vv}=(\sum_i\vv_i)/\|\sum_i\vv_i\|_2$, then every colluder has score
\begin{equation}
  \langle \bar{\vv},\vv_j\rangle=\frac{1}{\sqrt{K}}.
\end{equation}
\end{proposition}

\begin{proof}
Let $\vs=\sum_{i=1}^{K}\vv_i$. Orthonormality gives
$\|\vs\|_2^2=K$ and $\langle \vs,\vv_j\rangle=1$ for every colluder $j$.
Therefore
$\langle \bar{\vv},\vv_j\rangle
=\langle \vs/\|\vs\|_2,\vv_j\rangle=1/\sqrt{K}$.
\end{proof}

\section{Detailed Detection and Payload Recovery Results}
\label{sec:supp_current_results}

This section contains detailed tables underlying the compact main-paper
summaries.

\subsection{Per-noise Detection Results}
\label{subsec:supp_detailed_robustness}

\Cref{tab:supp_tpr_all} reports per-noise detection true positive rates (TPR) at fixed false positive rate (FPR).

\begin{table}[tbp]
\centering
\caption{\textbf{Detection TPR at FPR~$=10^{-6}$ across all 46 distortion conditions.} Cell shading: \colorbox{green!18}{\strut$\geq$0.9}, \colorbox{yellow!30}{\strut0.3--0.9}, \colorbox{red!15}{\strut$<$0.3}. VectorMark and SS-{*} methods use a continuous cosine-similarity null; all other methods use a discrete bit-accuracy null. $^\ddagger$SS-FM: the empirical null deviates significantly from the theoretical Beta null, so the detection thresholds are set from a Beta distribution fitted to unwatermarked scores ($\hat\mu=0.211$, $\hat\sigma=0.062$, $\tau=0.488$).}
\label{tab:supp_tpr_all}
\setlength{\tabcolsep}{2.8pt}
\resizebox{\linewidth}{!}{%
\small
\begin{tabular}{@{}lrrrrrrrrrrrrr@{}}
\toprule
Distortion & \textbf{Ours} & SS-FM$^\ddagger$ & SS-DCT & SS-DFT & \makecell{Mask-\\WM} & \makecell{Stega-\\Stamp} & VINE & \makecell{Trust-\\Mark} & MBRS & \makecell{Riva-\\GAN} & \makecell{DWT\\DCT\\SVD} & \makecell{DWT\\DCT} & HiDDeN \\
\midrule
Clean & \cellcolor{green!18}1.000 & \cellcolor{yellow!30}0.813 & \cellcolor{green!18}1.000 & \cellcolor{green!18}1.000 & \cellcolor{green!18}1.000 & \cellcolor{green!18}1.000 & \cellcolor{green!18}1.000 & \cellcolor{green!18}1.000 & \cellcolor{green!18}1.000 & \cellcolor{green!18}0.996 & \cellcolor{green!18}0.996 & \cellcolor{yellow!30}0.716 & \cellcolor{red!15}0.293  \\
\midrule
\textit{Valuemetric (aggregated)} & \cellcolor{green!18}1.000 & \cellcolor{yellow!30}0.680 & \cellcolor{green!18}1.000 & \cellcolor{green!18}0.997 & \cellcolor{green!18}1.000 & \cellcolor{green!18}1.000 & \cellcolor{green!18}1.000 & \cellcolor{green!18}0.978 & \cellcolor{green!18}0.967 & \cellcolor{green!18}0.971 & \cellcolor{yellow!30}0.561 & \cellcolor{yellow!30}0.467 & \cellcolor{yellow!30}0.425  \\
JPEG $q{=}60$ & \cellcolor{green!18}0.999 & \cellcolor{yellow!30}0.774 & \cellcolor{green!18}1.000 & \cellcolor{green!18}1.000 & \cellcolor{green!18}1.000 & \cellcolor{green!18}1.000 & \cellcolor{green!18}1.000 & \cellcolor{green!18}1.000 & \cellcolor{green!18}1.000 & \cellcolor{green!18}0.992 & \cellcolor{green!18}0.997 & \cellcolor{yellow!30}0.746 & \cellcolor{yellow!30}0.459  \\
JPEG $q{=}50$ & \cellcolor{green!18}0.994 & \cellcolor{yellow!30}0.758 & \cellcolor{green!18}1.000 & \cellcolor{green!18}1.000 & \cellcolor{green!18}1.000 & \cellcolor{green!18}1.000 & \cellcolor{green!18}1.000 & \cellcolor{green!18}1.000 & \cellcolor{green!18}1.000 & \cellcolor{green!18}0.992 & \cellcolor{green!18}0.998 & \cellcolor{yellow!30}0.736 & \cellcolor{yellow!30}0.415  \\
\makecell[l]{Gauss.~Blur $\sigma{=}1.0$} & \cellcolor{green!18}1.000 & \cellcolor{yellow!30}0.768 & \cellcolor{green!18}1.000 & \cellcolor{green!18}1.000 & \cellcolor{green!18}1.000 & \cellcolor{green!18}1.000 & \cellcolor{green!18}1.000 & \cellcolor{green!18}1.000 & \cellcolor{green!18}1.000 & \cellcolor{green!18}0.995 & \cellcolor{green!18}0.997 & \cellcolor{yellow!30}0.768 & \cellcolor{yellow!30}0.492  \\
\makecell[l]{Gauss.~Blur $\sigma{=}1.5$} & \cellcolor{green!18}1.000 & \cellcolor{yellow!30}0.724 & \cellcolor{green!18}1.000 & \cellcolor{green!18}0.999 & \cellcolor{green!18}1.000 & \cellcolor{green!18}1.000 & \cellcolor{green!18}1.000 & \cellcolor{green!18}1.000 & \cellcolor{green!18}1.000 & \cellcolor{green!18}0.995 & \cellcolor{green!18}0.997 & \cellcolor{yellow!30}0.669 & \cellcolor{yellow!30}0.462  \\
\makecell[l]{Gauss.~Noise $\sigma{=}0.1$} & \cellcolor{green!18}1.000 & \cellcolor{yellow!30}0.711 & \cellcolor{green!18}1.000 & \cellcolor{green!18}0.999 & \cellcolor{green!18}1.000 & \cellcolor{green!18}1.000 & \cellcolor{green!18}1.000 & \cellcolor{green!18}1.000 & \cellcolor{green!18}1.000 & \cellcolor{green!18}0.989 & \cellcolor{green!18}0.919 & \cellcolor{yellow!30}0.707 & \cellcolor{red!15}0.248  \\
Median Blur $k{=}5$ & \cellcolor{green!18}1.000 & \cellcolor{yellow!30}0.468 & \cellcolor{green!18}1.000 & \cellcolor{green!18}0.998 & \cellcolor{green!18}1.000 & \cellcolor{green!18}1.000 & \cellcolor{green!18}1.000 & \cellcolor{green!18}1.000 & \cellcolor{green!18}1.000 & \cellcolor{green!18}0.992 & \cellcolor{green!18}0.995 & \cellcolor{red!15}0.248 & \cellcolor{yellow!30}0.431  \\
Salt\,\&\,Pepper $p{=}0.1$ & \cellcolor{green!18}1.000 & \cellcolor{red!15}0.194 & \cellcolor{green!18}1.000 & \cellcolor{green!18}0.967 & \cellcolor{green!18}1.000 & \cellcolor{green!18}0.996 & \cellcolor{green!18}0.994 & \cellcolor{yellow!30}0.673 & \cellcolor{yellow!30}0.512 & \cellcolor{green!18}0.931 & \cellcolor{yellow!30}0.589 & \cellcolor{yellow!30}0.691 & \cellcolor{red!15}0.140  \\
Resize ${\times}0.4$ & \cellcolor{green!18}1.000 & \cellcolor{yellow!30}0.572 & \cellcolor{green!18}1.000 & \cellcolor{green!18}0.999 & \cellcolor{green!18}1.000 & \cellcolor{green!18}1.000 & \cellcolor{green!18}1.000 & \cellcolor{green!18}1.000 & \cellcolor{green!18}1.000 & \cellcolor{green!18}0.993 & \cellcolor{green!18}0.997 & \cellcolor{yellow!30}0.724 & \cellcolor{yellow!30}0.448  \\
Brightness $0.5\times$ & \cellcolor{green!18}1.000 & \cellcolor{yellow!30}0.780 & \cellcolor{green!18}1.000 & \cellcolor{green!18}1.000 & \cellcolor{green!18}1.000 & \cellcolor{green!18}1.000 & \cellcolor{green!18}1.000 & \cellcolor{green!18}1.000 & \cellcolor{green!18}1.000 & \cellcolor{yellow!30}0.865 & \cellcolor{red!15}0.020 & \cellcolor{red!15}0.004 & \cellcolor{yellow!30}0.456  \\
Brightness $1.5\times$ & \cellcolor{green!18}1.000 & \cellcolor{yellow!30}0.647 & \cellcolor{green!18}1.000 & \cellcolor{green!18}0.998 & \cellcolor{green!18}1.000 & \cellcolor{green!18}1.000 & \cellcolor{green!18}1.000 & \cellcolor{green!18}0.999 & \cellcolor{green!18}0.997 & \cellcolor{green!18}0.956 & \cellcolor{red!15}0.279 & \cellcolor{yellow!30}0.570 & \cellcolor{yellow!30}0.501  \\
Contrast $0.5\times$ & \cellcolor{green!18}1.000 & \cellcolor{yellow!30}0.805 & \cellcolor{green!18}1.000 & \cellcolor{green!18}1.000 & \cellcolor{green!18}1.000 & \cellcolor{green!18}1.000 & \cellcolor{green!18}1.000 & \cellcolor{green!18}1.000 & \cellcolor{green!18}1.000 & \cellcolor{green!18}0.952 & \cellcolor{red!15}0.045 & \cellcolor{red!15}0.000 & \cellcolor{yellow!30}0.488  \\
Contrast $1.5\times$ & \cellcolor{green!18}1.000 & \cellcolor{yellow!30}0.656 & \cellcolor{green!18}1.000 & \cellcolor{green!18}1.000 & \cellcolor{green!18}1.000 & \cellcolor{green!18}0.999 & \cellcolor{green!18}1.000 & \cellcolor{green!18}1.000 & \cellcolor{green!18}0.999 & \cellcolor{green!18}0.964 & \cellcolor{red!15}0.191 & \cellcolor{yellow!30}0.371 & \cellcolor{yellow!30}0.516  \\
Saturation $0.5\times$ & \cellcolor{green!18}1.000 & \cellcolor{yellow!30}0.785 & \cellcolor{green!18}1.000 & \cellcolor{green!18}1.000 & \cellcolor{green!18}1.000 & \cellcolor{green!18}1.000 & \cellcolor{green!18}1.000 & \cellcolor{green!18}1.000 & \cellcolor{green!18}1.000 & \cellcolor{green!18}0.986 & \cellcolor{red!15}0.040 & \cellcolor{red!15}0.000 & \cellcolor{yellow!30}0.522  \\
Saturation $1.5\times$ & \cellcolor{green!18}1.000 & \cellcolor{yellow!30}0.772 & \cellcolor{green!18}1.000 & \cellcolor{green!18}1.000 & \cellcolor{green!18}1.000 & \cellcolor{green!18}1.000 & \cellcolor{green!18}1.000 & \cellcolor{green!18}1.000 & \cellcolor{green!18}1.000 & \cellcolor{green!18}0.987 & \cellcolor{red!15}0.262 & \cellcolor{yellow!30}0.378 & \cellcolor{yellow!30}0.506  \\
Hue Shift $0.15$ & \cellcolor{green!18}1.000 & \cellcolor{yellow!30}0.779 & \cellcolor{green!18}1.000 & \cellcolor{green!18}1.000 & \cellcolor{green!18}1.000 & \cellcolor{green!18}1.000 & \cellcolor{green!18}1.000 & \cellcolor{green!18}1.000 & \cellcolor{green!18}1.000 & \cellcolor{green!18}0.974 & \cellcolor{red!15}0.091 & \cellcolor{yellow!30}0.393 & \cellcolor{red!15}0.297  \\
\midrule
\textit{Geometric (aggregated)} & \cellcolor{green!18}0.985 & \cellcolor{red!15}0.041 & \cellcolor{red!15}0.000 & \cellcolor{red!15}0.000 & \cellcolor{yellow!30}0.409 & \cellcolor{red!15}0.161 & \cellcolor{red!15}0.000 & \cellcolor{yellow!30}0.349 & \cellcolor{red!15}0.000 & \cellcolor{yellow!30}0.491 & \cellcolor{red!15}0.005 & \cellcolor{red!15}0.003 & \cellcolor{red!15}0.048  \\
Horiz.~Flip & \cellcolor{green!18}1.000 & \cellcolor{yellow!30}0.813 & \cellcolor{red!15}0.000 & \cellcolor{red!15}0.000 & \cellcolor{green!18}1.000 & \cellcolor{red!15}0.000 & \cellcolor{red!15}0.000 & \cellcolor{green!18}1.000 & \cellcolor{red!15}0.000 & \cellcolor{red!15}0.000 & \cellcolor{red!15}0.001 & \cellcolor{red!15}0.001 & \cellcolor{red!15}0.001  \\
\makecell[l]{Rotate $5^\circ$ (grey)} & \cellcolor{green!18}1.000 & \cellcolor{red!15}0.039 & \cellcolor{red!15}0.000 & \cellcolor{red!15}0.000 & \cellcolor{green!18}1.000 & \cellcolor{green!18}1.000 & \cellcolor{red!15}0.000 & \cellcolor{yellow!30}0.824 & \cellcolor{red!15}0.000 & \cellcolor{green!18}0.985 & \cellcolor{red!15}0.003 & \cellcolor{red!15}0.000 & \cellcolor{red!15}0.158  \\
\makecell[l]{Rotate $10^\circ$ (grey)} & \cellcolor{green!18}1.000 & \cellcolor{red!15}0.049 & \cellcolor{red!15}0.000 & \cellcolor{red!15}0.000 & \cellcolor{green!18}1.000 & \cellcolor{yellow!30}0.475 & \cellcolor{red!15}0.000 & \cellcolor{red!15}0.000 & \cellcolor{red!15}0.000 & \cellcolor{green!18}0.927 & \cellcolor{red!15}0.000 & \cellcolor{red!15}0.000 & \cellcolor{red!15}0.045  \\
\makecell[l]{Rotate $15^\circ$ (grey)} & \cellcolor{green!18}1.000 & \cellcolor{red!15}0.050 & \cellcolor{red!15}0.000 & \cellcolor{red!15}0.000 & \cellcolor{green!18}1.000 & \cellcolor{red!15}0.000 & \cellcolor{red!15}0.000 & \cellcolor{red!15}0.000 & \cellcolor{red!15}0.000 & \cellcolor{yellow!30}0.665 & \cellcolor{red!15}0.000 & \cellcolor{red!15}0.000 & \cellcolor{red!15}0.005  \\
\makecell[l]{Rotate $10^\circ$ (crop)} & \cellcolor{green!18}0.999 & \cellcolor{red!15}0.000 & \cellcolor{red!15}0.000 & \cellcolor{red!15}0.000 & \cellcolor{green!18}0.938 & \cellcolor{red!15}0.000 & \cellcolor{red!15}0.000 & \cellcolor{red!15}0.000 & \cellcolor{red!15}0.000 & \cellcolor{yellow!30}0.849 & \cellcolor{red!15}0.001 & \cellcolor{red!15}0.000 & \cellcolor{red!15}0.034  \\
\makecell[l]{Rotate $10^\circ$ (r.crop)} & \cellcolor{green!18}0.999 & \cellcolor{red!15}0.000 & \cellcolor{red!15}0.000 & \cellcolor{red!15}0.000 & \cellcolor{green!18}0.969 & \cellcolor{red!15}0.001 & \cellcolor{red!15}0.000 & \cellcolor{red!15}0.000 & \cellcolor{red!15}0.000 & \cellcolor{yellow!30}0.858 & \cellcolor{red!15}0.000 & \cellcolor{red!15}0.000 & \cellcolor{red!15}0.040  \\
\makecell[l]{Rotate $15^\circ$ (mirror)} & \cellcolor{green!18}1.000 & \cellcolor{red!15}0.044 & \cellcolor{red!15}0.000 & \cellcolor{red!15}0.000 & \cellcolor{green!18}1.000 & \cellcolor{red!15}0.000 & \cellcolor{red!15}0.000 & \cellcolor{red!15}0.000 & \cellcolor{red!15}0.000 & \cellcolor{yellow!30}0.662 & \cellcolor{red!15}0.000 & \cellcolor{red!15}0.000 & \cellcolor{red!15}0.003  \\
\makecell[l]{Rotate $15^\circ$ (crop)} & \cellcolor{green!18}0.995 & \cellcolor{red!15}0.000 & \cellcolor{red!15}0.000 & \cellcolor{red!15}0.000 & \cellcolor{red!15}0.275 & \cellcolor{red!15}0.000 & \cellcolor{red!15}0.000 & \cellcolor{red!15}0.000 & \cellcolor{red!15}0.000 & \cellcolor{yellow!30}0.389 & \cellcolor{red!15}0.000 & \cellcolor{red!15}0.000 & \cellcolor{red!15}0.003  \\
\makecell[l]{Rotate $15^\circ$ (r.crop)} & \cellcolor{green!18}0.995 & \cellcolor{red!15}0.005 & \cellcolor{red!15}0.000 & \cellcolor{red!15}0.000 & \cellcolor{red!15}0.161 & \cellcolor{red!15}0.000 & \cellcolor{red!15}0.000 & \cellcolor{red!15}0.000 & \cellcolor{red!15}0.000 & \cellcolor{yellow!30}0.393 & \cellcolor{red!15}0.000 & \cellcolor{red!15}0.000 & \cellcolor{red!15}0.004  \\
\makecell[l]{Flip+Rot.$5^\circ$ (grey)} & \cellcolor{green!18}0.999 & \cellcolor{red!15}0.032 & \cellcolor{red!15}0.000 & \cellcolor{red!15}0.000 & \cellcolor{red!15}0.000 & \cellcolor{red!15}0.000 & \cellcolor{red!15}0.000 & \cellcolor{yellow!30}0.800 & \cellcolor{red!15}0.000 & \cellcolor{red!15}0.000 & \cellcolor{red!15}0.000 & \cellcolor{red!15}0.000 & \cellcolor{red!15}0.000  \\
\makecell[l]{ Flip+Rot.$10^\circ$ (grey)} & \cellcolor{green!18}0.996 & \cellcolor{red!15}0.055 & \cellcolor{red!15}0.000 & \cellcolor{red!15}0.000 & \cellcolor{red!15}0.000 & \cellcolor{red!15}0.000 & \cellcolor{red!15}0.000 & \cellcolor{red!15}0.000 & \cellcolor{red!15}0.000 & \cellcolor{red!15}0.000 & \cellcolor{red!15}0.000 & \cellcolor{red!15}0.000 & \cellcolor{red!15}0.000  \\
\makecell[l]{ Flip+Rot.$15^\circ$ (grey)} & \cellcolor{green!18}0.986 & \cellcolor{red!15}0.056 & \cellcolor{red!15}0.000 & \cellcolor{red!15}0.000 & \cellcolor{red!15}0.000 & \cellcolor{red!15}0.000 & \cellcolor{red!15}0.000 & \cellcolor{red!15}0.000 & \cellcolor{red!15}0.000 & \cellcolor{red!15}0.000 & \cellcolor{red!15}0.000 & \cellcolor{red!15}0.000 & \cellcolor{red!15}0.000  \\
\makecell[l]{ Flip+Rot.$10^\circ$ (crop)} & \cellcolor{green!18}0.973 & \cellcolor{red!15}0.000 & \cellcolor{red!15}0.000 & \cellcolor{red!15}0.000 & \cellcolor{red!15}0.000 & \cellcolor{red!15}0.000 & \cellcolor{red!15}0.000 & \cellcolor{red!15}0.000 & \cellcolor{red!15}0.000 & \cellcolor{red!15}0.000 & \cellcolor{red!15}0.000 & \cellcolor{red!15}0.000 & \cellcolor{red!15}0.000  \\
\makecell[l]{ Flip+Rot.$10^\circ$ (r.crop)} & \cellcolor{green!18}0.975 & \cellcolor{red!15}0.000 & \cellcolor{red!15}0.000 & \cellcolor{red!15}0.000 & \cellcolor{red!15}0.000 & \cellcolor{red!15}0.000 & \cellcolor{red!15}0.000 & \cellcolor{red!15}0.000 & \cellcolor{red!15}0.000 & \cellcolor{red!15}0.000 & \cellcolor{red!15}0.000 & \cellcolor{red!15}0.000 & \cellcolor{red!15}0.000  \\
\makecell[l]{ Flip+Rot.$15^\circ$ (mirror)} & \cellcolor{green!18}0.986 & \cellcolor{red!15}0.053 & \cellcolor{red!15}0.000 & \cellcolor{red!15}0.000 & \cellcolor{red!15}0.000 & \cellcolor{red!15}0.000 & \cellcolor{red!15}0.000 & \cellcolor{red!15}0.000 & \cellcolor{red!15}0.000 & \cellcolor{red!15}0.000 & \cellcolor{red!15}0.001 & \cellcolor{red!15}0.000 & \cellcolor{red!15}0.000  \\
\makecell[l]{ Flip+Rot.$15^\circ$ (crop)} & \cellcolor{yellow!30}0.895 & \cellcolor{red!15}0.001 & \cellcolor{red!15}0.000 & \cellcolor{red!15}0.000 & \cellcolor{red!15}0.000 & \cellcolor{red!15}0.000 & \cellcolor{red!15}0.000 & \cellcolor{red!15}0.000 & \cellcolor{red!15}0.000 & \cellcolor{red!15}0.000 & \cellcolor{red!15}0.000 & \cellcolor{red!15}0.000 & \cellcolor{red!15}0.000  \\
\makecell[l]{ Flip+Rot.$15^\circ$ (r.crop)} & \cellcolor{yellow!30}0.817 & \cellcolor{red!15}0.004 & \cellcolor{red!15}0.000 & \cellcolor{red!15}0.000 & \cellcolor{red!15}0.000 & \cellcolor{red!15}0.000 & \cellcolor{red!15}0.000 & \cellcolor{red!15}0.000 & \cellcolor{red!15}0.000 & \cellcolor{red!15}0.000 & \cellcolor{red!15}0.000 & \cellcolor{red!15}0.000 & \cellcolor{red!15}0.000  \\
Persp.~$s{=}0.1$ & \cellcolor{green!18}0.999 & \cellcolor{red!15}0.000 & \cellcolor{red!15}0.000 & \cellcolor{red!15}0.000 & \cellcolor{green!18}1.000 & \cellcolor{green!18}0.997 & \cellcolor{red!15}0.000 & \cellcolor{green!18}1.000 & \cellcolor{red!15}0.000 & \cellcolor{green!18}0.984 & \cellcolor{red!15}0.000 & \cellcolor{red!15}0.001 & \cellcolor{red!15}0.188  \\
Persp.~$s{=}0.2$ & \cellcolor{green!18}1.000 & \cellcolor{red!15}0.000 & \cellcolor{red!15}0.000 & \cellcolor{red!15}0.000 & \cellcolor{yellow!30}0.301 & \cellcolor{yellow!30}0.616 & \cellcolor{red!15}0.000 & \cellcolor{green!18}1.000 & \cellcolor{red!15}0.000 & \cellcolor{green!18}0.968 & \cellcolor{red!15}0.000 & \cellcolor{red!15}0.000 & \cellcolor{red!15}0.102  \\
Persp.~$s{=}0.3$ & \cellcolor{green!18}1.000 & \cellcolor{red!15}0.000 & \cellcolor{red!15}0.000 & \cellcolor{red!15}0.000 & \cellcolor{red!15}0.000 & \cellcolor{red!15}0.000 & \cellcolor{red!15}0.000 & \cellcolor{green!18}0.988 & \cellcolor{red!15}0.000 & \cellcolor{green!18}0.919 & \cellcolor{red!15}0.000 & \cellcolor{red!15}0.000 & \cellcolor{red!15}0.069  \\
\makecell[l]{CropRes. $1.1\times$ (r.crop)} & \cellcolor{green!18}1.000 & \cellcolor{red!15}0.000 & \cellcolor{red!15}0.000 & \cellcolor{red!15}0.000 & \cellcolor{yellow!30}0.499 & \cellcolor{yellow!30}0.595 & \cellcolor{red!15}0.000 & \cellcolor{green!18}1.000 & \cellcolor{red!15}0.000 & \cellcolor{green!18}0.986 & \cellcolor{red!15}0.000 & \cellcolor{red!15}0.000 & \cellcolor{red!15}0.147  \\
\makecell[l]{CropRes. $1.2\times$ (c.crop)} & \cellcolor{green!18}1.000 & \cellcolor{red!15}0.012 & \cellcolor{red!15}0.000 & \cellcolor{red!15}0.000 & \cellcolor{red!15}0.132 & \cellcolor{yellow!30}0.493 & \cellcolor{red!15}0.000 & \cellcolor{green!18}1.000 & \cellcolor{red!15}0.000 & \cellcolor{green!18}0.939 & \cellcolor{red!15}0.000 & \cellcolor{red!15}0.000 & \cellcolor{red!15}0.092  \\
\makecell[l]{CropRes. $1.3\times$ (r.crop)} & \cellcolor{green!18}0.996 & \cellcolor{red!15}0.003 & \cellcolor{red!15}0.000 & \cellcolor{red!15}0.000 & \cellcolor{red!15}0.000 & \cellcolor{red!15}0.000 & \cellcolor{red!15}0.000 & \cellcolor{yellow!30}0.852 & \cellcolor{red!15}0.000 & \cellcolor{yellow!30}0.839 & \cellcolor{red!15}0.000 & \cellcolor{red!15}0.000 & \cellcolor{red!15}0.051  \\
\makecell[l]{CropStr. (wide $1.3\times$)} & \cellcolor{green!18}0.999 & \cellcolor{red!15}0.000 & \cellcolor{red!15}0.001 & \cellcolor{red!15}0.002 & \cellcolor{red!15}0.108 & \cellcolor{yellow!30}0.310 & \cellcolor{red!15}0.000 & \cellcolor{green!18}1.000 & \cellcolor{red!15}0.000 & \cellcolor{green!18}0.952 & \cellcolor{red!15}0.000 & \cellcolor{red!15}0.000 & \cellcolor{red!15}0.200  \\
\makecell[l]{CropStr. (tall $1.3\times$)} & \cellcolor{green!18}1.000 & \cellcolor{red!15}0.000 & \cellcolor{red!15}0.000 & \cellcolor{red!15}0.000 & \cellcolor{red!15}0.004 & \cellcolor{yellow!30}0.350 & \cellcolor{red!15}0.000 & \cellcolor{green!18}1.000 & \cellcolor{red!15}0.000 & \cellcolor{green!18}0.968 & \cellcolor{red!15}0.120 & \cellcolor{red!15}0.079 & \cellcolor{red!15}0.246  \\
\makecell[l]{PadRes. $0.5\times$ (mirror)} & \cellcolor{green!18}0.945 & \cellcolor{red!15}0.000 & \cellcolor{red!15}0.000 & \cellcolor{red!15}0.000 & \cellcolor{red!15}0.000 & \cellcolor{red!15}0.000 & \cellcolor{red!15}0.000 & \cellcolor{red!15}0.000 & \cellcolor{red!15}0.000 & \cellcolor{red!15}0.000 & \cellcolor{red!15}0.000 & \cellcolor{red!15}0.000 & \cellcolor{red!15}0.000  \\
\makecell[l]{PadRes. $0.6\times$ (grey)} & \cellcolor{green!18}1.000 & \cellcolor{red!15}0.008 & \cellcolor{red!15}0.000 & \cellcolor{red!15}0.000 & \cellcolor{green!18}1.000 & \cellcolor{red!15}0.000 & \cellcolor{red!15}0.000 & \cellcolor{red!15}0.000 & \cellcolor{red!15}0.000 & \cellcolor{red!15}0.135 & \cellcolor{red!15}0.000 & \cellcolor{red!15}0.000 & \cellcolor{red!15}0.007  \\
\makecell[l]{PadRes. $0.7\times$ (mean)} & \cellcolor{green!18}1.000 & \cellcolor{red!15}0.000 & \cellcolor{red!15}0.000 & \cellcolor{red!15}0.000 & \cellcolor{green!18}1.000 & \cellcolor{red!15}0.000 & \cellcolor{red!15}0.000 & \cellcolor{red!15}0.000 & \cellcolor{red!15}0.000 & \cellcolor{yellow!30}0.622 & \cellcolor{red!15}0.000 & \cellcolor{red!15}0.000 & \cellcolor{red!15}0.034  \\
\makecell[l]{PadStr. (wide $0.5\times$)} & \cellcolor{green!18}1.000 & \cellcolor{red!15}0.001 & \cellcolor{red!15}0.000 & \cellcolor{red!15}0.001 & \cellcolor{yellow!30}0.898 & \cellcolor{red!15}0.000 & \cellcolor{red!15}0.000 & \cellcolor{red!15}0.000 & \cellcolor{red!15}0.000 & \cellcolor{red!15}0.162 & \cellcolor{red!15}0.000 & \cellcolor{red!15}0.000 & \cellcolor{red!15}0.011  \\
\makecell[l]{PadStr. (tall $0.5\times$)} & \cellcolor{green!18}1.000 & \cellcolor{red!15}0.000 & \cellcolor{red!15}0.000 & \cellcolor{red!15}0.000 & \cellcolor{red!15}0.000 & \cellcolor{red!15}0.000 & \cellcolor{red!15}0.000 & \cellcolor{red!15}0.000 & \cellcolor{red!15}0.000 & \cellcolor{yellow!30}0.542 & \cellcolor{red!15}0.010 & \cellcolor{red!15}0.000 & \cellcolor{red!15}0.007  \\
\bottomrule
\end{tabular}
}
\end{table}

\subsection{Per-noise Payload Recovery Results}
\label{subsec:supp_payload}

\Cref{tab:supp_payload_scores} reports per-noise payload decoding accuracies.
Payload decoding is an
additional single-key capability of VectorMark.

\begin{table}[tbp]
\centering
\caption{\textbf{Payload bit accuracy across all 46 distortion conditions.} All entries are bit accuracies $\hat p$ (green: \colorbox{green!18}{\strut$\geq$0.85}, yellow: \colorbox{yellow!30}{\strut0.65--0.85}, red: \colorbox{red!15}{\strut$<$0.65}). For VectorMark, \emph{raw} denotes 1024-bit codeword sign accuracy, \emph{key} denotes decoded 100-bit payload accuracy after ECC decoding through a random linear code. \emph{2-pass} rows are blind flip-invariant decoding results: the decoder evaluates both $\hat e$ and $\hat e^{\leftrightarrow}$, re-encodes all candidates, and selects by codeword-match score. Baseline columns report native payload bit accuracies.}
\label{tab:supp_payload_scores}
\setlength{\tabcolsep}{2.8pt}
\resizebox{\linewidth}{!}{%
\small
\begin{tabular}{@{}lrrrrrrrrrrrrr@{}}
\toprule
Distortion & \makecell{Ours\\raw} & \makecell{Ours\\key} & \makecell{Ours\\2-pass\\raw} & \makecell{Ours\\2-pass\\key} & \makecell{Mask-\\WM} & \makecell{Stega-\\Stamp} & VINE & \makecell{Trust-\\Mark} & MBRS & \makecell{Riva-\\GAN} & \makecell{DWT\\DCT\\SVD} & \makecell{DWT\\DCT} & HiDDeN \\
\midrule
\textit{score} & \textit{acc} & \textit{acc} & \textit{acc} & \textit{acc} & \textit{acc} & \textit{acc} & \textit{acc} & \textit{acc} & \textit{acc} & \textit{acc} & \textit{acc} & \textit{acc} & \textit{acc} \\
\textit{dim/bits} & $1024^b$ & $100^b$ & $1024^b$ & $100^b$ & $64^b$ & $100^b$ & $100^b$ & $100^b$ & $64^b$ & $32^b$ & $64^b$ & $64^b$ & $30^b$ \\
\midrule
Clean & \cellcolor{green!18}1.000 & \cellcolor{green!18}1.000 & \cellcolor{green!18}1.000 & \cellcolor{green!18}1.000 & \cellcolor{green!18}1.000 & \cellcolor{green!18}0.999 & \cellcolor{green!18}1.000 & \cellcolor{green!18}1.000 & \cellcolor{green!18}1.000 & \cellcolor{green!18}0.997 & \cellcolor{green!18}0.998 & \cellcolor{green!18}0.855 & \cellcolor{green!18}0.884 \\
\midrule
\textit{Valuemetric (aggregated)} & \cellcolor{green!18}0.945 & \cellcolor{green!18}0.956 & \cellcolor{green!18}0.944 & \cellcolor{green!18}0.957 & \cellcolor{green!18}1.000 & \cellcolor{green!18}0.991 & \cellcolor{green!18}0.994 & \cellcolor{green!18}0.981 & \cellcolor{green!18}0.983 & \cellcolor{green!18}0.987 & \cellcolor{yellow!30}0.770 & \cellcolor{yellow!30}0.730 & \cellcolor{green!18}0.866 \\
JPEG $q{=}60$ & \cellcolor{yellow!30}0.789 & \cellcolor{yellow!30}0.832 & \cellcolor{yellow!30}0.788 & \cellcolor{yellow!30}0.834 & \cellcolor{green!18}1.000 & \cellcolor{green!18}0.998 & \cellcolor{green!18}0.999 & \cellcolor{green!18}0.998 & \cellcolor{green!18}1.000 & \cellcolor{green!18}0.996 & \cellcolor{green!18}0.996 & \cellcolor{green!18}0.857 & \cellcolor{green!18}0.872 \\
JPEG $q{=}50$ & \cellcolor{yellow!30}0.720 & \cellcolor{yellow!30}0.655 & \cellcolor{yellow!30}0.716 & \cellcolor{yellow!30}0.658 & \cellcolor{green!18}1.000 & \cellcolor{green!18}0.998 & \cellcolor{green!18}0.999 & \cellcolor{green!18}0.998 & \cellcolor{green!18}1.000 & \cellcolor{green!18}0.996 & \cellcolor{green!18}0.995 & \cellcolor{green!18}0.854 & \cellcolor{green!18}0.868 \\
Gauss.~Blur $\sigma{=}1.0$ & \cellcolor{green!18}0.999 & \cellcolor{green!18}1.000 & \cellcolor{green!18}0.999 & \cellcolor{green!18}1.000 & \cellcolor{green!18}1.000 & \cellcolor{green!18}0.999 & \cellcolor{green!18}0.999 & \cellcolor{green!18}1.000 & \cellcolor{green!18}1.000 & \cellcolor{green!18}0.997 & \cellcolor{green!18}0.997 & \cellcolor{green!18}0.865 & \cellcolor{green!18}0.880 \\
Gauss.~Blur $\sigma{=}1.5$ & \cellcolor{green!18}0.999 & \cellcolor{green!18}1.000 & \cellcolor{green!18}0.999 & \cellcolor{green!18}1.000 & \cellcolor{green!18}1.000 & \cellcolor{green!18}0.999 & \cellcolor{green!18}0.999 & \cellcolor{green!18}1.000 & \cellcolor{green!18}1.000 & \cellcolor{green!18}0.996 & \cellcolor{green!18}0.997 & \cellcolor{yellow!30}0.823 & \cellcolor{green!18}0.875 \\
Gauss.~Noise $\sigma{=}0.1$ & \cellcolor{yellow!30}0.809 & \cellcolor{green!18}0.887 & \cellcolor{yellow!30}0.809 & \cellcolor{green!18}0.888 & \cellcolor{green!18}1.000 & \cellcolor{green!18}0.997 & \cellcolor{green!18}0.999 & \cellcolor{green!18}0.986 & \cellcolor{green!18}0.993 & \cellcolor{green!18}0.994 & \cellcolor{green!18}0.946 & \cellcolor{yellow!30}0.844 & \cellcolor{yellow!30}0.826 \\
Median Blur $k{=}5$ & \cellcolor{green!18}0.961 & \cellcolor{green!18}1.000 & \cellcolor{green!18}0.961 & \cellcolor{green!18}1.000 & \cellcolor{green!18}1.000 & \cellcolor{green!18}0.999 & \cellcolor{green!18}0.987 & \cellcolor{green!18}0.999 & \cellcolor{green!18}0.999 & \cellcolor{green!18}0.995 & \cellcolor{green!18}0.995 & \cellcolor{yellow!30}0.679 & \cellcolor{green!18}0.869 \\
Salt\,\&\,Pepper $p{=}0.1$ & \cellcolor{green!18}0.998 & \cellcolor{green!18}1.000 & \cellcolor{green!18}0.998 & \cellcolor{green!18}1.000 & \cellcolor{green!18}1.000 & \cellcolor{green!18}0.900 & \cellcolor{green!18}0.932 & \cellcolor{yellow!30}0.773 & \cellcolor{yellow!30}0.781 & \cellcolor{green!18}0.966 & \cellcolor{yellow!30}0.744 & \cellcolor{yellow!30}0.841 & \cellcolor{yellow!30}0.808 \\
Resize ${\times}0.4$ & \cellcolor{green!18}0.984 & \cellcolor{green!18}1.000 & \cellcolor{green!18}0.984 & \cellcolor{green!18}1.000 & \cellcolor{green!18}1.000 & \cellcolor{green!18}0.997 & \cellcolor{green!18}0.999 & \cellcolor{green!18}0.999 & \cellcolor{green!18}0.999 & \cellcolor{green!18}0.996 & \cellcolor{green!18}0.997 & \cellcolor{yellow!30}0.849 & \cellcolor{green!18}0.874 \\
Brightness $0.5\times$ & \cellcolor{green!18}1.000 & \cellcolor{green!18}1.000 & \cellcolor{green!18}1.000 & \cellcolor{green!18}1.000 & \cellcolor{green!18}1.000 & \cellcolor{green!18}0.995 & \cellcolor{green!18}0.999 & \cellcolor{green!18}0.995 & \cellcolor{green!18}0.992 & \cellcolor{green!18}0.950 & \cellcolor{red!15}0.477 & \cellcolor{red!15}0.490 & \cellcolor{green!18}0.875 \\
Brightness $1.5\times$ & \cellcolor{green!18}0.945 & \cellcolor{green!18}0.977 & \cellcolor{green!18}0.944 & \cellcolor{green!18}0.978 & \cellcolor{green!18}1.000 & \cellcolor{green!18}0.991 & \cellcolor{green!18}0.999 & \cellcolor{green!18}0.991 & \cellcolor{green!18}0.990 & \cellcolor{green!18}0.983 & \cellcolor{yellow!30}0.650 & \cellcolor{yellow!30}0.776 & \cellcolor{green!18}0.879 \\
Contrast $0.5\times$ & \cellcolor{green!18}1.000 & \cellcolor{green!18}1.000 & \cellcolor{green!18}1.000 & \cellcolor{green!18}1.000 & \cellcolor{green!18}1.000 & \cellcolor{green!18}0.996 & \cellcolor{green!18}0.999 & \cellcolor{green!18}0.998 & \cellcolor{green!18}0.999 & \cellcolor{green!18}0.978 & \cellcolor{red!15}0.487 & \cellcolor{red!15}0.498 & \cellcolor{green!18}0.879 \\
Contrast $1.5\times$ & \cellcolor{green!18}0.971 & \cellcolor{green!18}0.994 & \cellcolor{green!18}0.971 & \cellcolor{green!18}0.994 & \cellcolor{green!18}1.000 & \cellcolor{green!18}0.994 & \cellcolor{green!18}0.999 & \cellcolor{green!18}0.992 & \cellcolor{green!18}0.993 & \cellcolor{green!18}0.986 & \cellcolor{red!15}0.618 & \cellcolor{yellow!30}0.700 & \cellcolor{green!18}0.879 \\
Saturation $0.5\times$ & \cellcolor{green!18}1.000 & \cellcolor{green!18}1.000 & \cellcolor{green!18}1.000 & \cellcolor{green!18}1.000 & \cellcolor{green!18}1.000 & \cellcolor{green!18}0.998 & \cellcolor{green!18}0.999 & \cellcolor{green!18}0.999 & \cellcolor{green!18}1.000 & \cellcolor{green!18}0.993 & \cellcolor{red!15}0.479 & \cellcolor{red!15}0.497 & \cellcolor{green!18}0.882 \\
Saturation $1.5\times$ & \cellcolor{green!18}1.000 & \cellcolor{green!18}1.000 & \cellcolor{green!18}1.000 & \cellcolor{green!18}1.000 & \cellcolor{green!18}1.000 & \cellcolor{green!18}0.998 & \cellcolor{green!18}0.999 & \cellcolor{green!18}0.999 & \cellcolor{green!18}1.000 & \cellcolor{green!18}0.994 & \cellcolor{red!15}0.641 & \cellcolor{yellow!30}0.690 & \cellcolor{green!18}0.881 \\
Hue Shift $0.15$ & \cellcolor{green!18}0.998 & \cellcolor{green!18}1.000 & \cellcolor{green!18}0.998 & \cellcolor{green!18}1.000 & \cellcolor{green!18}1.000 & \cellcolor{green!18}0.998 & \cellcolor{green!18}0.999 & \cellcolor{green!18}0.998 & \cellcolor{green!18}1.000 & \cellcolor{green!18}0.987 & \cellcolor{red!15}0.535 & \cellcolor{yellow!30}0.688 & \cellcolor{green!18}0.850 \\
\midrule
\textit{Geometric (aggregated)} & \cellcolor{yellow!30}0.838 & \cellcolor{yellow!30}0.827 & \cellcolor{green!18}0.921 & \cellcolor{green!18}0.975 & \cellcolor{yellow!30}0.725 & \cellcolor{red!15}0.588 & \cellcolor{red!15}0.500 & \cellcolor{yellow!30}0.668 & \cellcolor{red!15}0.499 & \cellcolor{yellow!30}0.787 & \cellcolor{red!15}0.535 & \cellcolor{red!15}0.519 & \cellcolor{yellow!30}0.686 \\
Horiz.~Flip & \cellcolor{yellow!30}0.700 & \cellcolor{red!15}0.501 & \cellcolor{green!18}1.000 & \cellcolor{green!18}1.000 & \cellcolor{green!18}1.000 & \cellcolor{red!15}0.515 & \cellcolor{red!15}0.497 & \cellcolor{green!18}0.999 & \cellcolor{red!15}0.503 & \cellcolor{red!15}0.503 & \cellcolor{red!15}0.503 & \cellcolor{red!15}0.505 & \cellcolor{red!15}0.563 \\
Rotate $5^\circ$ (grey) & \cellcolor{green!18}0.999 & \cellcolor{green!18}1.000 & \cellcolor{green!18}0.999 & \cellcolor{green!18}1.000 & \cellcolor{green!18}1.000 & \cellcolor{green!18}0.967 & \cellcolor{red!15}0.508 & \cellcolor{yellow!30}0.797 & \cellcolor{red!15}0.501 & \cellcolor{green!18}0.990 & \cellcolor{red!15}0.562 & \cellcolor{red!15}0.536 & \cellcolor{yellow!30}0.818 \\
Rotate $10^\circ$ (grey) & \cellcolor{green!18}0.995 & \cellcolor{green!18}1.000 & \cellcolor{green!18}0.995 & \cellcolor{green!18}1.000 & \cellcolor{green!18}0.999 & \cellcolor{yellow!30}0.741 & \cellcolor{red!15}0.504 & \cellcolor{red!15}0.507 & \cellcolor{red!15}0.501 & \cellcolor{green!18}0.966 & \cellcolor{red!15}0.544 & \cellcolor{red!15}0.522 & \cellcolor{yellow!30}0.769 \\
Rotate $15^\circ$ (grey) & \cellcolor{green!18}0.963 & \cellcolor{green!18}0.999 & \cellcolor{green!18}0.963 & \cellcolor{green!18}0.999 & \cellcolor{green!18}0.996 & \cellcolor{red!15}0.527 & \cellcolor{red!15}0.500 & \cellcolor{red!15}0.506 & \cellcolor{red!15}0.504 & \cellcolor{green!18}0.907 & \cellcolor{red!15}0.526 & \cellcolor{red!15}0.511 & \cellcolor{yellow!30}0.709 \\
Rotate $10^\circ$ (crop) & \cellcolor{green!18}0.955 & \cellcolor{green!18}0.999 & \cellcolor{green!18}0.955 & \cellcolor{green!18}0.999 & \cellcolor{green!18}0.850 & \cellcolor{red!15}0.550 & \cellcolor{red!15}0.498 & \cellcolor{red!15}0.511 & \cellcolor{red!15}0.495 & \cellcolor{green!18}0.945 & \cellcolor{red!15}0.555 & \cellcolor{red!15}0.527 & \cellcolor{yellow!30}0.757 \\
Rotate $10^\circ$ (r.crop) & \cellcolor{green!18}0.951 & \cellcolor{green!18}0.999 & \cellcolor{green!18}0.951 & \cellcolor{green!18}1.000 & \cellcolor{green!18}0.862 & \cellcolor{red!15}0.558 & \cellcolor{red!15}0.501 & \cellcolor{red!15}0.509 & \cellcolor{red!15}0.500 & \cellcolor{green!18}0.947 & \cellcolor{red!15}0.552 & \cellcolor{red!15}0.525 & \cellcolor{yellow!30}0.762 \\
Rotate $15^\circ$ (mirror) & \cellcolor{green!18}0.961 & \cellcolor{green!18}0.998 & \cellcolor{green!18}0.961 & \cellcolor{green!18}0.998 & \cellcolor{green!18}0.989 & \cellcolor{red!15}0.501 & \cellcolor{red!15}0.500 & \cellcolor{red!15}0.504 & \cellcolor{red!15}0.504 & \cellcolor{green!18}0.907 & \cellcolor{red!15}0.539 & \cellcolor{red!15}0.514 & \cellcolor{yellow!30}0.708 \\
Rotate $15^\circ$ (crop) & \cellcolor{green!18}0.850 & \cellcolor{green!18}0.985 & \cellcolor{green!18}0.850 & \cellcolor{green!18}0.986 & \cellcolor{yellow!30}0.740 & \cellcolor{red!15}0.494 & \cellcolor{red!15}0.501 & \cellcolor{red!15}0.501 & \cellcolor{red!15}0.504 & \cellcolor{green!18}0.857 & \cellcolor{red!15}0.537 & \cellcolor{red!15}0.521 & \cellcolor{yellow!30}0.692 \\
Rotate $15^\circ$ (r.crop) & \cellcolor{yellow!30}0.840 & \cellcolor{green!18}0.980 & \cellcolor{yellow!30}0.840 & \cellcolor{green!18}0.982 & \cellcolor{yellow!30}0.716 & \cellcolor{red!15}0.499 & \cellcolor{red!15}0.495 & \cellcolor{red!15}0.503 & \cellcolor{red!15}0.499 & \cellcolor{green!18}0.859 & \cellcolor{red!15}0.533 & \cellcolor{red!15}0.517 & \cellcolor{yellow!30}0.695 \\
Flip+Rot.$5^\circ$ (grey) & \cellcolor{yellow!30}0.689 & \cellcolor{red!15}0.499 & \cellcolor{green!18}0.999 & \cellcolor{green!18}1.000 & \cellcolor{red!15}0.521 & \cellcolor{red!15}0.512 & \cellcolor{red!15}0.504 & \cellcolor{yellow!30}0.798 & \cellcolor{red!15}0.502 & \cellcolor{red!15}0.498 & \cellcolor{red!15}0.513 & \cellcolor{red!15}0.505 & \cellcolor{red!15}0.541 \\
Flip+Rot.$10^\circ$ (grey) & \cellcolor{yellow!30}0.685 & \cellcolor{red!15}0.500 & \cellcolor{green!18}0.995 & \cellcolor{green!18}1.000 & \cellcolor{red!15}0.509 & \cellcolor{red!15}0.511 & \cellcolor{red!15}0.500 & \cellcolor{red!15}0.506 & \cellcolor{red!15}0.503 & \cellcolor{red!15}0.497 & \cellcolor{red!15}0.512 & \cellcolor{red!15}0.505 & \cellcolor{red!15}0.532 \\
Flip+Rot.$15^\circ$ (grey) & \cellcolor{yellow!30}0.672 & \cellcolor{red!15}0.497 & \cellcolor{green!18}0.964 & \cellcolor{green!18}0.999 & \cellcolor{red!15}0.508 & \cellcolor{red!15}0.510 & \cellcolor{red!15}0.496 & \cellcolor{red!15}0.502 & \cellcolor{red!15}0.500 & \cellcolor{red!15}0.502 & \cellcolor{red!15}0.512 & \cellcolor{red!15}0.504 & \cellcolor{red!15}0.525 \\
Flip+Rot.$10^\circ$ (crop) & \cellcolor{yellow!30}0.658 & \cellcolor{red!15}0.503 & \cellcolor{green!18}0.955 & \cellcolor{green!18}0.999 & \cellcolor{red!15}0.510 & \cellcolor{red!15}0.507 & \cellcolor{red!15}0.498 & \cellcolor{red!15}0.509 & \cellcolor{red!15}0.502 & \cellcolor{red!15}0.497 & \cellcolor{red!15}0.502 & \cellcolor{red!15}0.498 & \cellcolor{red!15}0.531 \\
Flip+Rot.$10^\circ$ (r.crop) & \cellcolor{yellow!30}0.658 & \cellcolor{red!15}0.499 & \cellcolor{green!18}0.948 & \cellcolor{green!18}0.999 & \cellcolor{red!15}0.512 & \cellcolor{red!15}0.508 & \cellcolor{red!15}0.499 & \cellcolor{red!15}0.510 & \cellcolor{red!15}0.503 & \cellcolor{red!15}0.497 & \cellcolor{red!15}0.501 & \cellcolor{red!15}0.499 & \cellcolor{red!15}0.528 \\
Flip+Rot.$15^\circ$ (mirror) & \cellcolor{yellow!30}0.670 & \cellcolor{red!15}0.500 & \cellcolor{green!18}0.961 & \cellcolor{green!18}0.998 & \cellcolor{red!15}0.505 & \cellcolor{red!15}0.514 & \cellcolor{red!15}0.494 & \cellcolor{red!15}0.502 & \cellcolor{red!15}0.500 & \cellcolor{red!15}0.505 & \cellcolor{red!15}0.511 & \cellcolor{red!15}0.502 & \cellcolor{red!15}0.528 \\
Flip+Rot.$15^\circ$ (crop) & \cellcolor{red!15}0.628 & \cellcolor{red!15}0.499 & \cellcolor{yellow!30}0.845 & \cellcolor{green!18}0.983 & \cellcolor{red!15}0.502 & \cellcolor{red!15}0.500 & \cellcolor{red!15}0.499 & \cellcolor{red!15}0.501 & \cellcolor{red!15}0.502 & \cellcolor{red!15}0.497 & \cellcolor{red!15}0.505 & \cellcolor{red!15}0.500 & \cellcolor{red!15}0.523 \\
Flip+Rot.$15^\circ$ (r.crop) & \cellcolor{red!15}0.624 & \cellcolor{red!15}0.501 & \cellcolor{yellow!30}0.831 & \cellcolor{green!18}0.975 & \cellcolor{red!15}0.511 & \cellcolor{red!15}0.500 & \cellcolor{red!15}0.502 & \cellcolor{red!15}0.501 & \cellcolor{red!15}0.499 & \cellcolor{red!15}0.497 & \cellcolor{red!15}0.504 & \cellcolor{red!15}0.498 & \cellcolor{red!15}0.523 \\
Persp.~$s{=}0.1$ & \cellcolor{green!18}0.997 & \cellcolor{green!18}1.000 & \cellcolor{green!18}0.997 & \cellcolor{green!18}1.000 & \cellcolor{green!18}0.986 & \cellcolor{green!18}0.988 & \cellcolor{red!15}0.509 & \cellcolor{green!18}1.000 & \cellcolor{red!15}0.502 & \cellcolor{green!18}0.992 & \cellcolor{red!15}0.571 & \cellcolor{red!15}0.552 & \cellcolor{yellow!30}0.824 \\
Persp.~$s{=}0.2$ & \cellcolor{green!18}0.981 & \cellcolor{green!18}1.000 & \cellcolor{green!18}0.981 & \cellcolor{green!18}1.000 & \cellcolor{yellow!30}0.743 & \cellcolor{yellow!30}0.760 & \cellcolor{red!15}0.499 & \cellcolor{green!18}0.999 & \cellcolor{red!15}0.506 & \cellcolor{green!18}0.985 & \cellcolor{red!15}0.553 & \cellcolor{red!15}0.531 & \cellcolor{yellow!30}0.795 \\
Persp.~$s{=}0.3$ & \cellcolor{green!18}0.912 & \cellcolor{green!18}0.999 & \cellcolor{green!18}0.912 & \cellcolor{green!18}0.999 & \cellcolor{red!15}0.555 & \cellcolor{red!15}0.494 & \cellcolor{red!15}0.499 & \cellcolor{green!18}0.954 & \cellcolor{red!15}0.501 & \cellcolor{green!18}0.965 & \cellcolor{red!15}0.538 & \cellcolor{red!15}0.517 & \cellcolor{yellow!30}0.777 \\
CropRes. $1.1\times$ (r.crop) & \cellcolor{green!18}0.987 & \cellcolor{green!18}0.999 & \cellcolor{green!18}0.987 & \cellcolor{green!18}0.999 & \cellcolor{yellow!30}0.770 & \cellcolor{yellow!30}0.762 & \cellcolor{red!15}0.500 & \cellcolor{green!18}1.000 & \cellcolor{red!15}0.502 & \cellcolor{green!18}0.991 & \cellcolor{red!15}0.587 & \cellcolor{red!15}0.546 & \cellcolor{yellow!30}0.815 \\
CropRes. $1.2\times$ (c.crop) & \cellcolor{green!18}0.968 & \cellcolor{green!18}0.999 & \cellcolor{green!18}0.968 & \cellcolor{green!18}0.999 & \cellcolor{yellow!30}0.707 & \cellcolor{yellow!30}0.729 & \cellcolor{red!15}0.501 & \cellcolor{green!18}1.000 & \cellcolor{red!15}0.499 & \cellcolor{green!18}0.974 & \cellcolor{red!15}0.540 & \cellcolor{red!15}0.526 & \cellcolor{yellow!30}0.795 \\
CropRes. $1.3\times$ (r.crop) & \cellcolor{yellow!30}0.798 & \cellcolor{green!18}0.941 & \cellcolor{yellow!30}0.798 & \cellcolor{green!18}0.942 & \cellcolor{red!15}0.530 & \cellcolor{red!15}0.494 & \cellcolor{red!15}0.500 & \cellcolor{green!18}0.880 & \cellcolor{red!15}0.501 & \cellcolor{green!18}0.942 & \cellcolor{red!15}0.531 & \cellcolor{red!15}0.518 & \cellcolor{yellow!30}0.757 \\
CropStr. (wide $1.3\times$) & \cellcolor{green!18}0.966 & \cellcolor{green!18}1.000 & \cellcolor{green!18}0.966 & \cellcolor{green!18}0.999 & \cellcolor{yellow!30}0.687 & \cellcolor{yellow!30}0.706 & \cellcolor{red!15}0.492 & \cellcolor{green!18}0.999 & \cellcolor{red!15}0.420 & \cellcolor{green!18}0.976 & \cellcolor{red!15}0.559 & \cellcolor{red!15}0.540 & \cellcolor{yellow!30}0.832 \\
CropStr. (tall $1.3\times$) & \cellcolor{green!18}0.958 & \cellcolor{green!18}0.999 & \cellcolor{green!18}0.958 & \cellcolor{green!18}0.999 & \cellcolor{red!15}0.625 & \cellcolor{yellow!30}0.713 & \cellcolor{red!15}0.498 & \cellcolor{green!18}0.998 & \cellcolor{red!15}0.500 & \cellcolor{green!18}0.986 & \cellcolor{yellow!30}0.686 & \cellcolor{red!15}0.649 & \cellcolor{yellow!30}0.839 \\
PadRes. $0.5\times$ (mirror) & \cellcolor{red!15}0.639 & \cellcolor{red!15}0.503 & \cellcolor{red!15}0.606 & \cellcolor{red!15}0.499 & \cellcolor{red!15}0.499 & \cellcolor{red!15}0.501 & \cellcolor{red!15}0.501 & \cellcolor{red!15}0.500 & \cellcolor{red!15}0.498 & \cellcolor{red!15}0.578 & \cellcolor{red!15}0.512 & \cellcolor{red!15}0.498 & \cellcolor{red!15}0.567 \\
PadRes. $0.6\times$ (grey) & \cellcolor{yellow!30}0.815 & \cellcolor{green!18}0.954 & \cellcolor{yellow!30}0.815 & \cellcolor{green!18}0.956 & \cellcolor{green!18}0.968 & \cellcolor{red!15}0.498 & \cellcolor{red!15}0.502 & \cellcolor{red!15}0.502 & \cellcolor{red!15}0.502 & \cellcolor{yellow!30}0.784 & \cellcolor{red!15}0.503 & \cellcolor{red!15}0.502 & \cellcolor{yellow!30}0.687 \\
PadRes. $0.7\times$ (mean) & \cellcolor{green!18}0.948 & \cellcolor{green!18}0.999 & \cellcolor{green!18}0.948 & \cellcolor{green!18}0.999 & \cellcolor{green!18}0.999 & \cellcolor{red!15}0.495 & \cellcolor{red!15}0.501 & \cellcolor{red!15}0.499 & \cellcolor{red!15}0.504 & \cellcolor{green!18}0.899 & \cellcolor{red!15}0.507 & \cellcolor{red!15}0.501 & \cellcolor{yellow!30}0.748 \\
PadStr. (wide $0.5\times$) & \cellcolor{yellow!30}0.849 & \cellcolor{green!18}0.980 & \cellcolor{yellow!30}0.849 & \cellcolor{green!18}0.982 & \cellcolor{green!18}0.865 & \cellcolor{red!15}0.537 & \cellcolor{red!15}0.497 & \cellcolor{red!15}0.541 & \cellcolor{red!15}0.519 & \cellcolor{yellow!30}0.795 & \cellcolor{red!15}0.513 & \cellcolor{red!15}0.505 & \cellcolor{yellow!30}0.719 \\
PadStr. (tall $0.5\times$) & \cellcolor{yellow!30}0.827 & \cellcolor{green!18}0.970 & \cellcolor{yellow!30}0.827 & \cellcolor{green!18}0.966 & \cellcolor{red!15}0.574 & \cellcolor{red!15}0.535 & \cellcolor{red!15}0.502 & \cellcolor{red!15}0.501 & \cellcolor{red!15}0.499 & \cellcolor{green!18}0.882 & \cellcolor{red!15}0.527 & \cellcolor{red!15}0.505 & \cellcolor{yellow!30}0.718 \\
\bottomrule
\end{tabular}
}
\end{table}

\FloatBarrier

\section{Training Details}
\label{sec:supp_implementation_rewrite}

\paragraph{Architecture.}
VectorMark uses a $D=1024$ continuous key space and a 100-bit payload before
random linear-code expansion. The encoder and extractor are U-Net-style
networks~\cite{ronneberger2015unet} with GroupNorm. The key vector is projected
to a spatial conditioning map and concatenated with image features. The
extractor output is centered by subtracting the held-out unwatermarked mean and
then normalized before cosine search.

\paragraph{JND masking.}
The JND mask follows luminance-adaptation and contrast-masking principles
\cite{wu2017enhanced} and is used as a spatial gain map for the residual.

\begin{table}[tb]
\centering
\caption{\textbf{Selected VectorMark checkpoint and training evidence.}
The selected checkpoint is the 280k-step model used for the reported main
experiments.}
\label{tab:supp_model_training_evidence}
\small
\begin{tabular}{@{}ll@{}}
\toprule
Field & Evidence \\
\midrule
Training dataset & MS-COCO 2014 train, 82,783 images \\
Evaluation dataset & MS-COCO 2014 val subset, 1,000 images \\
Checkpoint training step & 280,000 \\
Encoder parameters & 2,980,006 \\
Extractor parameters & 3,000,049 \\
Stored key-generator tensors & 1,150,976 \\
Payload / code dimensions & 100 payload bits expanded to a 1024-dim codeword \\
Training image resolution & $256\times256$ \\
Evaluation image resolution & $512\times512$ \\
Training batch size & 32 \\
Evaluation seed & 42 \\
Training seed & 42 \\
Optimizer evidence & AdamW, weight decay $10^{-2}$ \\
\bottomrule
\end{tabular}
\end{table}

\paragraph{Training}
\label{sec:supp_training_curriculum}
The selected VectorMark checkpoint uses JND masking, centering, AdamW, and the loss
defined in the main paper. The runs used seed 42. 

The model is trained with a progressive
curriculum over 280k global steps. Every step includes the single-key path:
image reconstruction loss, vector-direction recovery loss, and ECC supervision.
After collusion training is activated, each step also includes a mean-collusion
path: $K$ independent keys are embedded into the same image, the resulting
watermarked images are averaged, and the extractor is trained toward the
average key direction. The collusion path also includes the pixel-space
linearity loss between the averaged watermarked image and direct embedding of
the normalized average key. The first
three phases of training curriculum build embedding, perceptual quality, and strong geometric
robustness. The final three phases reduce augmentation severity and tighten the
perceptual residual budget to recover image quality while retaining the
previously learned robustness.

Throughout collusion-active phases, the linearity loss weight is
$\lambda_{\rm lin}=100$. The sampled colluder count is an integer in the
phase-specific range. Photometric augmentations include JPEG
quality 50, Gaussian blur, Gaussian noise, median blur, salt-and-pepper noise,
resize-compression, and color transforms for brightness, contrast, saturation,
and hue.

\paragraph{Computational resources} Training is done on 1$\times$NVIDIA H200 GPU for approximately 25 hours for the selected 280k-step checkpoint. Main evaluation grid takes around 280 seconds. \Cref{tab:supp_throughput} gives throughput estimations for baselines on watermark encoding and extraction.

\begin{table}[tb]
\centering
\caption{\textbf{Watermark encoding and extraction throughput (images/s)} on a single NVIDIA H200 GPU, batch size~100, all evaluated at $512\times512$. \textit{Native Res.}: the resolution at which each method's encoder and extractor operate internally. Timings use wall-clock \texttt{perf\_counter} with CUDA synchronisation; first batch excluded as warm-up. Methods marked $\dag$ run on the CPU via the \texttt{invisible-watermark} library~\cite{shieldmnt2020invisible}. VectorMark extraction timing excludes ECC codeword decoding, which is only required when payload extraction is needed.}

\label{tab:supp_throughput}
\small
\begin{tabular}{@{} l c cc @{}}
\toprule
Method & Native Res. & Encode (img/s)$\uparrow$ & Extract (img/s)$\uparrow$ \\
\midrule
\multicolumn{4}{@{}l}{\textit{Traditional (non-learned)}} \\
DWT-DCT-SVD$^\dag$ & $256^2$ &    54 &      80   \\
DWT-DCT$^\dag$     & $256^2$ &   138 &     222   \\
SS-DCT             & $256^2$ & 11{,}937 & 54{,}578 \\
SS-FM              & $256^2$ & 35{,}329 & 81{,}974 \\
SS-DFT             & $256^2$ & 38{,}103 & 111{,}782 \\
\midrule
\multicolumn{4}{@{}l}{\textit{Learning-based}} \\
RivaGAN$^\dag$     & $256^2$ &      10  &      11   \\
StegaStamp         & $400^2$ &      10  &      64   \\
VINE-R             & $256^2$ &     109  &     819   \\
MaskWM             & $256^2$ &   1{,}006 &     483   \\
TrustMark-Q        & $256^2$ &   2{,}264 &   5{,}486 \\
MBRS               & $256^2$ &   2{,}554 &   5{,}211 \\
HiDDeN             & $128^2$ &   9{,}072 &   7{,}535 \\
\midrule
\textbf{VectorMark (Ours)} & $256^2$ & {2{,}258} & {2{,}105} \\
\bottomrule
\end{tabular}
\end{table}

\section{Evaluation Protocol}
\label{sec:supp_eval_protocol}

\paragraph{Transform suites.}
The photometric suite contains JPEG compression, Gaussian blur and noise,
median filtering, salt-and-pepper noise, brightness/contrast/saturation/hue
shifts, and downscaling. The geometric suite contains flips, rotations,
flip-rotations, perspective warps, crop-and-resize, non-uniform stretching, and
pad-and-resize transforms. Full parameter values are listed in
\Cref{tab:supp_noise_eval}.

\begin{table}[tb]
  \caption{\textbf{Deterministic evaluation noise parameters.} These distinct, fixed configurations are used to evaluate the robustness of all watermarking methods to valuemetric and geometric transformations.}
  \label{tab:supp_noise_eval}
  \centering
  \small
  \begin{tabular}{@{} l p{0.75\linewidth} @{}}
    \toprule
    \textbf{Noise Category} & \textbf{Deterministic Configurations} \\
    \midrule
    \multicolumn{2}{@{}l}{\textit{Valuemetric Operations}} \\
    \midrule
    JPEG Compression     & Quality: $50$, $60$ \\
    Gaussian Blur        & Kernel $5\times5$ with $\sigma \in \{1.0, 1.5\}$ \\
    Gaussian Noise       & $\sigma = 0.1$ (Fixed seed: 42) \\
    Median Filter        & Kernel size: $5$ \\
    Salt \& Pepper       & Probability $p = 0.10$ (Fixed seed: 42) \\
    Downscale \& Resize  & $0.4\times$ original resolution, bilinear interpolation \\
    Color Jitter         & Brightness ($0.5$, $1.5$), Contrast ($0.5$, $1.5$), Satur. ($0.5$, $1.5$) \\
    Hue Shift            & $\Delta = 0.15$ \\
    \midrule
    \multicolumn{2}{@{}l}{\textit{Geometric \& Spatial Operations}} \\
    \midrule
    Horizontal Flip      & Probability $p=1.0$ \\
    Rotate (Fill)   & $\theta \in \{5^\circ, 10^\circ, 15^{\circ}\}$ with grey fill,  $\theta \in \{10^\circ, 15^\circ\}$ with mirror fill \\
    Rotate (Crop) & $\theta \in \{10^\circ, 15^\circ\}$ followed by centre and random crops (seed: 42)\\
    Flip + Rotate        & $p_\text{flip}=1.0$ combined with all of the above rotations \\
    Perspective Warp     & Distortion scale $\in \{0.10, 0.20, 0.30\}$ (seed: 42) \\
    Crop \& Resize (Iso) & Zoom scales $1.1\times$ (random), $1.2\times$ (center), $1.3\times$ (random) \\
    Crop Stretch (Aniso) & Wide ($1.3\times_x, 1.0\times_y$) and Tall ($1.0\times_x, 1.3\times_y$) with centre crop \\
    Pad \& Resize (Iso)  & Shrink scales $0.5\times$ (reflection), $0.6\times$ (grey), $0.7\times$ (mean) \\
    Pad Stretch (Aniso)  & Wide ($0.5\times_x, 1.0\times_y$) and Tall ($1.0\times_x, 0.5\times_y$) with grey fill \\
    \bottomrule
  \end{tabular}
\end{table}

\paragraph{Confidence intervals.}
For TPR, we report Wilson 95\% binomial confidence intervals computed from the
number of successes and trials. For transformed single-key groups, a trial is
one transform--image cell. This interval captures deterministic evaluation
success over the evaluated grid.

\begin{table}[tb]
\centering
\caption{\textbf{MS-COCO success counts and Wilson intervals.}
Counts are from the main evaluation grid at $\mathrm{FPR}=10^{-6}$. The
photometric group contains 15 transforms, the geometric group contains 30
transforms.}
\label{tab:supp_ci_headline}
\small
\begin{tabular}{@{}lrrrr@{}}
\toprule
Setting & Trials $n$ & Successes & TPR & Wilson 95\% CI \\
\midrule
Clean & 1,000 & 1,000 & 1.0000 & [0.9962, 1.0000] \\
Photometric & 15,000 & 14,993 & 0.9995 & [0.9990, 0.9998] \\
Geometric & 30,000 & 29,554 & 0.9851 & [0.9837, 0.9864] \\
Averaging collusion, $K=2$ & 1,000 & 1,000 & 1.0000 & [0.9962, 1.0000] \\
Averaging collusion, $K=5$ & 1,000 & 1,000 & 1.0000 & [0.9962, 1.0000] \\
Averaging collusion, $K=10$ & 1,000 & 996 & 0.9960 & [0.9898, 0.9984] \\
\bottomrule
\end{tabular}
\end{table}

\section{Full-Registry Attribution and Capacity}
\label{sec:supp_registry}

\paragraph{Returned-set protocol.}
For a true colluder set $C$ and a registry $\mathcal U$ of size $N$, the
returned set is $\widehat C=\{u\in\mathcal U:s_u\geq\tau_N\}$.
We explicitly score every registered identity, including the contributors. Recall measures the fraction of
contributors detected, $|\widehat C\cap C|/|C|$; precision measures the fraction
of returned identities that are contributors, $|\widehat C\cap C|/|\widehat C|$
for nonempty returned sets. All-colluder TPR is the frequency of
$C\subseteq\widehat C$, whereas exact-set recovery requires
$\widehat C=C$. No-FA is the frequency of $\widehat C\subseteq C$, and empirical
FWER is the frequency of $\widehat C\setminus C\neq\varnothing$.

\paragraph{Registry scaling.}
\Cref{tab:registry_full} evaluates 2,000 test images and three independently
sampled registries at $N\in\{10^3,10^4,10^5,10^6\}$ and
$K\in\{5,10,15,20\}$. The nominal family-wise budget is $\delta=0.01$, so
$\alpha_N=\delta/N$; at $N=10^6$, $\alpha_N=10^{-8}$.
The analytic million-user ($N=10^6$) threshold is $\tau_N=0.1742$. At $K=10$, mean returned-set size is 9.99,
recall is .998, precision is .999, and exact-set recovery is .980.
Across the evaluated registry/coalition settings, empirical FWER ranges from .009 to .013, close to the nominal 1\% budget.

\begin{table}[tbp]
\centering
\caption{\textbf{Complete database-scale attribution.} Each registry is fully
scanned at nominal $\alpha_N=0.01/N$. Results use 2,000 test images and three
independently sampled registries. No-FA excludes all innocent identities,
exact-set recovery requires both No-FA and detection of every colluder.}
\label{tab:registry_full}
\small
\begin{tabular}{@{}rrcccccc@{}}
\toprule
$N$ & $K$ & Recall & Precision & All-colluder & No-FA & Exact set & FWER \\
\midrule
$10^{3}$ & 5 & 1.000 & .998 & 1.000 & .990 & .990 & .010 \\
 & 10 & 1.000 & .999 & .999 & .990 & .989 & .010 \\
 & 15 & .995 & .999 & .950 & .989 & .940 & .011 \\
 & 20 & .951 & 1.000 & .538 & .991 & .533 & .009 \\
\midrule
$10^{4}$ & 5 & 1.000 & .999 & 1.000 & .991 & .991 & .009 \\
 & 10 & 1.000 & .999 & .998 & .990 & .988 & .010 \\
 & 15 & .986 & .999 & .875 & .990 & .865 & .010 \\
 & 20 & .893 & .999 & .244 & .987 & .241 & .013 \\
\midrule
$10^{5}$ & 5 & 1.000 & .998 & 1.000 & .990 & .990 & .010 \\
 & 10 & .999 & .999 & .996 & .988 & .984 & .012 \\
 & 15 & .968 & .999 & .732 & .990 & .725 & .010 \\
 & 20 & .804 & .998 & .065 & .991 & .064 & .009 \\
\midrule
$10^{6}$ & 5 & 1.000 & .998 & 1.000 & .990 & .990 & .010 \\
 & 10 & .998 & .999 & .990 & .989 & .980 & .011 \\
 & 15 & .936 & .998 & .525 & .989 & .521 & .011 \\
 & 20 & .685 & .997 & .007 & .991 & .007 & .009 \\
\bottomrule
\end{tabular}
\end{table}

\paragraph{Scaling under ideal setting.}
For an isotropic innocent direction, $\operatorname{Var}(S)=1/D$.
The large-$D$ approximation and ideal contributor signal give
\begin{equation}
 \tau_{N,\delta}\approx\sqrt{\frac{2\log(N/\delta)}{D}},
 \qquad s_{\mathrm{true}}\approx\frac{1}{\sqrt K},
 \qquad K\lesssim\frac{D}{2\log(N/\delta)}.
 \label{eq:supp_registry_scaling}
\end{equation}
This shows that under an ideal setting a typical pairwise
correlations have scale $1/\sqrt D$, while a rough extreme-value estimate
for $N$ random keys is
$\max_{i\ne j}|\langle v_i,v_j\rangle|\approx\sqrt{4\log N/D}$,
approximately 0.232 at $D=1024,N=10^6$.

At $K=10$, the reported individual true-score mean is 0.285, below the ideal
$1/\sqrt{10}\approx 0.316$. The first, fifth, and tenth percentiles of the
minimum true score per query are 0.174, 0.200, and 0.213. Under the same null
model, mapping these through
$N_{\max}=\delta/\Pr_{H_0}(S\geq\tau)$ gives approximate registry-size
projections of $1.1\times10^6$, $2.0\times10^8$, and $3.2\times10^9$ for
99\%, 95\%, and 90\% all-colluder TPR, respectively. These are
\emph{model-based extrapolations} under the ideal setting, ignoring the noise, the approximation and prediction errors. 

\paragraph{Identity collisions versus false attribution.}
For uniformly sampled 100-bit identities, the birthday union bound is
$\Pr(\text{identity collision})\leq N(N-1)/2^{101}$, approximately
$3.94\times10^{-19}$ at $N=10^6$ and $3.94\times10^{-13}$ at $N=10^9$.
This addresses concern regarding duplicate input identities. If the binary matrix $G$ has full column
rank, different identities also have distinct codewords, while the orthogonal
rotation preserves equality and pairwise inner products.

\FloatBarrier

\section{Dimension and Soft-Versus-Hard Scoring Controls}
\label{sec:supp_controls}

\Cref{tab:controls_full} uses 600 disjoint test images,
three database seeds, and full matching against $N=10^6$ identities with a
nominal 1\% family-wise budget. VectorMark uses analytical thresholds for soft cosine
and hard agreement; the baseline scoring rules are calibrated separately using held-out images. 

For VectorMark, the hard control uses hard codeword agreement rather than soft cosine similarity. For TrustMark-Q,
ChunkySeal~\cite{petrov2025chunkyseal}, PixelSeal~\cite{soucek2025pixelseal},
and VideoSeal~\cite{fernandez2024videoseal}, the paired rows similarly compare
native hard agreement with cosine scoring of the same decoder logits.
DWT-DCT supplies a training-free payload-dimension sweep from 256 to 1024 bits.
PSNR is reported at each method's evaluated operating point.

\begin{table}[tbp]
\centering
\caption{\textbf{Complete dimension and scoring controls.}
Each coalition-size cell shows \emph{exact-set recovery} on the first line and
\emph{recall / precision} on the second line, all three rates are percentages.
Results use $N=10^6$, nominal $\delta=0.01$, 600 disjoint test images, and
three database seeds. \emph{PSNR} is measured against the unmarked host.
Bold indicates the largest value for each metric and coalition size,
including ties. }
\label{tab:controls_full}
\small
\setlength{\tabcolsep}{3pt}
\renewcommand{\arraystretch}{1.05}
\begin{tabular}{@{}lrlccccr@{}}
\toprule
Method & $D$ & Score
& $K=5$ & $K=10$ & $K=12$ & $K=15$
& \makecell{PSNR\\(dB)} \\
\midrule

\textbf{VectorMark} & 1024 & Soft cosine
& \vmctl{\textbf{99.2}}{\textbf{100.0}}{\textbf{99.9}}
& \vmctl{\textbf{97.4}}{\textbf{99.8}}{\textbf{99.9}}
& \vmctl{\textbf{91.4}}{\textbf{98.9}}{\textbf{99.9}}
& \vmctl{\textbf{52.8}}{\textbf{93.5}}{\textbf{99.9}}
& 40.28 \\

VectorMark & 1024 & Hard agreement
& \vmctl{99.1}{\textbf{100.0}}{99.8}
& \vmctl{62.4}{94.4}{\textbf{99.9}}
& \vmctl{13.4}{82.3}{99.6}
& \vmctl{0.0}{56.5}{99.0}
& 40.28 \\

\midrule

DWT-DCT & 256 & Hard agreement
& \vmctl{0.3}{11.7}{27.4}
& \vmctl{0.0}{0.6}{5.0}
& \vmctl{0.0}{0.2}{2.3}
& \vmctl{0.0}{0.1}{1.0}
& 36.79 \\

DWT-DCT & 512 & Hard agreement
& \vmctl{9.4}{18.8}{32.3}
& \vmctl{0.0}{3.5}{12.6}
& \vmctl{0.0}{1.8}{10.1}
& \vmctl{0.0}{0.6}{7.2}
& 36.79 \\

DWT-DCT & 1024 & Hard agreement
& \vmctl{49.0}{71.8}{88.9}
& \vmctl{3.6}{22.3}{58.4}
& \vmctl{1.0}{14.5}{48.4}
& \vmctl{0.0}{7.2}{31.7}
& 36.79 \\

SS-DFT & 1024 & Cosine
& \vmctl{86.2}{94.2}{98.4}
& \vmctl{0.8}{40.1}{85.5}
& \vmctl{0.0}{19.4}{72.2}
& \vmctl{0.0}{5.7}{46.4}
& 33.85 \\

\addlinespace[2pt]

TrustMark-Q & 100 & Hard agreement
& \vmctl{0.0}{1.3}{6.7}
& \vmctl{0.0}{0.0}{0.2}
& \vmctl{0.0}{0.0}{0.1}
& \vmctl{0.0}{0.0}{0.0}
& 40.52 \\

TrustMark-Q & 100 & Logit cosine
& \vmctl{0.0}{5.6}{24.6}
& \vmctl{0.0}{0.1}{0.7}
& \vmctl{0.0}{0.0}{0.5}
& \vmctl{0.0}{0.0}{0.0}
& 40.52 \\

\addlinespace[2pt]

ChunkySeal & 1024 & Hard agreement
& \vmctl{92.3}{95.4}{96.0}
& \vmctl{0.2}{42.4}{81.0}
& \vmctl{0.0}{22.5}{71.5}
& \vmctl{0.0}{9.1}{56.2}
& 44.70 \\

ChunkySeal & 1024 & Logit cosine
& \vmctl{94.9}{96.2}{96.1}
& \vmctl{44.1}{74.8}{83.4}
& \vmctl{10.8}{58.4}{77.8}
& \vmctl{0.0}{34.7}{70.9}
& 44.70 \\

\addlinespace[2pt]

PixelSeal & 256 & Hard agreement
& \vmctl{0.1}{34.8}{90.1}
& \vmctl{0.0}{1.3}{12.2}
& \vmctl{0.0}{0.4}{5.3}
& \vmctl{0.0}{0.1}{2.2}
& \textbf{46.13} \\

PixelSeal & 256 & Logit cosine
& \vmctl{43.3}{83.9}{99.7}
& \vmctl{0.0}{7.9}{54.1}
& \vmctl{0.0}{3.0}{31.3}
& \vmctl{0.0}{0.6}{9.1}
& \textbf{46.13} \\

\addlinespace[2pt]

VideoSeal & 256 & Hard agreement
& \vmctl{0.0}{19.4}{67.4}
& \vmctl{0.0}{0.3}{3.4}
& \vmctl{0.0}{0.1}{0.6}
& \vmctl{0.0}{0.0}{0.2}
& 45.07 \\

VideoSeal & 256 & Logit cosine
& \vmctl{9.1}{61.4}{98.1}
& \vmctl{0.0}{2.9}{25.4}
& \vmctl{0.0}{1.0}{11.2}
& \vmctl{0.0}{0.2}{2.1}
& 45.07 \\

\bottomrule
\end{tabular}
\end{table}

\paragraph{Interpretation.}
Soft scoring improves VectorMark's $K=10$ exact-set recovery from 0.624 to
0.974 on identical outputs, and improves ChunkySeal from 0.002 to 0.441.
Increasing DWT-DCT from 256 to 1024 bits raises $K=5$ exact-set recovery
from 0.003 to 0.490. The short-run ablations in \Cref{tab:main_ablation_studies}
also show gains from dimension and collusion-aware training. Together these controls support the roles of
dimensionality, retained soft evidence, and learned robustness. 

\section{Scheme-Aware, Black-Box and White-Box Attacks}
\label{sec:supp_attacks}

\paragraph{Access model and evaluation.}
The scheme-aware adversary knows the construction and owns aligned,
differently marked copies of the same host image. It does not receive the
clean host, secret keys, target encoder/extractor parameters, or target
detector outputs. The residual subspace removal attack below is query-free
and scheme-aware, it uses the collection of marked images from a set of colluders. All six evaluated methods receive identical attack settings and images. Detection scans a
million-identity registry at nominal $\delta=0.01$, where the binary methods use
native hard agreement with binomial-null thresholds. PSNR is measured
against the clean host for evaluation.

\paragraph{Scheme-aware patchwise residual-subspace removal.}
For each non-overlapping $8\times8$ patch $p$, let the rows of $R^{(p)}$ be
$r_i^{(p)}=\operatorname{vec}(X_i^{(p)}-\bar X^{(p)})$.
Let $\mu^{(p)}$ be the spatial mean of $\bar X^{(p)}$, computed separately
per channel and broadcast over the patch, and let
$\tilde x^{(p)}=\operatorname{vec}(\bar X^{(p)}-\mu^{(p)})$.
The attack is
\begin{align}
 P_R^{(p)}&=(R^{(p)})^\top
 \bigl(R^{(p)}(R^{(p)})^\top+\lambda I\bigr)^{-1}R^{(p)},\\
 x_{\mathrm{rm}}^{\gamma,(p)}&=\operatorname{vec}(\bar X^{(p)})
 -\gamma P_R^{(p)}\tilde x^{(p)}.
 \label{eq:supp_subspace_remove}
\end{align}
Patchwise estimation separates unrelated spatial regions, while per-channel
centering avoids constant color shifts. We evaluate $K=10$ and
$\gamma\in\{0.25,0.50,1\}$.

\paragraph{Black-box reconstruction and optimization.}
WmForger~\cite{soucek2025forging} uses an external image-preference model for
50 optimization steps. The black-box Regen-VAE attack follows the reconstruction-based
removal setting of Zhao et al.~\cite{zhao2024removable} and uses CompressAI's
pretrained \texttt{bmshj2018-factorized} model at quality level 8
\cite{balle2018variational}. This is the factorized-prior learned image codec, and not a Stable Diffusion VAE. The other black-box removal attack, SD1.5 img2img, uses Stable Diffusion 1.5
\cite{rombach2022high} at strength .02 for diffusion-based denoising. These three attacks use $K=5$.

\begin{table}[tbp]
\centering
\caption{\textbf{Complete attack results.}
Each cell shows \emph{exact-set recovery $\vert$ precision} above
\emph{recall $\vert$ PSNR}. Rates are percentages, PSNR is in dB relative
to the unmarked host. ``Subsp. ($\gamma$)'' denotes residual-subspace
removal; SD1.5 denotes img2img denoising attack. Results use identical images and attack
parameters, $N=10^6$, and nominal $\delta=0.01$; binary methods use their
native hard matcher. Bold indicates the largest value for each metric
within each attack row, including ties. }
\label{tab:attacks_full}
\small
\setlength{\tabcolsep}{2pt}
\renewcommand{\arraystretch}{1.05}
\begin{tabular}{@{}lrcccccc@{}}
\toprule
Attack & $K$
& \textbf{VectorMark}
& SS-DFT
& TrustMark
& ChunkySeal
& PixelSeal
& VideoSeal \\
\midrule

Mean & 10
& \vmcell{\textbf{97.9}}{\textbf{99.9}}{\textbf{99.8}}{41.03}
& \vmcell{1.3}{85.4}{40.5}{43.54}
& \vmcell{0.0}{0.2}{0.0}{42.39}
& \vmcell{0.6}{68.7}{32.9}{\textbf{51.32}}
& \vmcell{0.0}{10.8}{1.1}{50.49}
& \vmcell{0.0}{3.3}{0.4}{50.83} \\

\addlinespace[3pt]

Subsp. ($.25$) & 10
& \vmcell{\textbf{70.4}}{\textbf{99.9}}{\textbf{94.5}}{37.73}
& \vmcell{0.1}{73.9}{28.0}{36.18}
& \vmcell{0.0}{0.1}{0.0}{35.63}
& \vmcell{0.0}{32.1}{9.5}{\textbf{40.08}}
& \vmcell{0.0}{3.2}{0.3}{36.55}
& \vmcell{0.0}{2.1}{0.2}{36.23} \\

\addlinespace[3pt]

Subsp. ($.50$) & 10
& \vmcell{\textbf{8.3}}{\textbf{87.8}}{\textbf{44.8}}{34.03}
& \vmcell{0.0}{28.5}{5.1}{30.71}
& \vmcell{0.0}{0.0}{0.0}{30.55}
& \vmcell{0.0}{8.3}{2.0}{\textbf{34.39}}
& \vmcell{0.0}{0.7}{0.1}{30.72}
& \vmcell{0.0}{0.1}{0.0}{30.34} \\

\addlinespace[3pt]

Subsp. ($1$) & 10
& \vmcell{\textbf{0.0}}{\textbf{5.8}}{\textbf{1.2}}{\textbf{28.99}}
& \vmcell{\textbf{0.0}}{0.0}{0.0}{24.82}
& \vmcell{\textbf{0.0}}{0.0}{0.0}{24.98}
& \vmcell{\textbf{0.0}}{0.9}{0.2}{28.51}
& \vmcell{\textbf{0.0}}{0.0}{0.0}{24.78}
& \vmcell{\textbf{0.0}}{0.0}{0.0}{24.37} \\

\midrule

WmForger & 5
& \vmcell{\textbf{72.5}}{\textbf{90.4}}{\textbf{83.8}}{32.68}
& \vmcell{61.0}{89.5}{74.3}{32.98}
& \vmcell{0.0}{1.0}{0.2}{32.67}
& \vmcell{10.5}{25.0}{18.7}{33.26}
& \vmcell{0.0}{7.5}{1.9}{33.24}
& \vmcell{0.0}{4.0}{0.9}{\textbf{33.29}} \\

\addlinespace[3pt]

Regen-VAE & 5
& \vmcell{\textbf{98.0}}{\textbf{99.8}}{\textbf{99.8}}{36.81}
& \vmcell{80.5}{97.0}{91.2}{36.30}
& \vmcell{0.0}{3.0}{0.6}{36.73}
& \vmcell{3.5}{10.5}{6.8}{\textbf{37.98}}
& \vmcell{0.0}{75.5}{22.0}{37.93}
& \vmcell{0.0}{48.8}{12.0}{37.94} \\

\addlinespace[3pt]

SD1.5 & 5
& \vmcell{0.0}{0.0}{0.0}{25.50}
& \vmcell{\textbf{0.5}}{\textbf{28.0}}{\textbf{10.2}}{25.37}
& \vmcell{0.0}{1.0}{0.2}{25.51}
& \vmcell{0.0}{0.0}{0.0}{\textbf{25.54}}
& \vmcell{0.0}{1.0}{0.2}{25.52}
& \vmcell{0.0}{0.0}{0.0}{25.51} \\

\bottomrule
\end{tabular}
\end{table}

\paragraph{Interpretation.}
\Cref{tab:attacks_full} uses the same attack
settings across methods. VectorMark has
0.725 exact-set recovery under WmForger and 0.980 under Regen-VAE black-box attacks.
Stronger scheme-aware adaptive subspace removal attack also reduces attribution at the cost of image quality.
SD1.5 removes all VectorMark contributor detections at 25.50\,dB PSNR.

\paragraph{White-box attacks.}
\label{subsec:framing}
A separate experiment grants the attacker the ability to re-embed a chosen
innocent identity with the white-box encoder access. Re-embedding the mean of five marked copies detects that
target innocent identity in 1,000/1,000 images, reduces source recall to zero, and returns
1.018 identities on average at 37.45\,dB.
This access is stronger than knowing the scheme or having an encoder without
the chosen victim's key. The result shows that watermark attribution alone
does not provide secure authentication or prevent watermark forgery. An encoder
and key service needs appropriate authentication and access control, and
attribution should be corroborated with independent evidence.

\FloatBarrier

\section{Deliberate Multilabel Embedding and Localized Attribution}
\label{sec:supp_multilabel}

\paragraph{One-pass global embedding.}
A publisher can intentionally embed several registered labels by a single
encoder call with
\begin{equation}
 v_M=\frac{\sum_{j=1}^{M}v_j}{\|\sum_{j=1}^{M}v_j\|_2},
 \qquad I_M=E_\theta(I,v_M).
 \label{eq:supp_multilabel}
\end{equation}
This is not averaging $K$ independently marked copies: the normalized
combined direction is embedded directly. \Cref{tab:multilabel_full} reports
1,000 test images, a fixed million-identity registry, analytical thresholds,
and a nominal 1\% family-wise budget. At $M=15$, recall is 0.995, precision
0.999, all-label detection 0.926, and exact-set recovery 0.915; reported image
quality is 40.16\,dB. Larger label sets reduce recovery, as the full sweep to
$M=100$ shows.

\begin{table}[tbp]
\centering
\caption{\textbf{One-pass deliberate multilabel embedding.} A single encoder
call embeds the normalized sum of $M$ intended keys. Results use 1,000 test
images, a fixed million-identity registry, and nominal
$\delta=0.01$. Reported PSNR is 40.16\,dB.}
\label{tab:multilabel_full}
\small
\begin{tabular}{@{}rcccccc@{}}
\toprule
$M$ & Recall & Precision & All-label & No-FA & Exact set & FWER \\
\midrule
1 & 1.000 & .994 & 1.000 & .988 & .988 & .012 \\
5 & 1.000 & .998 & 1.000 & .988 & .988 & .012 \\
10 & 1.000 & .999 & 1.000 & .994 & .994 & .006 \\
15 & .995 & .999 & .926 & .989 & .915 & .011 \\
20 & .924 & .999 & .184 & .989 & .183 & .011 \\
25 & .750 & 1.000 & .002 & .992 & .002 & .008 \\
30 & .543 & .999 & .000 & .990 & .000 & .010 \\
40 & .250 & .999 & .000 & .990 & .000 & .010 \\
50 & .113 & .997 & .000 & .992 & .000 & .008 \\
75 & .021 & .780 & .000 & .983 & .000 & .017 \\
100 & .006 & .458 & .000 & .991 & .000 & .009 \\
\bottomrule
\end{tabular}
\end{table}

\paragraph{Multi-label embedding with localized watermarking.}
We compare global identity-set retrieval with Watermark
Anything (WAM)~\cite{sander2025watermark} and MaskWM~\cite{hu2025mask}, which
can assign messages to spatially distinct regions. For these localized
methods, $M$ similar-area rectangles are separated by 12-pixel clean gutters.
Blind decoding hides the region coordinates: WAM uses its published
spatial-clustering decoder, and MaskWM decodes connected components of its
predicted mask. We assume that the ground-truth (GT) condition supplies the embedding masks.
For every $M$ and localization mode, thresholds are calibrated using 250
disjoint held-out images at nominal $\delta=0.01$, and all $10^6$ registry
identities are scored. Evaluation uses 1,000 test images.

\begin{table}[tbp]
\centering
\caption{\textbf{Localized versus global multi-identity attribution.}
Each cell shows \emph{exact-set recovery $\vert$ precision} above
\emph{recall $\vert$ PSNR}. Rates are percentages, PSNR is in dB.
All rows use 1,000 test images, $N=10^6$, and nominal $\delta=0.01$.
VectorMark uses the analytical threshold; WAM and MaskWM use held-out thresholds.
GT supplies ground-truth embedded regions; blind decoding
does not assume their availability. VectorMark embeds a single vector corresponding to average of all labels. Bold indicates the largest value
for each metric at each $M$, including ties. }
\label{tab:localized_full}
\small
\setlength{\tabcolsep}{5pt}
\renewcommand{\arraystretch}{1.05}
\begin{tabular}{@{}rccccc@{}}
\toprule
& \multicolumn{2}{c}{WAM}
& \multicolumn{2}{c}{MaskWM}
& \textbf{VectorMark} \\
\cmidrule(lr){2-3}
\cmidrule(lr){4-5}
$M$ & Blind & GT mask & Blind & GT mask & Global \\
\midrule

1
& \vmcell{98.5}{99.1}{99.6}{37.01}
& \vmcell{98.5}{99.1}{99.6}{37.01}
& \vmcell{\textbf{99.2}}{\textbf{99.6}}{\textbf{100.0}}{35.76}
& \vmcell{\textbf{99.2}}{\textbf{99.6}}{\textbf{100.0}}{35.76}
& \vmcell{98.8}{99.4}{\textbf{100.0}}{\textbf{40.16}} \\

\addlinespace[3pt]

5
& \vmcell{88.3}{\textbf{99.9}}{96.9}{37.58}
& \vmcell{87.8}{\textbf{99.9}}{96.8}{37.58}
& \vmcell{94.3}{\textbf{99.9}}{98.1}{30.97}
& \vmcell{\textbf{99.4}}{\textbf{99.9}}{\textbf{100.0}}{30.97}
& \vmcell{98.8}{99.8}{\textbf{100.0}}{\textbf{40.16}} \\

\addlinespace[3pt]

10
& \vmcell{30.2}{99.7}{82.9}{37.96}
& \vmcell{29.4}{99.7}{82.4}{37.96}
& \vmcell{79.7}{\textbf{99.9}}{95.1}{30.41}
& \vmcell{99.0}{\textbf{99.9}}{\textbf{100.0}}{30.41}
& \vmcell{\textbf{99.4}}{\textbf{99.9}}{\textbf{100.0}}{\textbf{40.16}} \\

\addlinespace[3pt]

12
& \vmcell{14.1}{99.4}{74.4}{38.10}
& \vmcell{13.1}{99.4}{74.0}{38.10}
& \vmcell{77.5}{\textbf{100.0}}{94.8}{30.35}
& \vmcell{\textbf{99.8}}{\textbf{100.0}}{\textbf{100.0}}{30.35}
& \vmcell{98.6}{99.9}{\textbf{100.0}}{\textbf{40.16}} \\

\addlinespace[3pt]

15
& \vmcell{1.9}{99.2}{63.6}{38.28}
& \vmcell{1.7}{99.2}{63.1}{38.28}
& \vmcell{67.9}{99.9}{93.1}{30.32}
& \vmcell{\textbf{99.3}}{\textbf{100.0}}{\textbf{100.0}}{30.32}
& \vmcell{91.5}{99.9}{99.5}{\textbf{40.16}} \\

\addlinespace[3pt]

17
& \vmcell{0.5}{98.3}{56.2}{38.40}
& \vmcell{0.5}{98.5}{55.6}{38.40}
& \vmcell{66.2}{99.9}{93.0}{30.31}
& \vmcell{\textbf{99.4}}{\textbf{100.0}}{\textbf{100.0}}{30.31}
& \vmcell{72.9}{99.9}{98.2}{\textbf{40.16}} \\

\addlinespace[3pt]

20
& \vmcell{0.0}{97.4}{40.6}{38.56}
& \vmcell{0.0}{97.2}{39.7}{38.56}
& \vmcell{62.7}{99.9}{93.1}{30.34}
& \vmcell{\textbf{97.1}}{\textbf{100.0}}{\textbf{99.9}}{30.34}
& \vmcell{18.3}{99.9}{92.4}{\textbf{40.16}} \\

\bottomrule
\end{tabular}
\end{table}

At $M=20$, MaskWM with GT masks reaches 0.971 exact-set recovery at
30.34\,dB, versus 0.627 with blind localization. The global VectorMark entry
in this comparison is 0.183 at 40.16\,dB. WAM's blind recall is 0.406.
The localization gap and differing image qualities show the trade-off:
localized schemes can retain more labels when spatial masks are available,
whereas VectorMark retrieves a global set without spatial partitioning at
a higher-PSNR operating point. 

\section{Limitations}
\label{sec:limitations}

\paragraph{Model assumptions and calibration.}
The geometric analysis characterizes ideal equal-weight mixtures of
near-orthogonal keys. Its application to images depends on the learned
encoder and extractor preserving the corresponding mixture direction.
Recovery noise, nonlinear image processing, key correlations, and unequal
contributions can reduce the minimum contributor score and therefore
all-colluder recovery. Similarly, the shifted-Beta thresholds are exact
under the spherical-null assumption, not for every learned extractor or
image distribution. Centering and a distributional fit on held-out
samples do not by themselves certify rare-event tail probabilities.
Database-level Bonferroni control requires valid marginal null
probabilities, although it does not require independent scores.
Deployment on a new distribution therefore requires checking the
calibration and attribution performance under the intended conditions.

\paragraph{Security and attribution.}
The attack experiments establish comparative robustness under the
specified access assumptions. Strong removal or regeneration can suppress attribution, and
the evaluation is not exhaustive over white-box optimization,
detector-query attacks, or adversarially selected registries.
An attacker with the ability to embed a chosen registered identity can
also create misleading attribution, as demonstrated by the framing
experiment. Secure deployment therefore requires protecting the
embedding and key-management services and distinguishing evidence of a
watermark from evidence that a particular person performed an action.
Watermark scores should be corroborated with independent records and
human review. The method alone cannot prevent undisclosed tracking or
resolve disputes about ownership.

\paragraph{Computational scaling.}
For key dimension $D$ and registry size $N$, exhaustive attribution
requires $O(ND)$ arithmetic per query, separately from neural extraction.
Materializing all registry vectors also requires $O(ND)$ storage.
Approximate retrieval may reduce search cost, but candidates omitted
during retrieval cannot be recovered by subsequent exact re-ranking.
Consequently, approximate indexing introduces an additional
recall--efficiency trade-off that must be evaluated before deployment
at larger scales.

\section{Broader Impact}
\label{sec:broader_impact}

\paragraph{Intended use.}
The intended use is forensic provenance for images distributed by an owner who
embeds user-specific marks and later investigates leaked copies. A watermark
score is probabilistic forensic evidence, and it should not be used as a standalone proof of wrongdoing.
Operational deployments should combine watermark evidence with access logs,
audit records, and human review.

\paragraph{Potential benefits.}
Robust user-specific watermarking can help content owners investigate leaks,
discourage unauthorized redistribution, and support provenance infrastructure
for digital media.

\paragraph{Potential harms and safeguards.}
Fingerprinting can also be misused for undisclosed tracking or punitive
attribution. Deployments should disclose watermarking where appropriate, define
retention and appeal policies, avoid high-stakes decisions based only on a
watermark score, and assess the privacy implications of user-specific marks.
Open release improves reproducibility but may also help adversaries study
removal attacks. 

\section{Qualitative Comparison}
\label{subsec:supp_quality}

Qualitative comparisons show original images, watermarked images, magnified
residuals, and JND maps, where applicable for various baseline methods for further imperceptibility analysis.

\begin{figure}[tbp]
\centering
\setlength{\tabcolsep}{2pt}
\renewcommand{\arraystretch}{0.85}
\small
\begin{tabular}{@{}cccc@{}}
  Original & Watermarked & Residual~${\times}10$ & JND Mask \\[2pt]
  \includegraphics[width=0.24\linewidth]{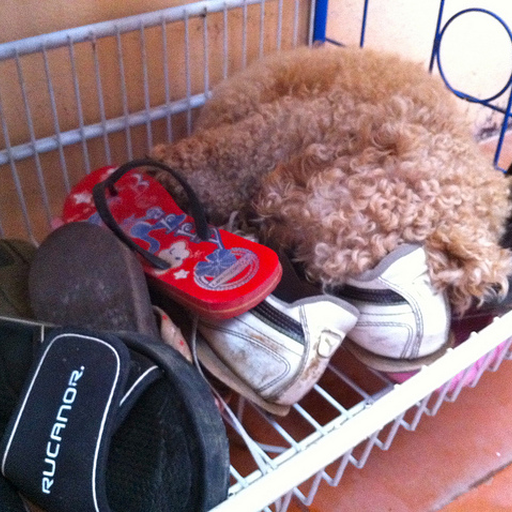} &
  \includegraphics[width=0.24\linewidth]{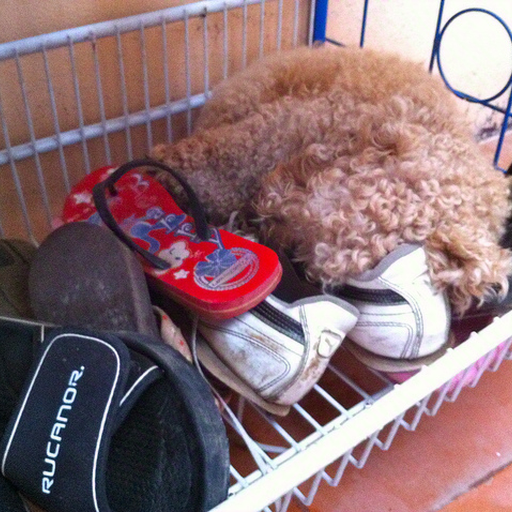} &
  \includegraphics[width=0.24\linewidth]{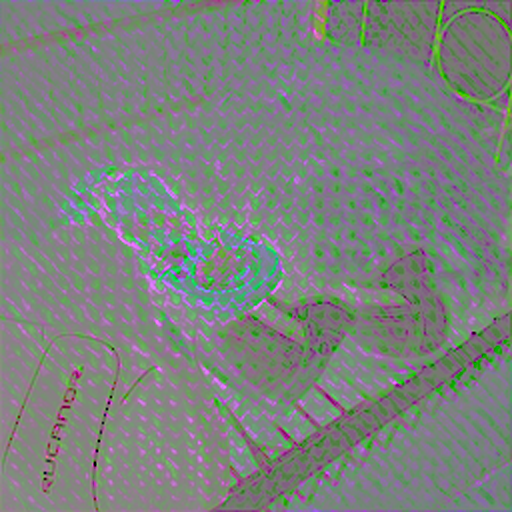} &
  \includegraphics[width=0.24\linewidth]{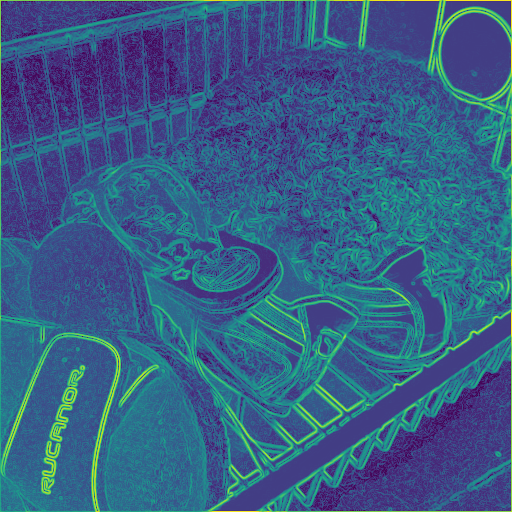} \\[2pt]
  \includegraphics[width=0.24\linewidth]{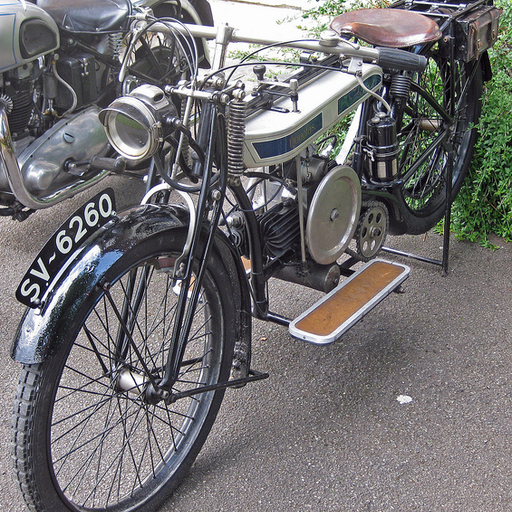} &
  \includegraphics[width=0.24\linewidth]{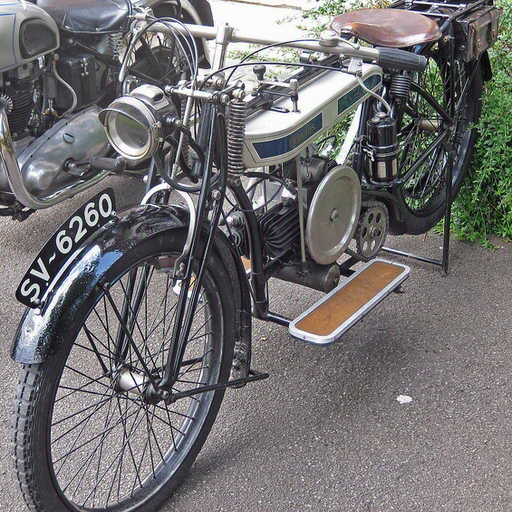} &
  \includegraphics[width=0.24\linewidth]{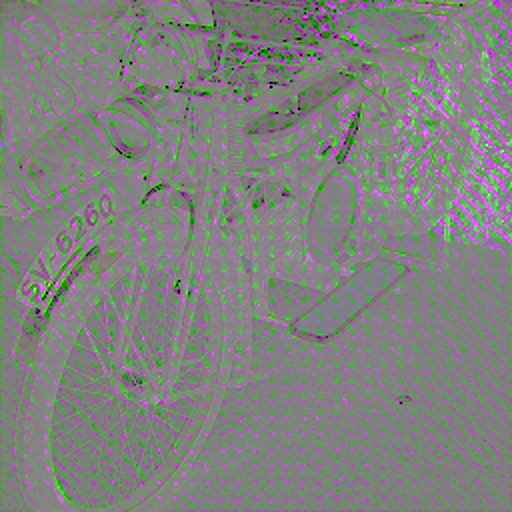} &
  \includegraphics[width=0.24\linewidth]{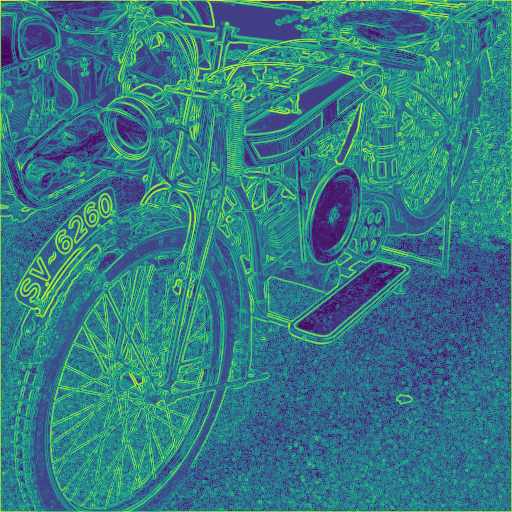} \\[2pt]
  \includegraphics[width=0.24\linewidth]{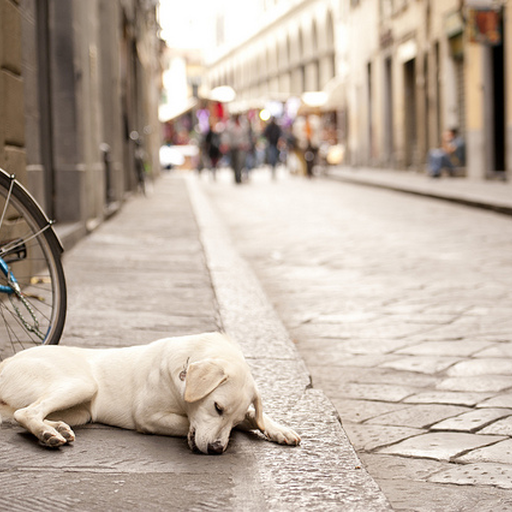} &
  \includegraphics[width=0.24\linewidth]{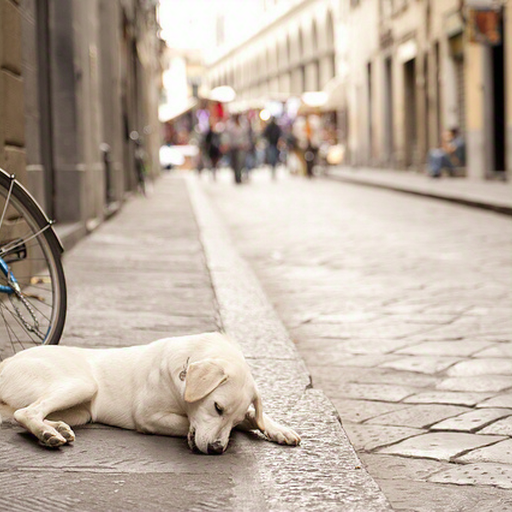} &
  \includegraphics[width=0.24\linewidth]{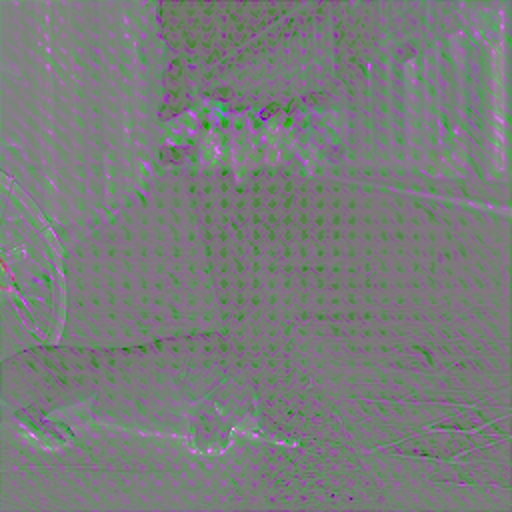} &
  \includegraphics[width=0.24\linewidth]{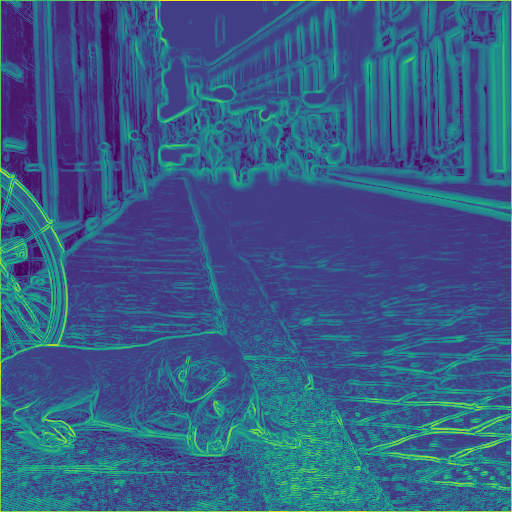} \\[2pt]
  \includegraphics[width=0.24\linewidth]{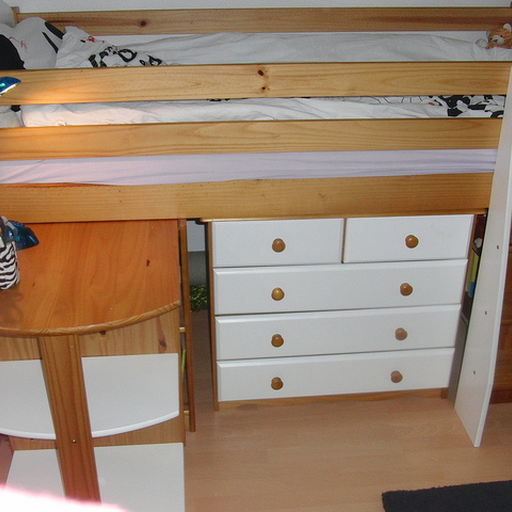} &
  \includegraphics[width=0.24\linewidth]{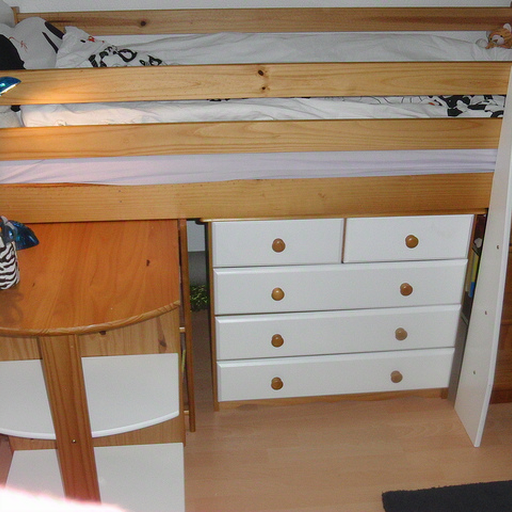} &
  \includegraphics[width=0.24\linewidth]{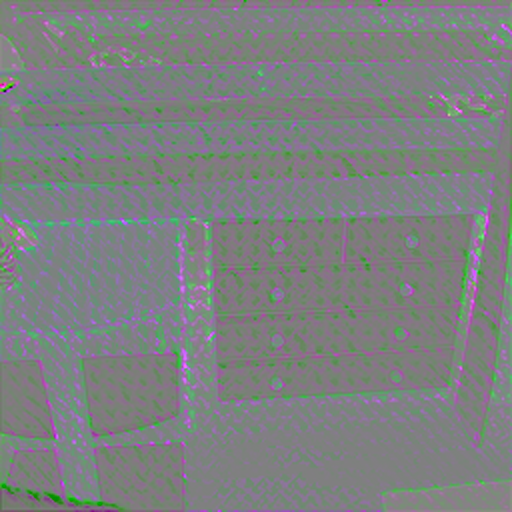} &
  \includegraphics[width=0.24\linewidth]{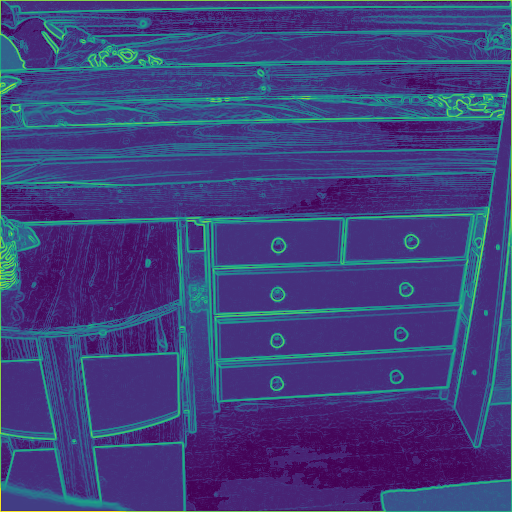} \\
\end{tabular}
\caption{\textbf{VectorMark invisibility analysis} across four MS-COCO test images. \textbf{Col~1:} Original image. \textbf{Col~2:} VectorMark watermarked image. \textbf{Col~3:} $10\times$ amplified residual. \textbf{Col~4:} JND spatial gain map~\cite{wu2017enhanced}. The raw multiplicative JND gain values (typically in the range of $[0.0001, 0.128]$) have been min-max normalized to $[0, 1]$ and rendered with a viridis colormap (bright yellow indicates higher perturbation tolerance) to make the spatial variance visible. The tight correlation between the residual footprint and the JND mask confirms that VectorMark dynamically steers embedding energy toward complex textures.}
\label{fig:supp_jnd_masks}
\end{figure}

\begin{figure*}[tbp]
\centering
\setlength{\tabcolsep}{2pt}
\begin{tabular}{@{} cccc @{}}
  \multicolumn{4}{@{}c}{\small\textbf{HiDDeN}} \\[-1pt]
  \multicolumn{2}{@{}c}{\scriptsize PSNR: 31.01 \textbar{} SSIM: 0.986 \textbar{} LPIPS: 0.056} &  
  \multicolumn{2}{c}{\scriptsize PSNR: 34.16 \textbar{} SSIM: 0.985 \textbar{} LPIPS: 0.079} \\[2pt]
  \includegraphics[width=0.245\linewidth]{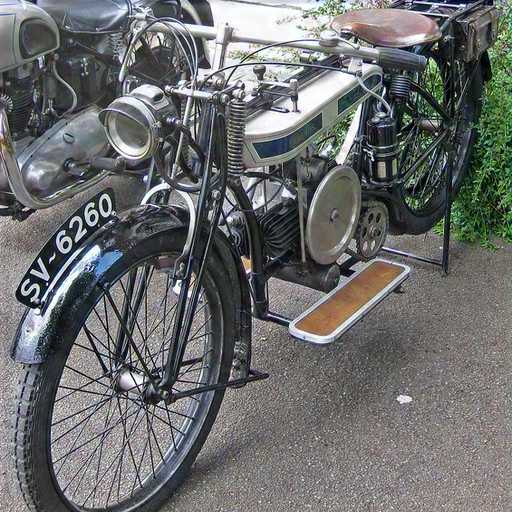} &
  \includegraphics[width=0.245\linewidth]{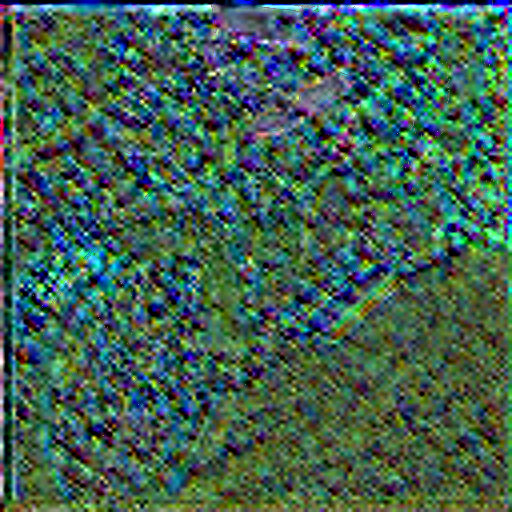} &
  \includegraphics[width=0.245\linewidth]{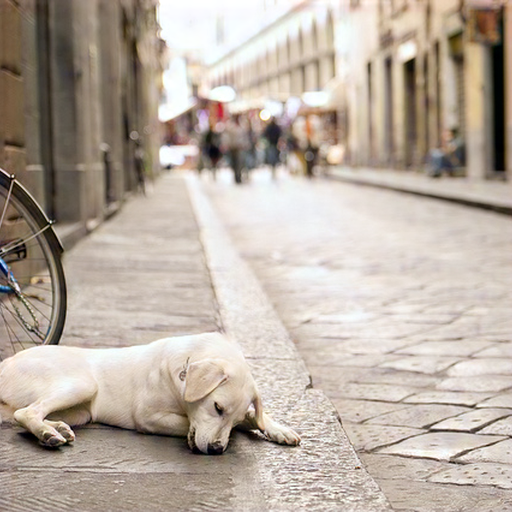} &
  \includegraphics[width=0.245\linewidth]{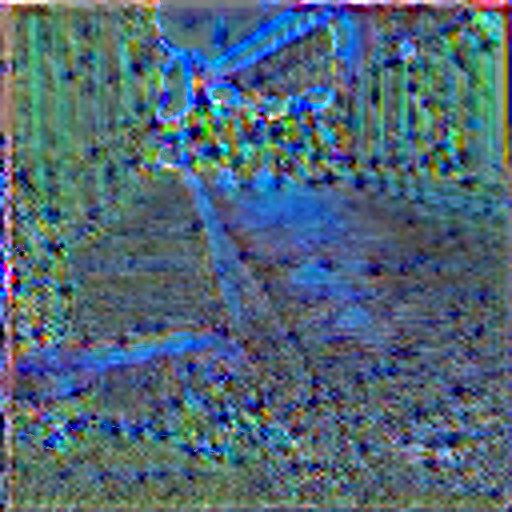} \\[5pt]

  \multicolumn{4}{@{}c}{\small\textbf{SS-DFT}} \\[-1pt]
  \multicolumn{2}{@{}c}{\scriptsize PSNR: 35.49 \textbar{} SSIM: 0.972 \textbar{} LPIPS: 0.051} &  
  \multicolumn{2}{c}{\scriptsize PSNR: 40.75 \textbar{} SSIM: 0.966 \textbar{} LPIPS: 0.152} \\[2pt]
  \includegraphics[width=0.245\linewidth]{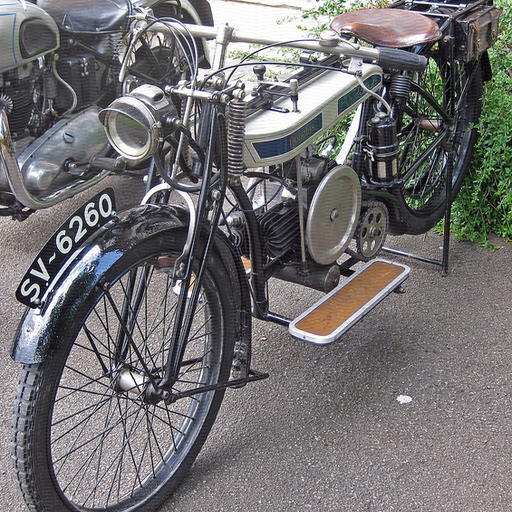} &
  \includegraphics[width=0.245\linewidth]{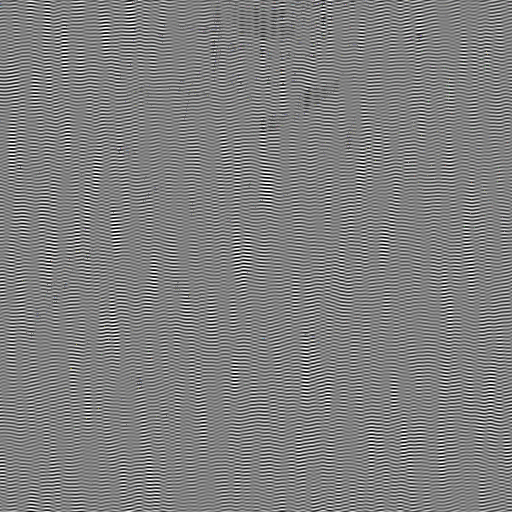} &
  \includegraphics[width=0.245\linewidth]{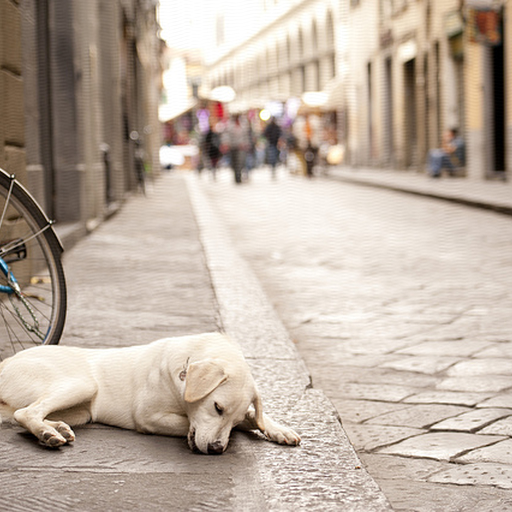} &
  \includegraphics[width=0.245\linewidth]{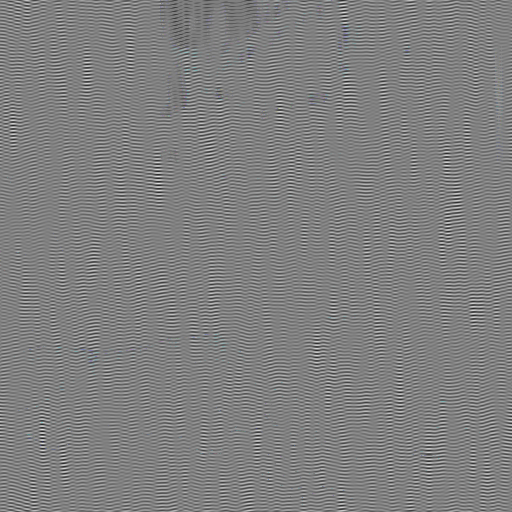} \\[5pt]

  \multicolumn{4}{@{}c}{\small\textbf{MaskWM}} \\[-1pt]
  \multicolumn{2}{@{}c}{\scriptsize PSNR: 36.07 \textbar{} SSIM: 0.992 \textbar{} LPIPS: 0.116} &  
  \multicolumn{2}{c}{\scriptsize PSNR: 36.87 \textbar{} SSIM: 0.977 \textbar{} LPIPS: 0.178} \\[2pt]
  \includegraphics[width=0.245\linewidth]{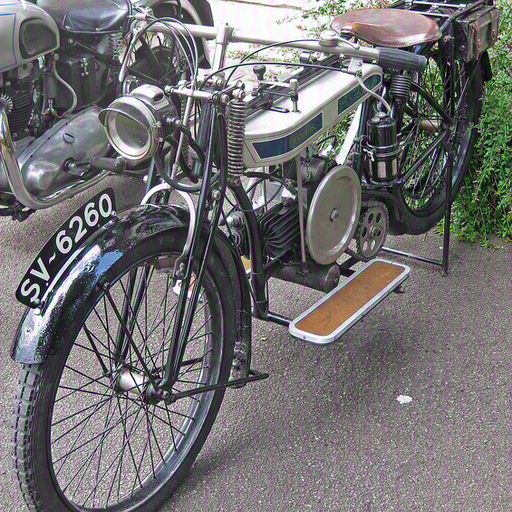} &
  \includegraphics[width=0.245\linewidth]{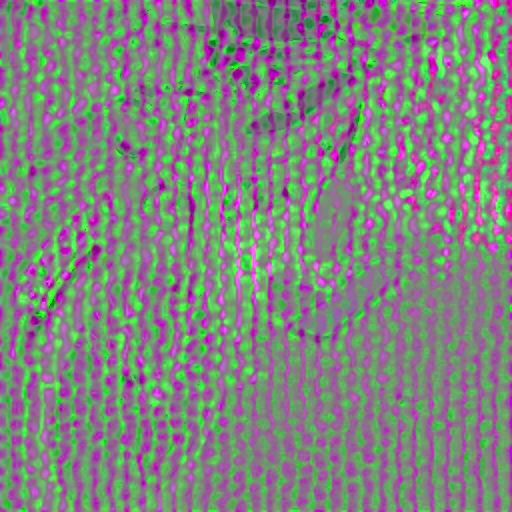} &
  \includegraphics[width=0.245\linewidth]{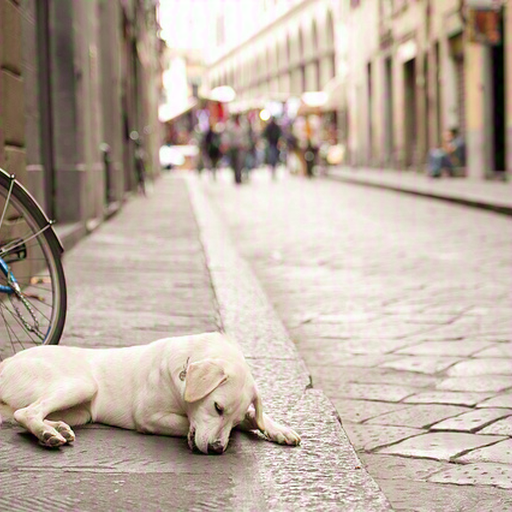} &
  \includegraphics[width=0.245\linewidth]{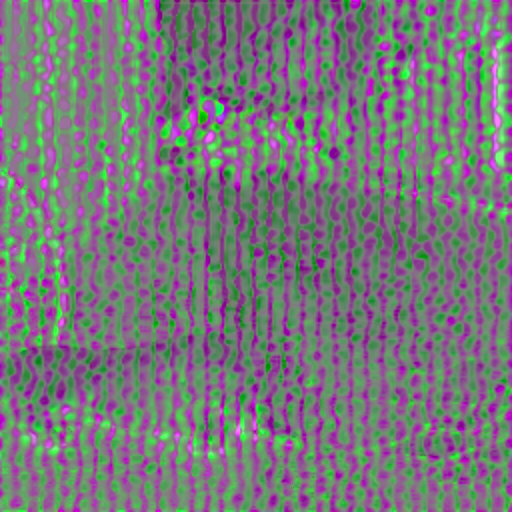} \\[5pt]

  \multicolumn{4}{@{}c}{\small\textbf{VINE-R}} \\[-1pt]
  \multicolumn{2}{@{}c}{\scriptsize PSNR: 37.33 \textbar{} SSIM: 0.996 \textbar{} LPIPS: 0.005} &  
  \multicolumn{2}{c}{\scriptsize PSNR: 38.14 \textbar{} SSIM: 0.993 \textbar{} LPIPS: 0.010} \\[2pt]
  \includegraphics[width=0.245\linewidth]{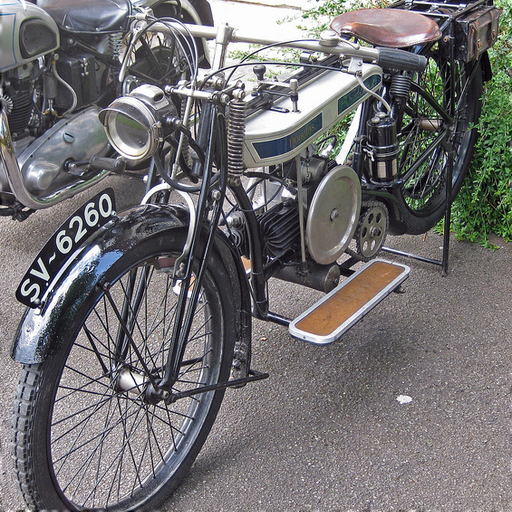} &
  \includegraphics[width=0.245\linewidth]{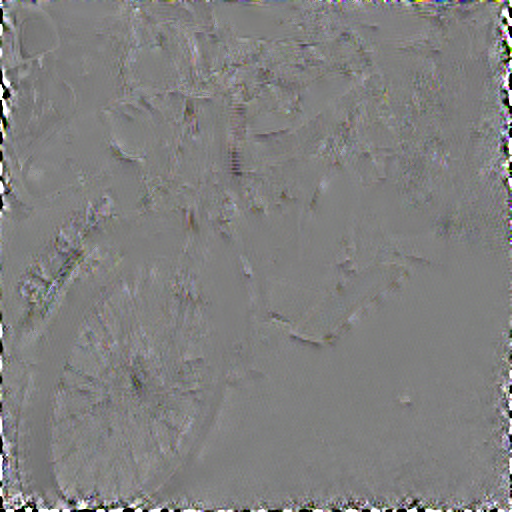} &
  \includegraphics[width=0.245\linewidth]{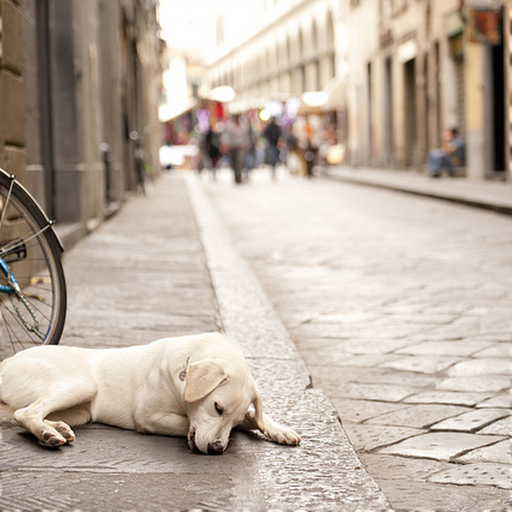} &
  \includegraphics[width=0.245\linewidth]{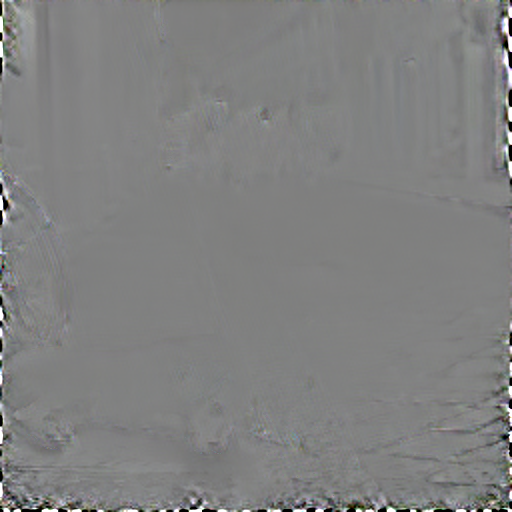} \\
\end{tabular}
\caption{\textbf{Qualitative invisibility comparison} across two MS-COCO validation images with exact per-image metrics. \textbf{Columns 1 \& 2:} Watermarked output and $10\times$ amplified residual for Image 1. \textbf{Columns 3 \& 4:} Watermarked output and $10\times$ amplified residual for Image 2. By viewing the amplified residuals side-by-side with their per-image metrics, structural failure modes become visible despite competitive PSNR scores.}
\label{fig:supp_vis_comparison}
\end{figure*}

\begin{figure*}[tbp]
\centering
\setlength{\tabcolsep}{2pt}
\begin{tabular}{@{} cccc @{}}
  \multicolumn{4}{@{}c}{\small\textbf{TrustMark-Q}} \\[-1pt]
  \multicolumn{2}{@{}c}{\scriptsize PSNR: 38.13 \textbar{} SSIM: 0.994 \textbar{} LPIPS: 0.009} &  
  \multicolumn{2}{c}{\scriptsize PSNR: 40.87 \textbar{} SSIM: 0.991 \textbar{} LPIPS: 0.028} \\[2pt]
  \includegraphics[width=0.245\linewidth]{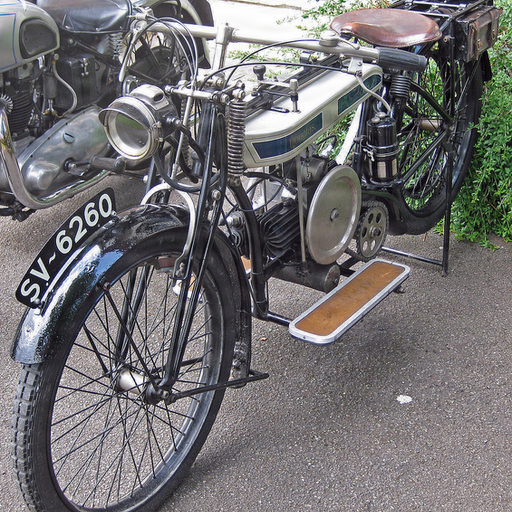} &
  \includegraphics[width=0.245\linewidth]{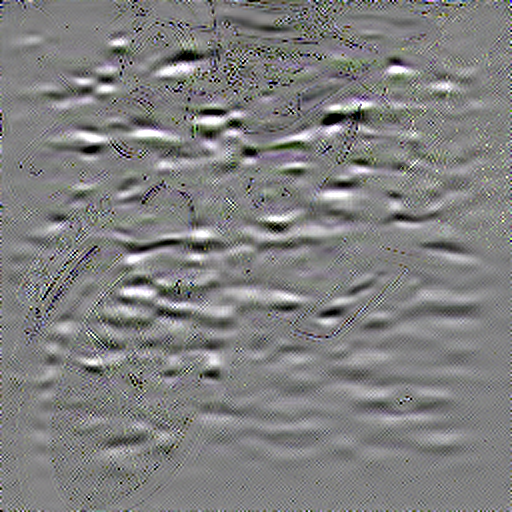} &
  \includegraphics[width=0.245\linewidth]{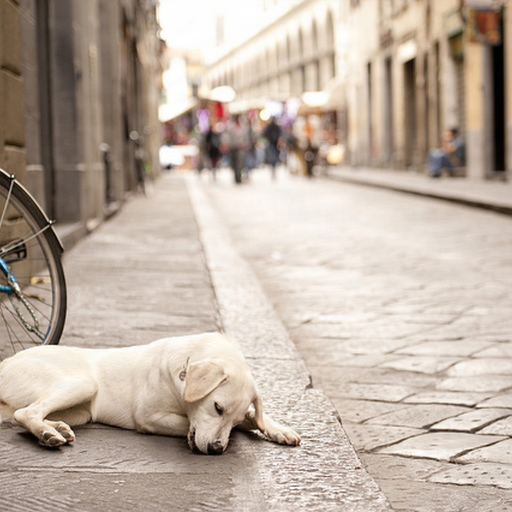} &
  \includegraphics[width=0.245\linewidth]{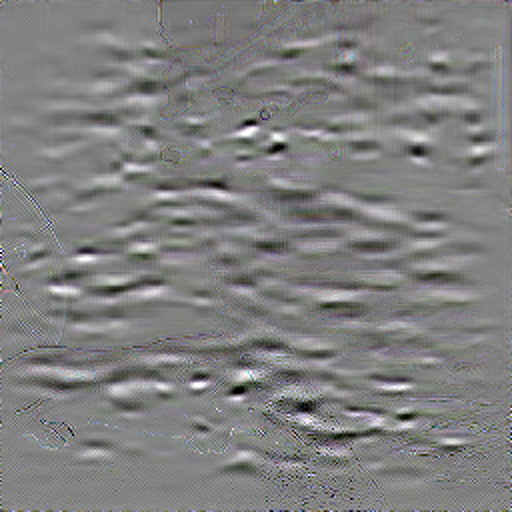} \\[5pt]

  \multicolumn{4}{@{}c}{\small\textbf{RivaGAN}} \\[-1pt]
  \multicolumn{2}{@{}c}{\scriptsize PSNR: 39.73 \textbar{} SSIM: 0.997 \textbar{} LPIPS: 0.054} &  
  \multicolumn{2}{c}{\scriptsize PSNR: 40.00 \textbar{} SSIM: 0.991 \textbar{} LPIPS: 0.098} \\[2pt]
  \includegraphics[width=0.245\linewidth]{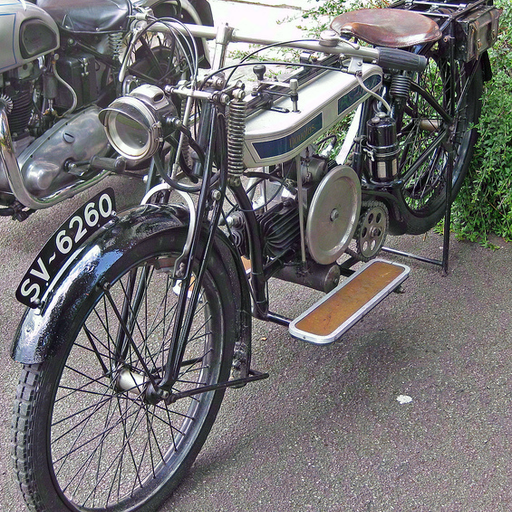} &
  \includegraphics[width=0.245\linewidth]{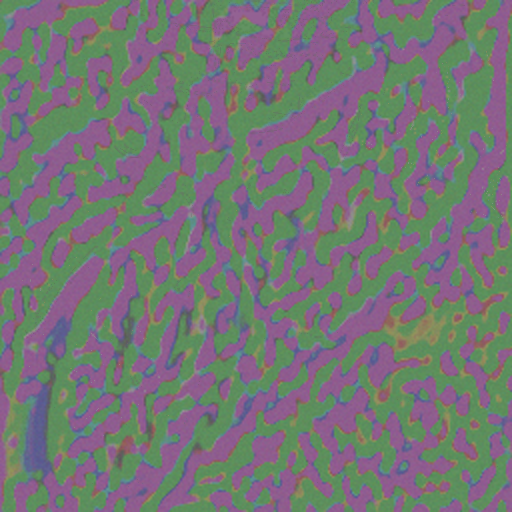} &
  \includegraphics[width=0.245\linewidth]{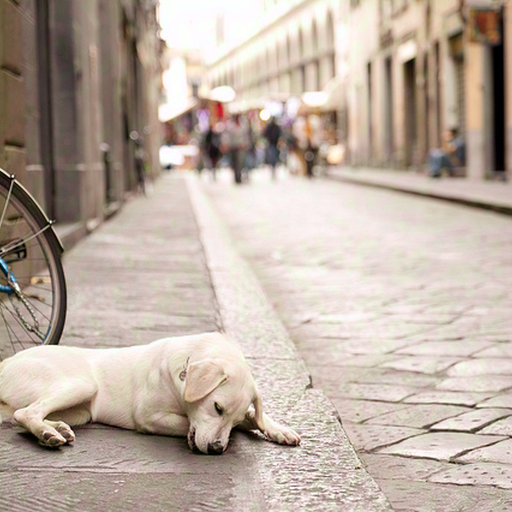} &
  \includegraphics[width=0.245\linewidth]{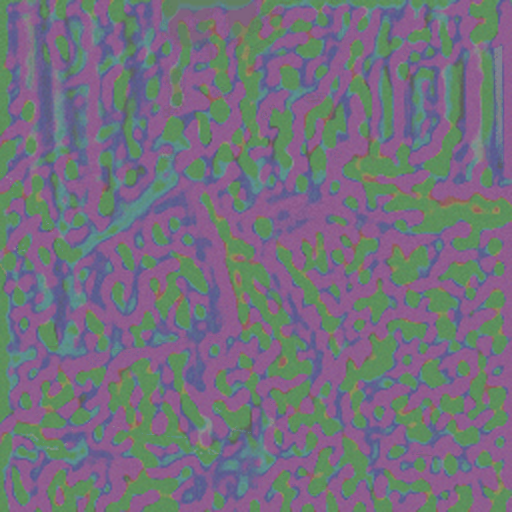} \\[5pt]

  \multicolumn{4}{@{}c}{\small\textbf{MBRS}} \\[-1pt]
  \multicolumn{2}{@{}c}{\scriptsize PSNR: 39.23 \textbar{} SSIM: 0.996 \textbar{} LPIPS: 0.014} &  
  \multicolumn{2}{c}{\scriptsize PSNR: 42.76 \textbar{} SSIM: 0.995 \textbar{} LPIPS: 0.022} \\[2pt]
  \includegraphics[width=0.245\linewidth]{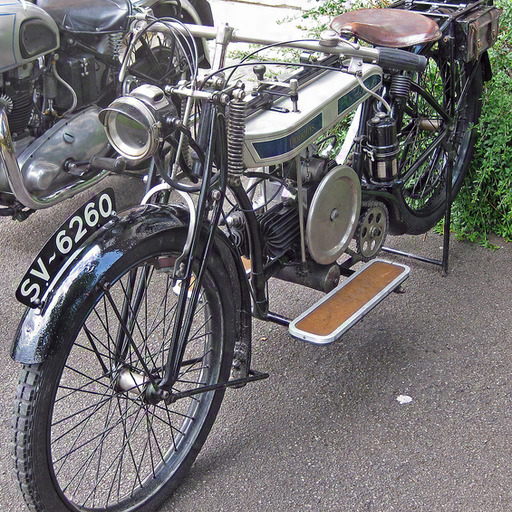} &
  \includegraphics[width=0.245\linewidth]{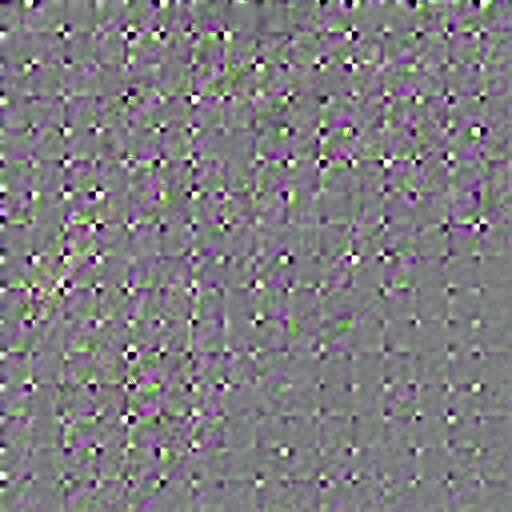} &
  \includegraphics[width=0.245\linewidth]{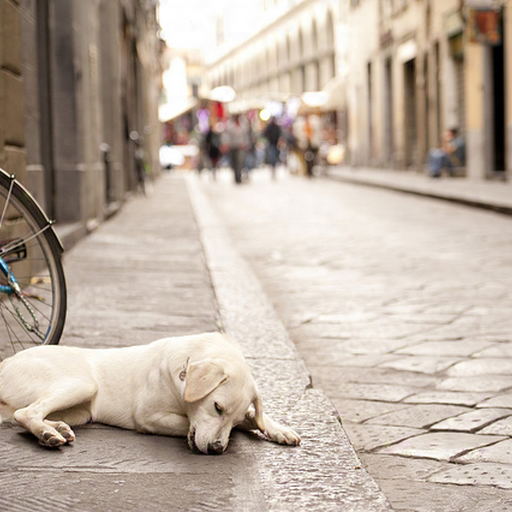} &
  \includegraphics[width=0.245\linewidth]{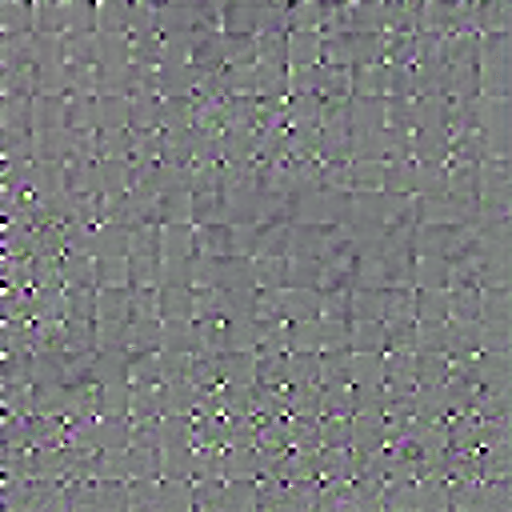} \\[5pt]

  \multicolumn{4}{@{}c}{\small\textbf{VectorMark (Ours)}} \\[-1pt]
  \multicolumn{2}{@{}c}{\scriptsize PSNR: 39.16 \textbar{} SSIM: 0.995 \textbar{} LPIPS: 0.032} &  
  \multicolumn{2}{c}{\scriptsize PSNR: 42.65 \textbar{} SSIM: 0.992 \textbar{} LPIPS: 0.042} \\[2pt]
  \includegraphics[width=0.245\linewidth]{images/invisibility/vectormark_best/img_01_watermarked.png} &
  \includegraphics[width=0.245\linewidth]{images/invisibility/vectormark_best/img_01_residual_10x.png} &
  \includegraphics[width=0.245\linewidth]{images/invisibility/vectormark_best/img_02_watermarked.png} &
  \includegraphics[width=0.245\linewidth]{images/invisibility/vectormark_best/img_02_residual_10x.png} \\
\end{tabular}
\caption{\textbf{Qualitative invisibility comparison} across two MS-COCO validation images with exact per-image metrics. \textbf{Columns 1 \& 2:} Watermarked and $10\times$ residual for Image 1. \textbf{Columns 3 \& 4:} Watermarked and $10\times$ residual for Image 2. By viewing the amplified residuals side-by-side with their per-image metrics, structural failure modes become visible despite competitive PSNR scores.}
\end{figure*}

% \newpage
% \clearpage
% \input{checklist.tex}

\end{document}